\documentclass{article}

\usepackage{arxiv}

\usepackage[utf8]{inputenc} 
\usepackage[T1]{fontenc}    
\usepackage{hyperref}       
\usepackage{url} 
\usepackage{amssymb}
\usepackage{pifont}
\usepackage{latexsym}
\usepackage{amsmath}
\usepackage{makecell}
\usepackage{multirow}
\usepackage{diagbox}
\usepackage{enumerate}
\usepackage{booktabs}       
\usepackage{amsfonts}       
\usepackage{nicefrac}       
\usepackage{microtype}      
\usepackage{lipsum}
\usepackage{graphicx}
\usepackage{subcaption}
\usepackage{xcolor}
\usepackage{algorithm}
\usepackage{algorithmic}
\graphicspath{ {./images/} }

\title{Initialization-enhanced physics-informed neural network with domain decomposition (IDPINN)}

\author{
  Chenhao Si \\
  School of Data Science\\
  The Chinese University of Hong Kong, Shenzhen (CUHK-Shenzhen)\\
  Shenzhen, 518172  \\
  \texttt{222042011@link.cuhk.edu.cn} \\
  \And
  Ming Yan* \\
  School of Data Science\\
  The Chinese University of Hong Kong, Shenzhen (CUHK-Shenzhen)\\
  Shenzhen, 518172  \\
  \texttt{yanming@cuhk.edu.cn} \\
}

\begin{document}
\maketitle
\begin{abstract}
We propose a new physics-informed neural network framework, IDPINN, based on the enhancement of initialization and domain decomposition to improve prediction accuracy. We train a PINN using a small dataset to obtain an initial network structure, including the weighted matrix and bias, which initializes the PINN for each subdomain. Moreover, we leverage the smoothness condition on the interface to enhance the prediction performance. We numerically evaluated it on several forward problems and demonstrated the benefits of IDPINN in terms of accuracy. 
\end{abstract}


\section{Introduction}
Physics-Informed Neural Network (PINN) has emerged as a promising approach to tackle both forward and inverse problems associated with Partial Differential Equations (PDEs)~\cite{Rasissietal2019}. PINNs incorporate the governing equation into the neural network's architecture at its core, augmenting the loss function with a residual term derived from the equation. This structural adjustment penalizes deviations from the equation, narrowing the range of viable solutions. Consequently, it transforms the process of determining PDE solutions into an optimization task focused on minimizing the loss function.

PINNs have demonstrated adaptability and effectiveness across various domains and interdisciplinary fields~\cite{Shuklaetal2020, Yinelta2021, Caielta202, Henkeselta2022}. Moreover, it terms out as a rapid expansion of research that applies across various scientific fields, such as fluid flows~\cite{RRaissietal2020, chew2024modelling, rao2020physics}, physics modeling~\cite{cai2021deepm, mao2021deepm}, solid mechanics~\cite{zhang2020physics, chen2021learning}, etc. Some recent works also used variation structures of PINNs in solving certain types of PDE~\cite{wang2024pinn}, including wave modeling~\cite{ren2024seismicnet}, multi-scale modeling in thermal conductivity~\cite{liu2024multi}, and hydrodynamic simulations~\cite{dai2024physics}.

As practical applications proliferate, interest in understanding the theoretical foundations of the PINN framework has surged. Researchers have embarked on meticulous studies to explore the convergence properties and error bounds of PINNs~\cite{Siddelta2023} to establish a robust theoretical foundation. These investigations seek to unravel the intricate mechanisms that drive PINN performance, anticipating that such understanding will contribute to developing more efficient and accurate models.

However, several challenges remain to be solved in PINNs. Many aspects of PINNs need to be improved, and these have attracted the attention of researchers in the field. One of the main limitations of PINN is its computationally intensive training process. Therefore, a key goal is accelerating model training while enhancing accuracy and efficiency without compromising the model's capabilities.

Numerous adapted variants of PINNs have shown promising progress in mitigating inaccuracies. For instance, Chakraborty et al. (2021) implemented a model of shared knowledge within PINNs, incorporating multiple levels of detail to enhance accuracy~\cite{Chakrabortyelta2021}. Additional research conducted by Desai et al. (2020), Goswami et al. (2020), Tang et al. (2022), and Xue et al. (2022) also demonstrated the successful application of shared knowledge in this context~\cite{Desaielta20, Goswamielta2020, Tangelta2022, Xuelta2205}. Furthermore, new constructions of PINNs have been developed, such as PINNs with enhanced gradients (gPINNs)~\cite{Yuelta2022, Mohammadian2022, Linelta2023}. These gPINNs have exhibited improved accuracy, particularly in training points for complex PDEs.

Notably, recent research, as exemplified by Lin et al. (2022) and Yang et al. (2021), has combined PINNs with Bayesian knowledge, revealing the limitations of the original structure of PINNs and the necessity of developing new frameworks \cite{Linkaelta2022, Yangelta2021}. Collectively, these works underscore the evolving landscape of PINNs, where innovation and adaptation are key to overcoming existing challenges and advancing the field.

To improve the accuracy and enhance the efficiency of solving both forward and inverse problems, researchers have adopted the domain decomposition approach~\cite{Kharazmielta2020}. Recent works showcase its effectiveness and practicality~\cite{ren2024seismicnet, yang2024adaptive, de2024physics, zhang2023multi}. This approach involves partitioning the entire domain into regular and irregular subdomains. Therefore, we can train each subdomain using different neural networks simultaneously and independently.

The benefits and advantages of domain decomposition can be summarized as follows \cite{Kharazmielta2020, Japtapelta2020, JaptapXelta2020}:

1. Distinct behaviors in varied domains: Systems show distinct behaviors in different domains in many real-world scenarios. Therefore, the specific physical characteristics may significantly diverge for different subdomains. For instance, this phenomenon is often observed in problems involving disruptive elements like shock waves.

2. Simplified neural network structures: Segmenting a vast domain into smaller subdomains helps reduce the complexity of the neural network structure required for each subdomain. Domain decomposition strategy simplifies the training process and makes it more efficient.

One significant challenge in domain decomposition for PINNs is that when some subdomains lack the necessary boundary conditions or initial conditions, PINNs (for the corresponding subdomains) will result in constant-zero predictions for certain PDEs. In this case, the interface conditions will be used as the boundary conditions to the corresponding subdomains enforced on the networks. Therefore, the interface conditions in the loss function become crucial in domain decomposition strategies.

One notable solution to this challenge is the conservative PINN (cPINN) framework, proposed by Japtapel et al. (2020) \cite{Japtapelta2020}. This framework divides the problem domain into subdomains and implements separate, localized high-capacity networks in these regions. The interface condition is established through flux continuity within the network. As a result, a set of individual local PINNs is generated, and each is adapted to the specific prior knowledge about the solution within its respective subdomain. The flux continuity condition effectively governs the behavior of predictions.

Expanding upon the cPINNs concept, Japtapel et al. further introduced the broadened PINNs (XPINNs) framework \cite{JaptapXelta2020}, which extends the previous approach to address general PDEs. XPINNs modify the loss function associated with the interface conditions. However, it should be noted that effectively managing the loss at the interfaces between subdomains remains a challenging task \cite{Huelta2020}. Consequently, the research landscape has witnessed the emergence of an adapted framework known as APINNs \cite{Hu2elta2022}, which builds upon the foundations of both PINNs and XPINNs. APINNs aim to address the challenges associated with interface conditions in domain decomposition strategies.

Despite significant advancements, the persistent challenges of high training costs and errors near the interface remain obstacles in domain decomposition approaches for PINNs. These challenges impact the efficiency and accuracy of the methodology, calling for innovative solutions. Recent studies conducted by Chuang et al. (2023) \cite{Chuangelta2023}, Ben et al. (2023) \cite{Benelta2023}, and Alena et al. (2023) \cite{Alenaelta2023} have shed light on the potential of domain decomposition to alleviate training cost problems and effectively handle errors occurring at the interface. However, these challenges persist, motivating researchers to explore new methods for improvement.

We proposed a new framework called Initialization-enhanced Physics-Informed Neural Network with Domain Decomposition (IDPINN) to address the above challenges. 
We train a PINN using a smaller dataset and fewer training steps. The trained model structures are then used to initialize the networks for each subdomain, demonstrating greater efficiency compared to the standard PINN model. Additionally, we adjust the loss functions and enforce a smoothness condition at the interface, significantly reducing the interface error.

The contribution and novelty of this paper can be summarized as follows:

\begin{enumerate}[ \textbullet]
    \item \textbf{Initialization enhancement}: Instead of using a random initialization for our network, we use the network construction from a previously trained PINN. This initialization strategy proves to be highly effective in enhancing the performance of the PINN, reducing both training costs and the relative error significantly.
    \item \textbf{Improvement for the domain decomposition}: We present a novel model that employs domain decomposition and new loss functions. Our empirical investigations reveal significant enhancements when used for network training with the domain decomposition approach.
\end{enumerate}

The organization of this paper is as follows: In Section~\ref{Sec2}, we provide a brief review of the PINN framework. The framework is introduced in Section~\ref{Sec3}, and the experimental results are shown in Section~\ref{Sec4}. Specifically, we apply our framework to the Helmholtz equation, 2D Poisson equation, heat equation. and viscous Burgers equation We compare our results with those obtained using PINN and XPINN. 
\section{Problem setup}
\label{Sec2}
\subsection{Neural networks}
We denote $\mathcal{NN}:\mathbb{R}^n \rightarrow \mathbb{R}^m$ as a feed-forward neural network consisting of $L$ hidden layers, where its $l^{th}$ hidden layer has $N_l$ neurons ($1 \leq l \leq L$). In particular, the input and output layers have $n$ and $m$ neurons, respectively. The weight matrix and bias vector in the $l^{th}$ hidden layer are denoted as $\mathbf{W}^l$ and $\mathbf{b}^l$, respectively. Let $h^{l}$ be the nonlinear activation function in the $l^{th}$ hidden layer, then the output vector of the $l^{th}$ hidden layer $\mathbf{z}^l$ is obtained by:
\begin{align}
    \mathbf{z}^l = h^l(\mathbf{W}^l \mathbf{z}^{l-1} + \mathbf{b}^l).
\end{align}
The input vector of the neural network is its input layer $\mathbf{z}^0$. The output of the neural network is
\begin{align}
    u_{\theta}(\mathbf{z}^0) = \mathcal{NN}(\mathbf{z}^0; \theta),
\end{align}
where $\theta = \{\mathbf{W}^l, \mathbf{b}^l\}_{l=1}^L$ is the collection set of all weight matrix and bias vectors of the neural network. For convenience, we may write $\mathcal{NN}(\mathbf{z})$ instead of $\mathcal{NN}(\mathbf{z};\theta)$ in the remaining sections when there is no confusion.

\subsection{Partial differential equations}
A partial differential equation (PDE) with its solution $u(\mathbf{x},t)$ takes the general form as follows:
\begin{align}
    \mathcal{F}(u,\mathbf{x},t)&=0, \quad\qquad \mathbf{x}\in\Omega\subseteq\mathbb{R}^n,~t\in[0,T]\label{(1)}\\
    u(\mathbf{x},t) &= f(\mathbf{x},t), \quad \mathbf{x}\in\partial\Omega,~t\in[0,T]\label{(2)}\\
    u(\mathbf{x},0) &= g(\mathbf{x}),  \label{(3)}
\end{align}
where $\Omega$ denotes the spatial domain with its boundary $\partial\Omega$. We also denote $\overline{\Omega} = \Omega\cup\partial\Omega$. Besides, $\mathbf{x}$ stands for the spatial coordinates such that $\mathbf{x} = (x_1,x_2,x_3,\ldots,x_n)\in\mathbb{R}^n$, $t$ stands for the time coordinate, and $\mathcal{F}$ is the differential operator. Equations \eqref{(2)} and \eqref{(3)} ensure the boundary and initial conditions for the solution $u$, respectively, where $f(\mathbf{x},t)$ and $g(\mathbf{x})$ are functions on the boundary with time and the spatial domain without time, respectively.
Here is an example of a PDE system, which is so-called the 1D viscous Burgers equation:
\begin{align}
&u_t + u u_x =\frac{0.01}{\pi} u_{xx}, \quad x\in[-1,1],~t\in[0,1],\label{Burgers1}\\
&u(-1, t) = u(1, t) = 0,\label{Burgers3}\\
&u(x, 0) = -\sin(\pi x),~\quad x\in (-1,1).\label{Burgers2}
\end{align}
The differential operator is $\mathcal{F}(u,\mathbf{x},t) = u_t + u u_x -\frac{0.01}{\pi} u_{xx}$ with spatial domain $\Omega = [-1,1]$ and its boundary $\partial\Omega = \{-1,1\}$.
\subsection{Physics-informed neural networks (PINNs)}
The fundamental concept of PINN is to utilize a neural network $\mathcal{NN}(\mathbf{x},t; \theta)$ with parameters $\theta$ to represent the unidentified solution $u(\mathbf{x}, t)$.
Let $\{(\mathbf{x}_k^R, t_k^R)\}_{k=1}^{N_R}$ be the set of $N_R$ training residual points in the domain $\Omega\times(0,T)$, $\{(\mathbf{x}_k^B, t_k^B)\}_{k=1}^{N_B}$ be the set of $N_B$ boundary points on the boundary $\partial\Omega\times(0,T)$, and $\{(\mathbf{x}_k^I, 0)\}_{k=1}^{N_I}$ be the set of $N_I$ initial points. The predicted solution from the neural network is denoted as $\hat{u}(\mathbf{x}, t)$. We construct the loss function with PDE residues, the boundary conditions, and initial conditions as:
\begin{align}
    \mathcal{L}_R(\theta) &=\frac{1}{N_R}\sum_{k=1}^{N_R}|\mathcal{F}(\hat{u}, \mathbf{x}_k^R, t_k^R)|^2, \label{5}\\
    \mathcal{L}_B(\theta) &= \frac{1}{N_B}\sum_{k=1}^{N_B}| \hat{u}(\mathbf{x}_k^B, t_k^B) - f(\mathbf{x}_k^B,t_k^B)|^2,\label{6}\\
    \mathcal{L}_I(\theta) &=\frac{1}{N_I}\sum_{k=1}^{N_I}| \hat{u}(\mathbf{x}_k^I, 0) - g(\mathbf{x}_k^I)|^2.\label{7}
\end{align}
Here $\mathcal{L}_R(\theta)$, $\mathcal{L}_B(\theta)$, and $\mathcal{L}_I(\theta)$ are called the residue loss of the PDE, boundary loss, and initial loss, respectively. The loss function for the standard PINN is defined as:
\begin{align}
\mathcal{L(\theta)} = \lambda_1\mathcal{L}_R(\theta)+\lambda_2\mathcal{L}_B(\theta)+\lambda_3\mathcal{L}_I(\theta),\label{4}
\end{align}
where $\lambda_1$, $\lambda_2$ and $\lambda_3$ are weights corresponding to the three losses.

\section{Our proposed framework for domain decomposition}
\label{Sec3}

We briefly introduce domain decomposition and the motivation for a new framework in Section~\ref{Sec3.1}. Section~\ref{Sec3.2} introduces a new framework named Initialization-enhanced Physics-Informed Neural Network with Domain Decomposition (IDPINN), whose overall structure is shown in Fig.~\ref{Fig: Structure of IDPINN}. By decomposing the entire domain into several disjoint subdomains, we build an individual neural network on each subdomain and train the neural networks together. As shown in Fig.~\ref{Fig: Structure of IDPINN}, the loss function comprises six parts, whose details will be explained later.

\subsection{Domain decomposition: issues and motivation}
\label{Sec3.1}
The main idea of domain decomposition is dividing a domain into several subdomains, and functions on different subdomains are represented by different functions. These subdomains could be disjoint or overlapped. In this paper, we focus on the disjoint case. Using domain decomposition in PINNs can utilize parallel computing to tackle larger-scale problems. Moreover, learning functions on a relatively small domain enhances the accuracy and stability of PINN~\cite{Kharazmielta2020, Japtapelta2020, JaptapXelta2020}.

Suppose that the entire domain $\Omega$ is decomposed into $M$ non-overlapping subdomains $\{\Omega_m\}_{m=1}^M$. We denote the interface between two adjacent subdomains $\Omega_i$ and $\Omega_j$ as $\partial\Omega_{ij}$ (here we let $i<j$ for the uniqueness). We denote $\{(\mathbf{x}^i_k, t^i_k)\}_{k=1}^{N_{f_i}}$ as the set of data points within the subdomain $i$, i.e., $(\mathbf{x}_k^i,t_k^i)\in \Omega_i\times (0, T)$. Moreover, to impose the interface condition, we denote $\{(\mathbf{x}^{ij}_k, t^{ij}_k)\}_{k=1}^{N_{ij}}$ as the set of $N_{ij}$ data points randomly generated on the interface $\partial\Omega_{ij}$. 
The total data points generated are denoted as $\{(\mathbf{x}^*, t^*)\}$. 

It is not surprising to use the difference of $\hat{u}_i(\mathbf{x}^{ij}_k,t^{ij}_k)$ and $\hat{u}_j(\mathbf{x}^{ij}_k,t^{ij}_k)$ as an additional penalty term, where $\hat{u}_i(\mathbf{x}^{ij}_k,t^{ij}_k)$ and $\hat{u}_j(\mathbf{x}^{ij}_k,t^{ij}_k)$ stand for the predictions on the interface $\partial\Omega_{ij}$ from the PINNs to the $\Omega_i$ and $\Omega_j$, respectively. This penalty term is as follows:
\begin{align}  
\mathcal{L}_{inter}(\theta) =&\sum_{ij}\frac{1}{N_{ij}}\sum_{k=1}^{N_{ij}} |\hat{u}_i(\mathbf{x}^{ij}_k,t^{ij}_k) - \hat{u}_j(\mathbf{x}^{ij}_k,t^{ij}_k)|^2. \label{(13)}
\end{align}
Here, we add the loss functions for all interfaces together. The notation $\sum_{ij}$ means that the summation is over all interfaces.
There are other types of loss functions to enforce the consistency between two adjacent subdomains. For instance, the remarkable decomposition model XPINN \cite{JaptapXelta2020} uses
\begin{align*}
\mathcal{L}_{inter}(\theta) = \lambda_{residual}\mathcal{L}_{residual}(\theta)+\lambda_{avg}\mathcal{L}_{avg}(\theta),
\end{align*}
where
\begin{align}  
\mathcal{L}_{residual}(\theta) =& \sum_{ij}\frac{1}{N_{ij}}\sum_{k=1}^{N_{ij}}|\mathcal{F}(\hat{u}_i, \mathbf{x}^{ij}_{k},t^{ij}_k)-\mathcal{F}(\hat{u}_j, \mathbf{x}^{ij}_{k},t^{ij}_k)|^2,\label{8}\\
\mathcal{L}_{avg}(\theta) =&\sum_{ij}\frac{1}{N_{ij}}\sum_{k=1}^{N_{ij}} \left|\hat{u}_i(\mathbf{x}^{ij}_k,t^{ij}_k) - \frac{\hat{u}_i(\mathbf{x}^{ij}_k,t^{ij}_k)+\hat{u}_j(\mathbf{x}^{ij}_k,t^{ij}_k)}{2}\right|^2.\label{99}
\end{align}

In the original XPINN~\cite{JaptapXelta2020}, the authors use $\theta_i$ as the parameters of the PINN for the subdomain $i$ and construct the loss function subdomain-wise assuming that the parameters for other PINNs are known. Here, we denote $\theta$ as the parameters for all PINNs in the domain for convenience. The loss functions introduced above are equivalent to the functions in XPINN. Note that the term $\mathcal{L}_{avg}(\theta)=\mathcal{L}_{inter}(\theta)/4$, where $\mathcal{L}_{inter}(\theta)$ is defined in equation~(\ref{(13)}).

\begin{figure}[!htb]
\centering
    \begin{minipage}{0.8\textwidth}
     \centering
     \includegraphics[width=.95\linewidth]{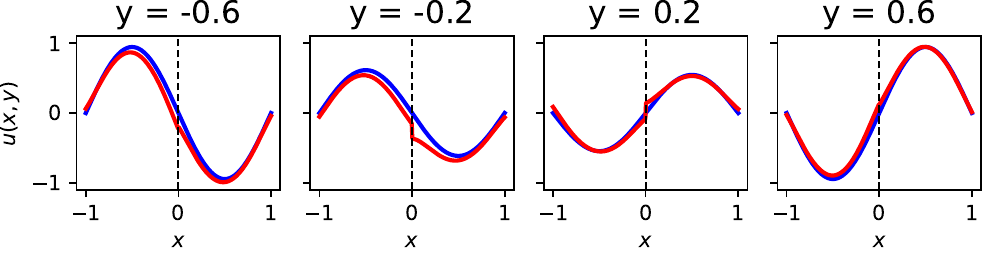}  
   \end{minipage} 
       \begin{minipage}{0.8\textwidth}
     \centering
     \includegraphics[width=.95\linewidth]{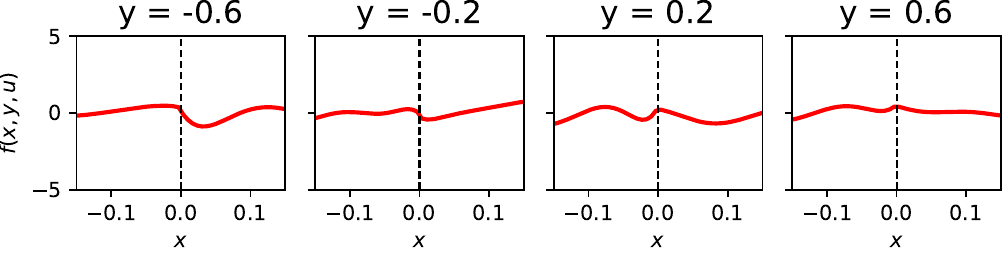}  
   \end{minipage}
   \begin{minipage}{0.8\textwidth}
     \centering
     \includegraphics[width=.95\linewidth]{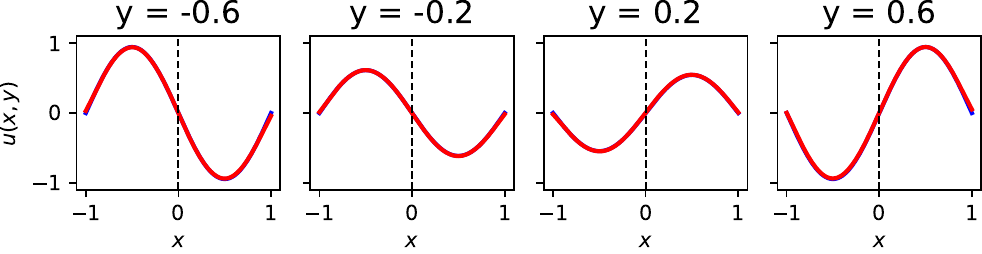} 
   \end{minipage}
    \begin{minipage}{0.8\textwidth}
     \centering
     \includegraphics[width=.95\linewidth]{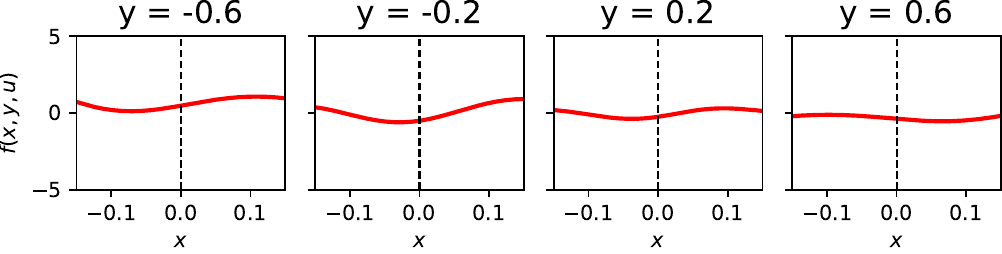}  
   \end{minipage}
    \caption{Slices of the exact (blue) and predicted (red) solution ($u(x,y)$) of the Helmholtz equation for given $y$ values, and their corresponding PDE residuals ($f(x,y,u)$). First and second row: We use the loss functions on the interface: $\mathcal{L}_{residual}(\theta)$~\eqref{8} and $\mathcal{L}_{avg}(\theta)$~\eqref{99}. Third and fourth row: We use IDPINN-3. ($\mathcal{L}_{\nabla}(\theta)$~\eqref{9} and $\mathcal{L}_{PDE_g}(\theta)$~\eqref{10}).}\label{pointwise error interface}
\end{figure}

We apply XPINN on the Helmholtz equation by dividing the domain $\Omega = [-1,1]\times [-1,1]$ into two subdomains with the line $x = 0$ being the interface (more details on this experiment can be found in Section~\ref{Sec4.1}) and plot some results in the first two rows of Fig.~\ref{pointwise error interface}. In the first row, we plot the output function of XPINN for four different $y$ values. We can see the changes in the slop near the interface between both subdomains. This figure shows that only enforcing consistency in the function value may cause a non-smooth connection across the interface. To be fair enough, we test it under several choices of the weights $\lambda$ and plot one of the best results.
Also, we plot the residual functions of PDE $f(x,y,u) =\mathcal{F}(u,\mathbf{x})$ for the same $y$ values in the second row of Fig.~\ref{pointwise error interface}. Though the true residuals are zero in the whole domain, the residuals from XPINN are not smooth around the interface. This observation shows that the consistency in the function and residual across the interface may not be sufficient, which motivates us to use high-order continuity.

\subsection{Our proposed IDPINN}
\label{Sec3.2}
Motivated by the above experiment, which shows that XPINN introduces non-smoothness across the interface, we propose to penalize the difference between the gradients of the function and the residual at the interface to ensure smoothness. We introduce the new loss function as follows:
\begin{align}
\mathcal{L}(\theta) = \lambda_1\mathcal{L}_r(\theta)+\lambda_2\mathcal{L}_b(\theta)+\lambda_3\mathcal{L}_I(\theta)+\lambda_4\mathcal{L}_{inter}(\theta)+\lambda_5\mathcal{L}_{\nabla}(\theta)+\lambda_6\mathcal{L}_{PDE_g}(\theta),
\end{align}
where $\mathcal{L}_{inter}(\theta)$ is defined in equation (\ref{(13)}), and
\begin{align}  
\mathcal{L}_{\nabla}(\theta) =& \sum_{ij}\frac{1}{N_{ij}}\sum_{k=1}^{N_{ij}}\|\nabla \hat{u}_i(\mathbf{x}_k^{ij},t_k^{ij}) - \nabla \hat{u}_j(\mathbf{x}_k^{ij},t_k^{ij})\|^2_2,\label{9}\\
\mathcal{L}_{PDE_g}(\theta) = &\sum_{ij}\frac{1}{N_{ij}}\sum_{k=1}^{N_{ij}}\|\nabla \mathcal{F}(\hat{u}_i, \mathbf{x}_k^{ij},t_k^{ij}) - \nabla \mathcal{F}(\hat{u}_j, \mathbf{x}_k^{ij},t_k^{ij})\|_2^2.\label{10}
\end{align}


The loss function $\mathcal{L}_{inter}(\theta)$ ensures continuity across the interface, 
and we add the smoothness term $\mathcal{L}_{\nabla}(\theta)$ at the interface, which plays a crucial role in our model. 
However, for the residuals, we replace the consistency term at the surface~\eqref{8} with the consistency in its gradient~\eqref{9}. We do not keep the consistency term~\eqref{8} since we already have the residual loss term~\eqref{5} within each subdomain. 

Besides the changes in the loss function, we introduce an initialization stage,  referred to as the IDPINN-init stage, to reduce the overall training time. During this stage, We randomly select a subset of data points $\{(\mathbf{x}^{init},t^{init})\}\subset \{(\mathbf{x}^*, t^*)\}$, which includes both boundary and initial data points, and train a single PINN using this small dataset for a few iterations. The trained parameters are denoted as $\theta^0$. Then, we use $\theta^0$ as the initial parameters for PINNs across all subdomains during the IDPINN-main stage. 
\begin{figure}[!ht]
   \begin{minipage}{0.95\textwidth}
    \centering
     \includegraphics[width=0.8\linewidth]{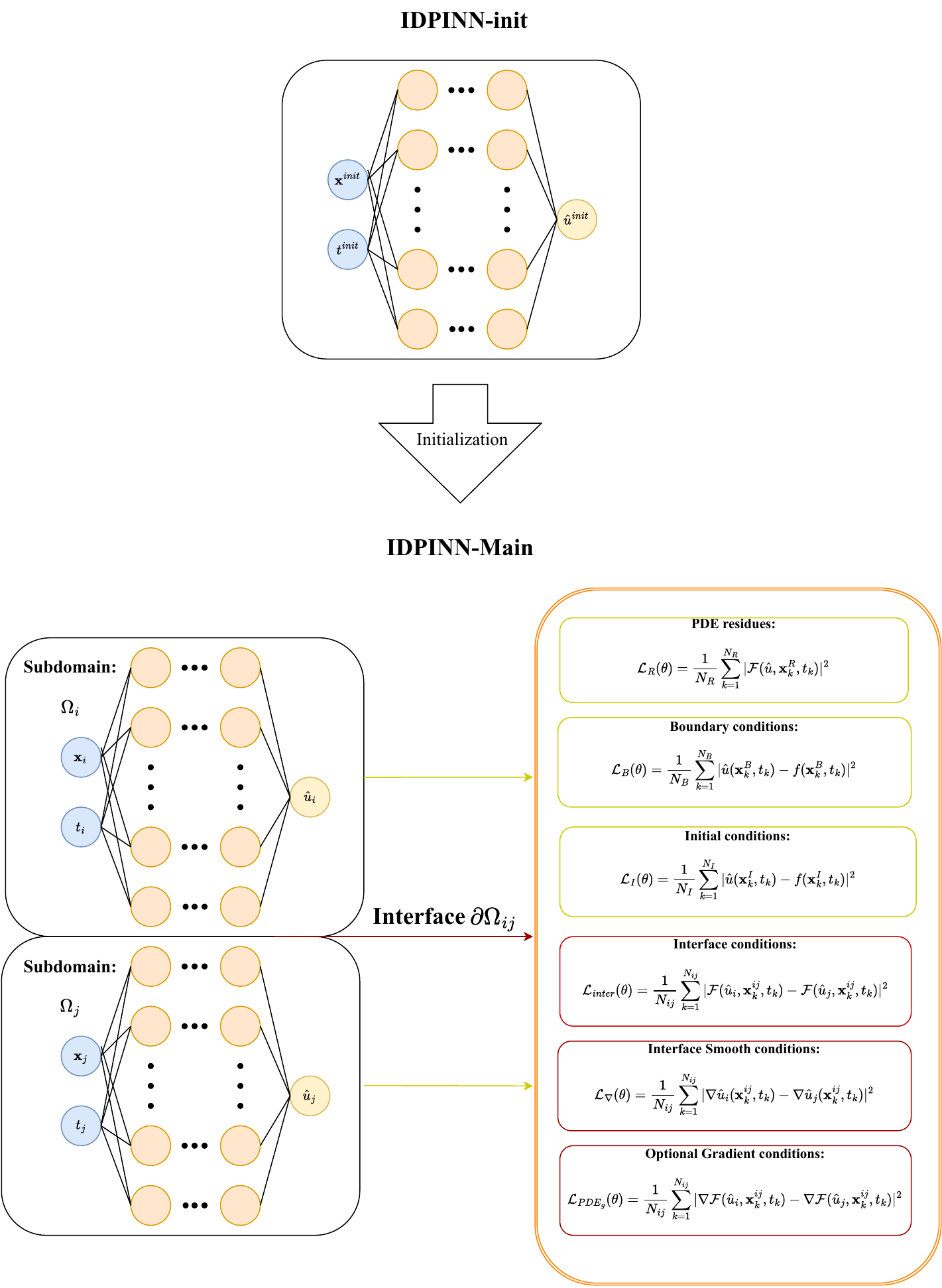}  
   \end{minipage}
    \caption{Structure of IDPINN, where each subdomain uses a neural network to approximate the function on the subdomain, and these neural networks are integrated through interface conditions. The PINN to the subdomains $\Omega_i$ and $\Omega_j$ have different parameters $\theta_i$ and $\theta_j$, respectively. Here, due to the simplicity, we use $\theta$.}\label{Fig: Structure of IDPINN}
\end{figure}
Fig.~\ref{Fig: Structure of IDPINN} shows the overall structure of our model. For simplicity, we use $\theta_k$ to represent the PINN parameters on subdomain $k$, and $\theta$ encompasses all parameters.

We apply IDPINN to the same Helmholtz equation problem discussed in the previous subsection, shown in the third and fourth rows of Fig.~\ref{pointwise error interface}. In the third row, the red curve (prediction) closely aligns with the entire blue curve (exact), exhibiting no discernible discontinuity or non-smoothness, even near the interface. In the fourth row, the influence of adding $\mathcal{L}_{PDE_g}(\theta)$ is further depicted. The curve exhibits overall smoother behavior, with no discontinuous shock occurring at the interface. The residual functions are also closer to zero compared with the second row. This improvement is attributed to the smoothness at the interface, thereby enabling each subdomain's PINN to benefit from a more stable interface condition.

\section{Numerical experiments}
\label{Sec4}
There are six terms in IDPINN: the first three terms come from PINN (PDE residual, boundary condition, and initial condition); the fourth and fifth terms are used to promote continuity and smoothness of the function across interfaces, respectively; the last term promotes the smoothness of the function and the PDE residual across interface using higher-order information. We keep the first four terms in the following experiments and demonstrate the benefit of adding the smoothness terms~\eqref{9} and~\eqref{10}. We denote IDPINN-1 as the model with the smoothness term \eqref{9} only, IDPINN-2 as the model with the smoothness term \eqref{10} only, and IDPINN-3 as the model with both~\eqref{9} and \eqref{10}. The differences of these three models are summarized in Table \ref{Table}. 
\begin{table}[!h]
  \centering
  \caption{Loss term information for IDPINN-1, IDPINN-2, and IDPINN-3.}
  \begin{tabular}{|c|c|c|c|}
    \hline
    Model &  IDPINN-1 &  IDPINN-2 & IDPINN-3  \\
    \hline
    Smoothness term $\mathcal{L}_{\nabla}(\theta)$    & \ding{51}  & \ding{55} & \ding{51}  \\
    \hline
    Smoothness term $\mathcal{L}_{PDE_g}(\theta)$ & \ding{55}  & \ding{51} & \ding{51} \\
    \hline
  \end{tabular}
  \label{Table}
\end{table} 

We use the following relative L2 error (referred to as L2 error) to measure the accuracy of the models:
\begin{align}
    \text{L2-error} = \frac{\sqrt{\sum_{k=1}^N|\hat{u}(\mathbf{x}_k, t_k) - u(\mathbf{x}_k, t_k)|^2}}{\sqrt{\sum_{k=1}^N |u(\mathbf{x}_k, t_k)|^2}},
\label{L2error}
\end{align}
where $\hat{u}$ and $u$ are predicted and exact solutions, respectively. The set of points $\{(\mathbf{x}_k, t_k)\}_{k=1}^N$ used to compute the L2 error will be explained in the subsequent experiments.


\subsection{Helmholtz equation}
\label{Sec4.1}
The Helmholtz partial differential equation (PDE) is frequently utilized to describe wave and diffusion phenomena. It is used to depict the development of wave propagation and diffusion processes within spatial or spatiotemporal domains. In this investigation, we focus on a particular Helmholtz PDE that exists exclusively within the spatial (x, y) domain by:
\begin{align}
&u_{xx} + u_{yy} + u - q(x,y)=0, \quad x\in[-1,1], y\in[-1,1],\\
&u(-1,y) = u(1,y) = 0,\\
&u(x, -1) = u(x, 1) = 0,
\end{align}
where
\begin{align}
q(x,y)& = -\pi^2\sin(\pi x)\sin(4\pi y) - (4\pi)^2\sin(\pi x)\sin(4\pi y) + \sin(\pi x)\sin(4\pi y),
\end{align}
with an exact solution
\begin{align}
    u(x, y) = \sin(\pi x) \sin(4\pi y) .
\end{align}

We partition the domain along the line $x=0$ and evaluate the performance of IDPINN-3 and XPINN using a 5-layer network with 55 neurons per layer. We generate a uniform grid of size $300\times 300$ across the domain $[-1,1]\times [-1,1]$ to calculate the L2 error. Additionally, we uniformly sample 5000 points along the interface $x=0$. We select training points from the total of 95000 generated points as follows: Initially, 200 boundary points and 3000 residual points are randomly chosen from the grid for each subdomain. Subsequently, 200 points are randomly selected from the interface $x=0$.

\begin{figure}[!htb]
\centering
    \begin{minipage}{0.5\textwidth}
     \centering
     \includegraphics[width=.95\linewidth]{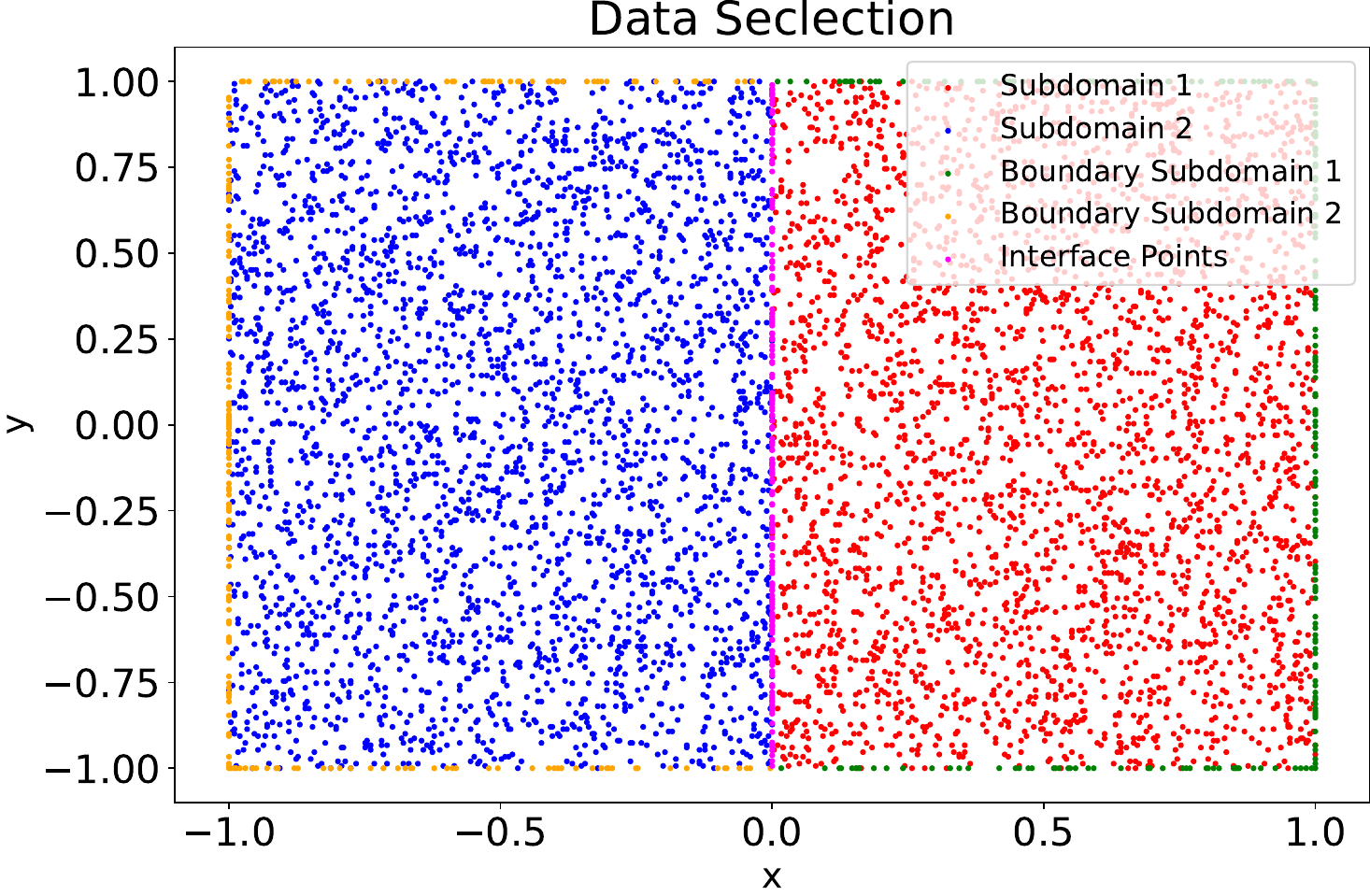}  
   \end{minipage}  
    \caption{Point selection for the Helmholtz equation with two subdomains. The selected points in each subdomain are represented by red (Subdomain 1) and blue (Subdomain 2), respectively. The boundary points in two subdomains and interface points are represented by green, orange, and purple, respectively.}
    \label{Helm data}
\end{figure}

The weights for the loss function in IDPINN-3 are specified as follows: $\lambda_1=1$, $\lambda_2 = 10$, $\lambda_3 = 0$, $\lambda_4 = 20$, $\lambda_5 = 2$, and $\lambda_6 = 5$. Among the 6400 points selected on the grid, we chose 80 boundary points and 500 residual points during the IDPINN-init stage. Subsequently, the model is trained using these 6600 points for 98,000 iterations via the Adam optimizer, employing a learning rate of $8\times 10^{-5}$. The resulting prediction L2 error amounts to $1.1\times 10^{-2}$. Please refer to Table~\ref{Table 4} and Fig.~\ref{Helm data} for detailed arrangement information.

\begin{figure}[!htb]
      \centering
      \begin{minipage}{0.45\textwidth}
      \centering
      \includegraphics[width=0.98\linewidth, height=0.65\textwidth]{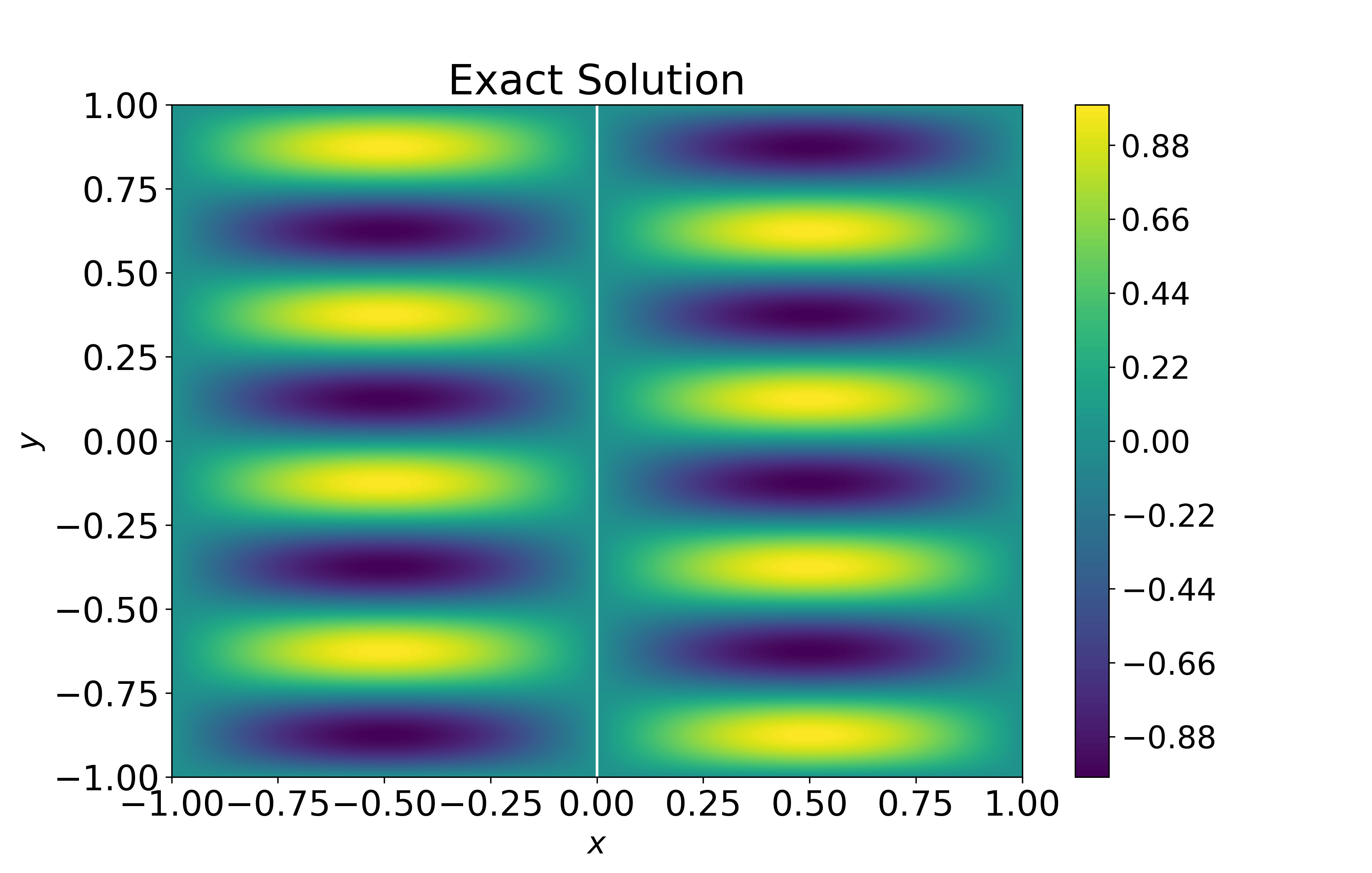}    
      \end{minipage}
      \vspace{1em}

      \begin{minipage}{0.45\textwidth}
      \centering
      \includegraphics[width=0.98\linewidth, height=0.65\textwidth]{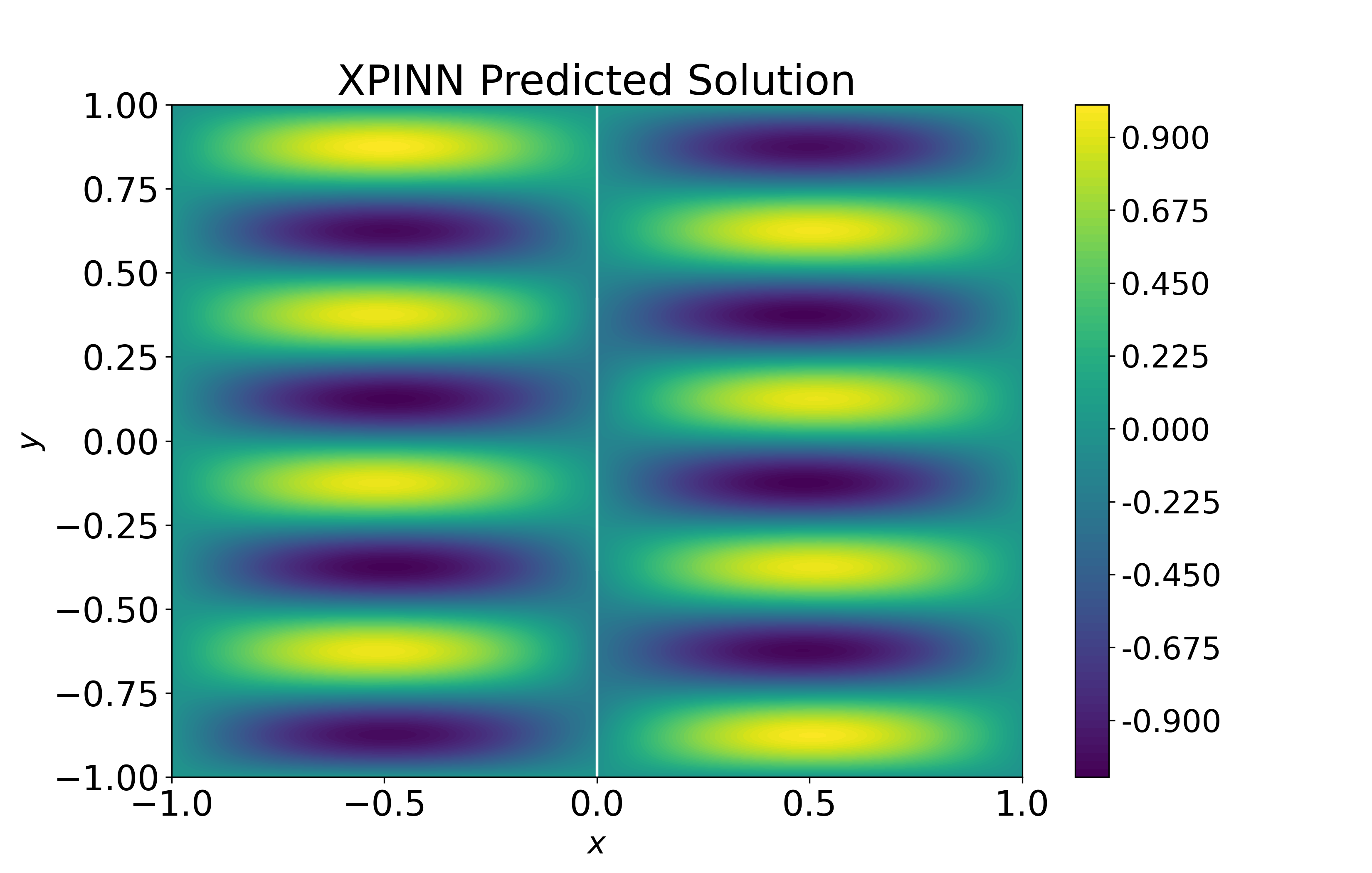}    
      \end{minipage}
      \begin{minipage}{0.45\textwidth}
      \centering
      \includegraphics[width=\linewidth, height=0.65\textwidth]{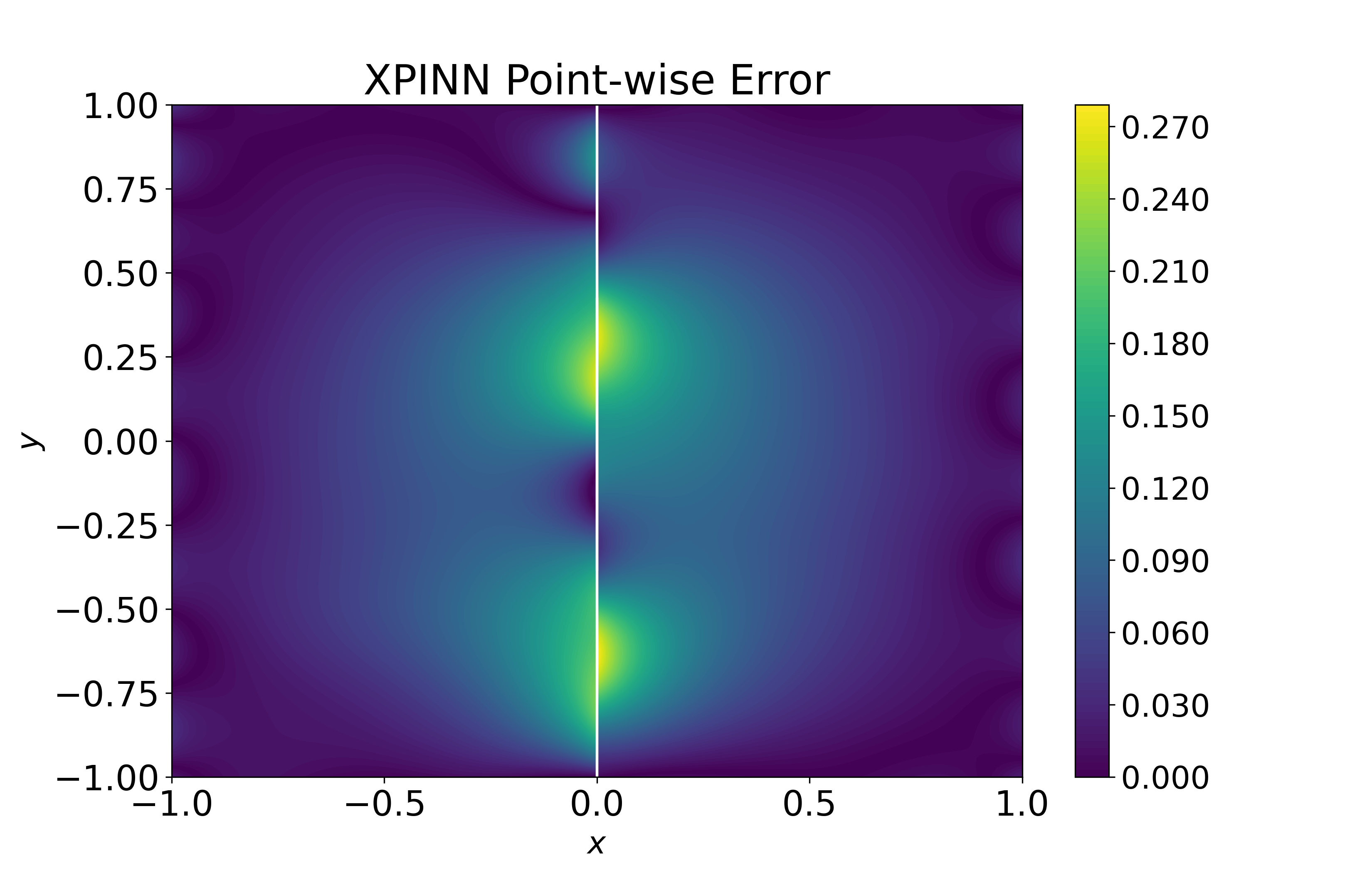}
      \end{minipage}
      \vspace{1em} 
    
      \begin{minipage}{0.45\textwidth}
      \centering
      \includegraphics[width=\linewidth, height=0.65\textwidth]{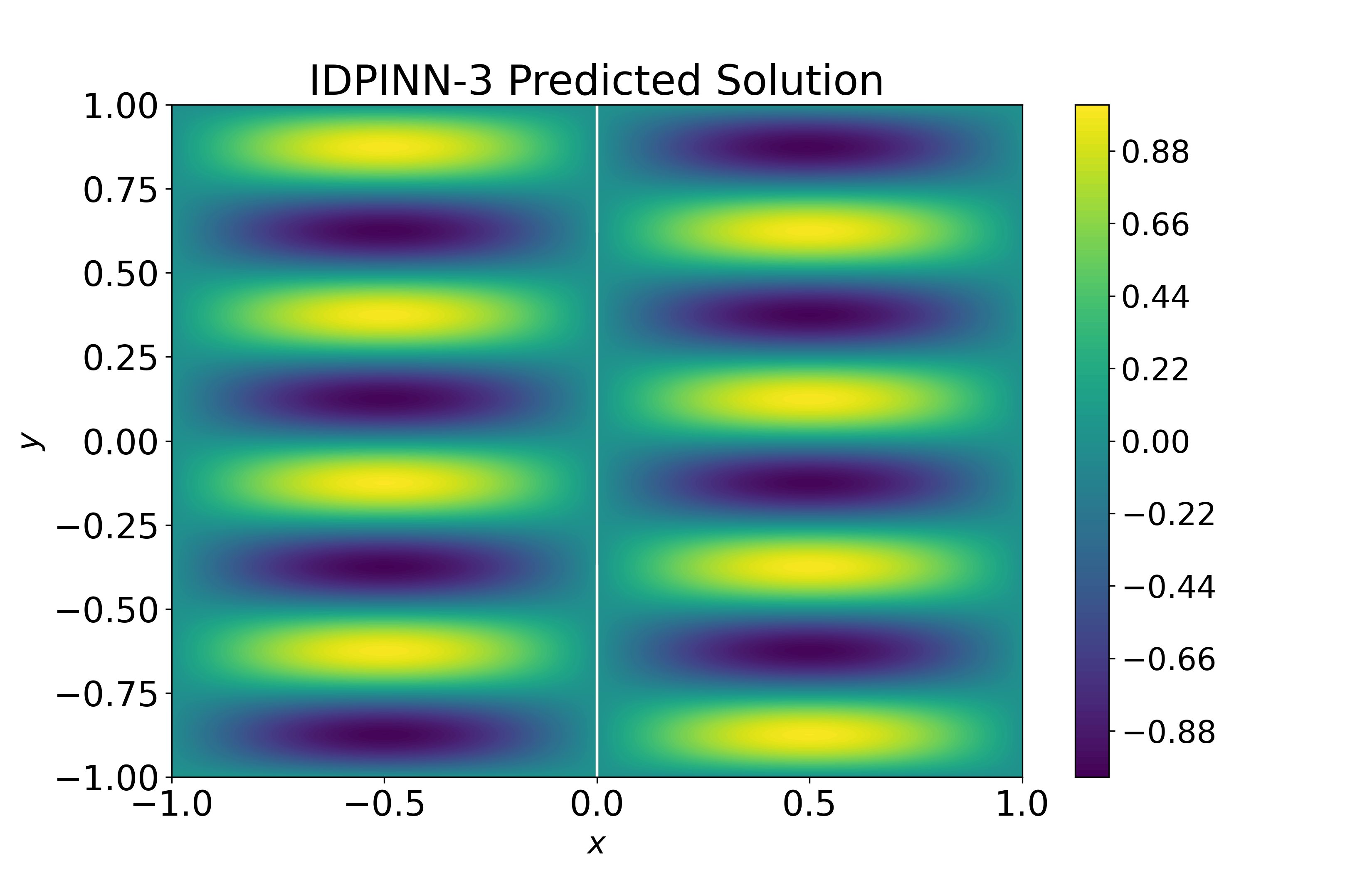}  
      \end{minipage}
      \begin{minipage}{0.45\textwidth}
      \centering
      \includegraphics[width=\linewidth, height=0.65\textwidth]{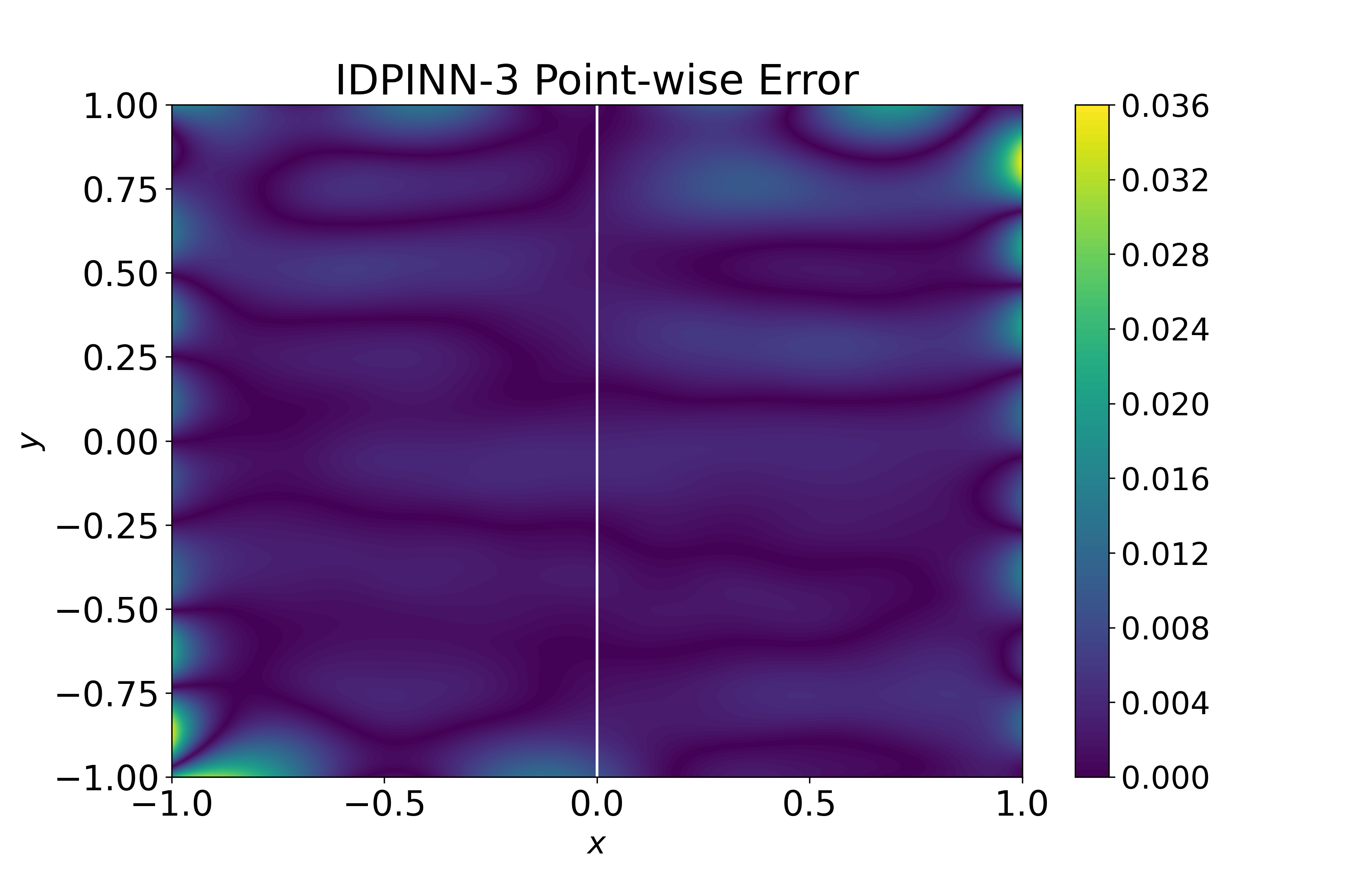}    
      \end{minipage}
    \begin{minipage}{0.45\textwidth}
      \centering
      \includegraphics[width=\linewidth, height=0.65\textwidth]{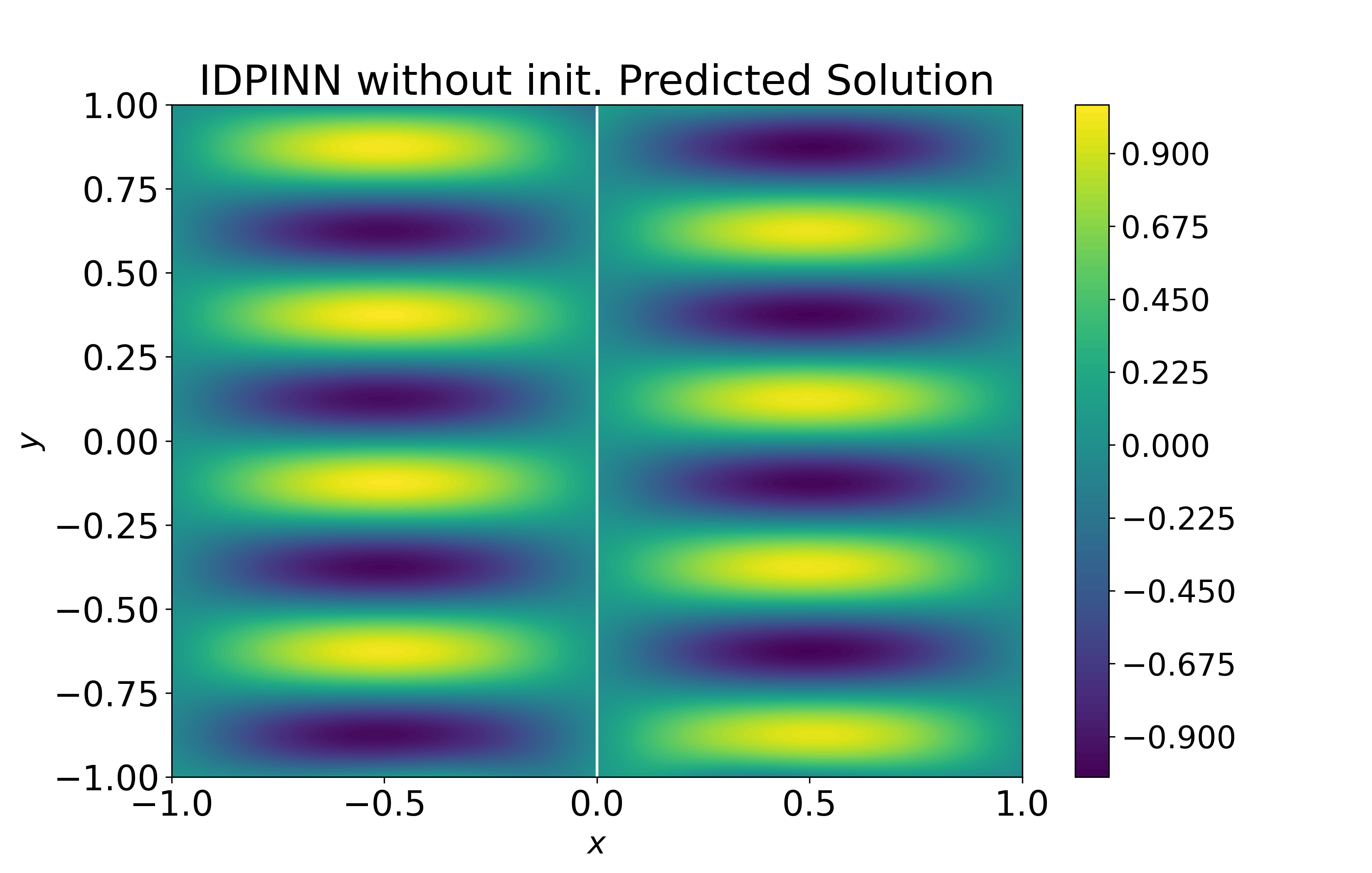}  
   \end{minipage}
    \begin{minipage}{0.45\textwidth}
      \centering
      \includegraphics[width=\linewidth, height=0.65\textwidth]{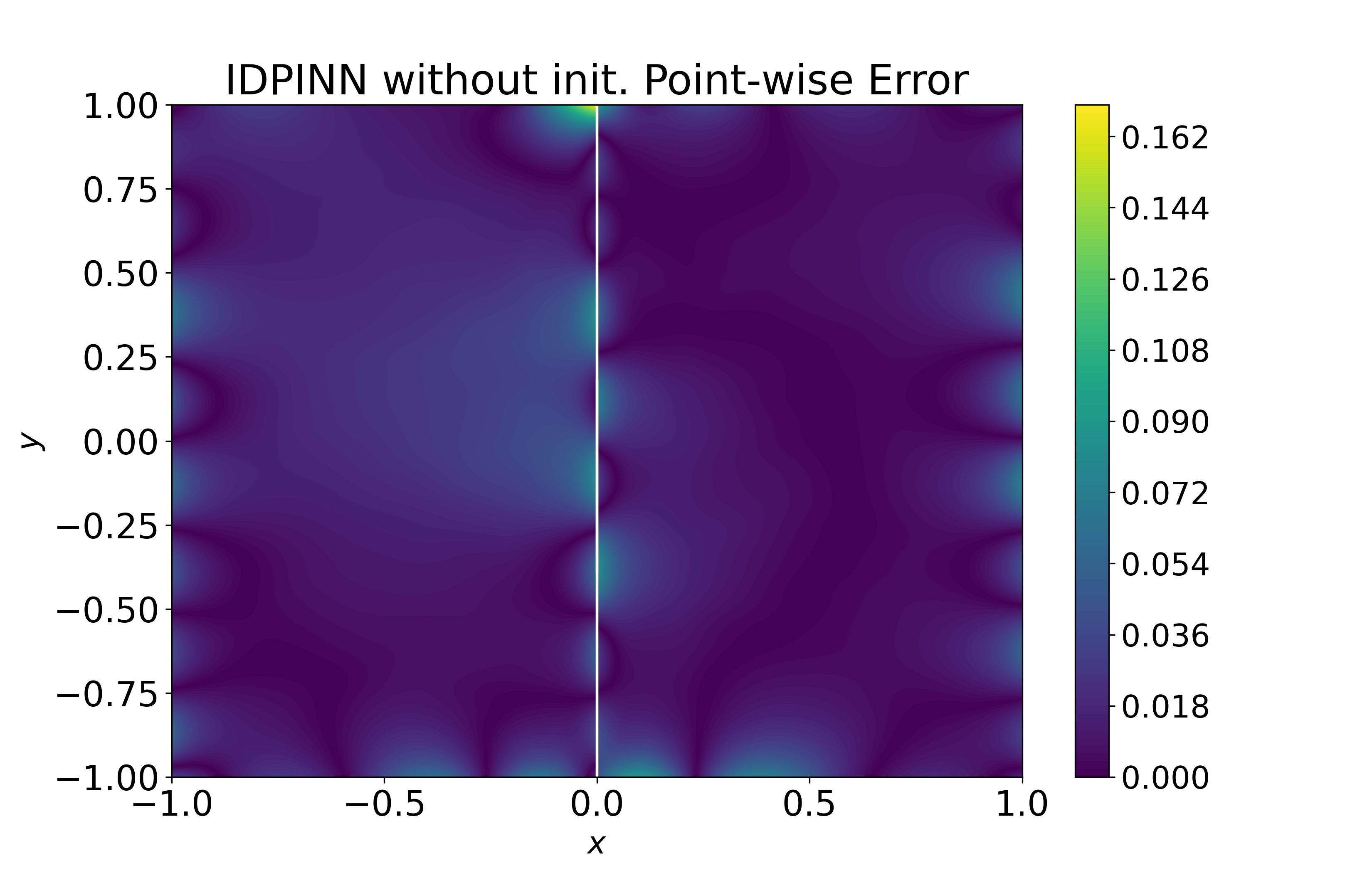}    
    \end{minipage}
     \caption{First row: exact solution of the Helmholtz equation. Second row: predicted solution and point-wise error for XPINN after 100k total iterations, resulting in a $1.41\times 10^{-1}$ L2 error. Third row: predicted solution and point-wise error for IDPINN-3 after 100k total iterations, resulting in a $1.1\times 10^{-2}$ L2 error. Last row: predicted solution and point-wise error for IDPINN-3 without initialization after 100k total iterations, resulting in a $4.82\times 10^{-2}$ L2 error.}\label{Helm result}
\end{figure}

Additionally, we explore various weight configurations for XPINN, where $\lambda_1 = 1$, $\lambda_2 = 20$, $\lambda_3 = 0$, $\lambda_{residual} = 20$, and $\lambda_{avg} = 80$ yield a comparatively lower L2 error ($1.41\times 10^{-1}$). These configurations are utilized to produce Fig.~\ref{pointwise error interface}. Fig.~\ref{Helm result} presents the visualization of the predictions. In our initial experiments, we adopted the same learning rate as that utilized for IDPINN-3. However, we observed that the L2 error plateaued after approximately 20k iterations. Consequently, we increased the learning rate to $1\times 10^{-2}$, which resulted in enhanced performance. Fig.~\ref{Helm Error history} illustrates the L2 error history of these models under both a lower learning rate ($8\times 10^{-5}$) and a higher learning rate ($1\times 10^{-2}$).

\begin{table}[!h]
  \centering
  \caption{Number of points and iterations in each stage (Helmholtz equation). The points for the IDPINN-init stage are collected from the training set of IDPINN-main.}
  \begin{tabular}{|c|c|c|c|c|c|c|}
    \hline
    stage & region &  residual  &  boundary & initial  & interface & iterations\\
    \hline
    IDPINN-init & all-region  & 500  & 80 & \multirow{3}{*}{NA} & NA & 2000\\
    \cline{1-4} \cline{6-7}
 \multirow{2}{*}{IDPINN-main}  & sub-domain 1 (red) & 3000  & 200 &  & \multirow{2}{*}{200} & \multirow{2}{*}{98000}\\
    \cline{2-4}
    & sub-domain 2 (blue) & 3000 & 200 &  & & \\
    \hline
    \end{tabular}
    \label{Table 4}
\end{table}

\subsubsection{Effect of the Initialization}
\label{Sec4.4.1}
In this section, we discuss the importance of initialization. With the same setting as above, we train IDPINN-3 without initialization for $100k$ iterations and compare it to IDPINN-3 with initialization.

\begin{figure}[!htb]
\centering
   \begin{minipage}{0.85\textwidth}
     \centering
     \includegraphics[width=0.6\linewidth]{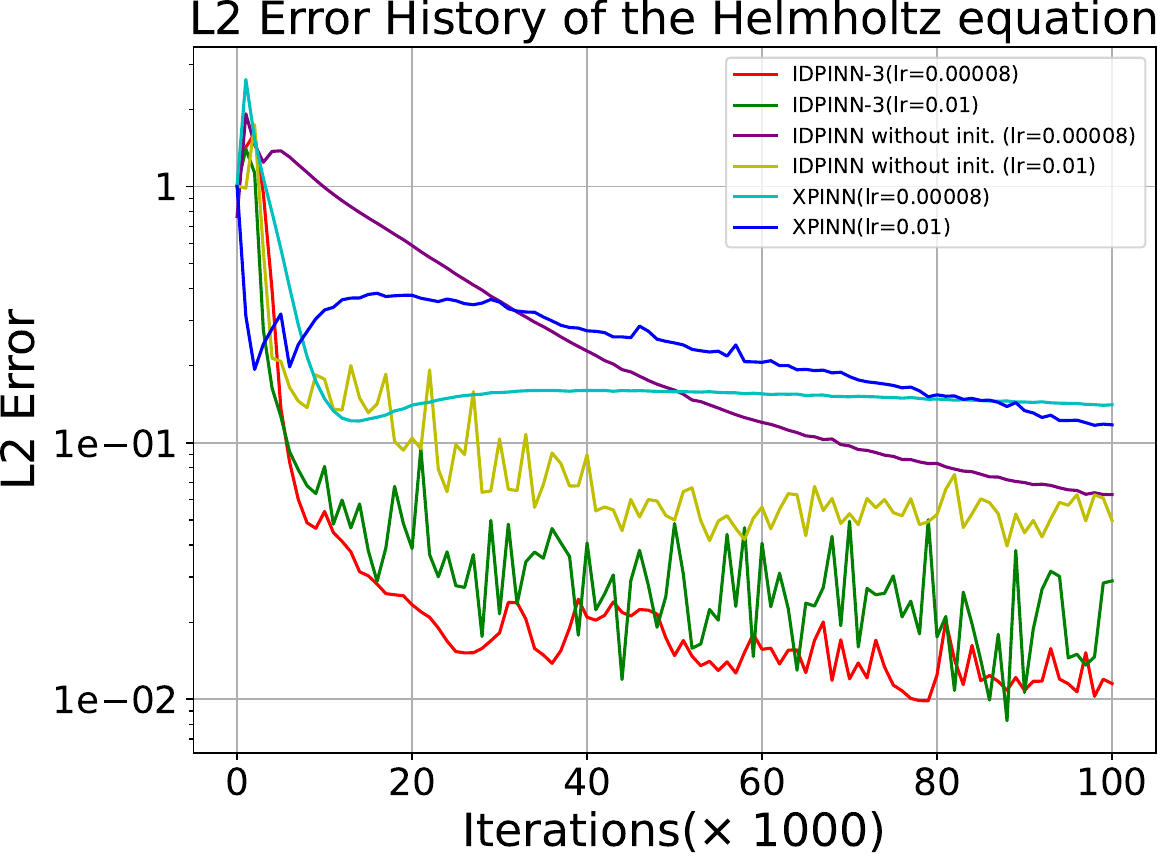}  
   \end{minipage}
    \caption{The L2 error history in 100,000 iterations for the Helmholtz equation for XPINN, IDPINN-3, and IDPINN-3 without initialization, under two different learning rates.}\label{Helm Error history}
\end{figure}

The last row in Fig.~\ref{Helm result} shows the performance of the IDPINN without the initialization. The overall performance is better than XPINN while there are still a small error occurring around the interface, comparing with IDPINN-3. Further, the L2 error history for IDPINN-3 without initialization, using a lower learning rate ($8\times 10^{-5}$), is presented in Fig.~\ref{Helm Error history} (purple curve). We realized that the L2 error decreased slowly in the initial iterations. Consequently, we increase the learning rate to $1\times 10^{-2}$ (yellow curve). Notably, IDPINN-3 outperforms other models under both low ($8\times 10^{-5}$) and high ($1\times 10^{-2}$) learning rates. Furthermore, the performance of IDPINN-3 remains relatively better and more stable at a lower learning rate. 

\begin{figure}[!htb]
    \centering
    \begin{subfigure}{0.31\textwidth}
        \centering
        \includegraphics[width=.95\linewidth]{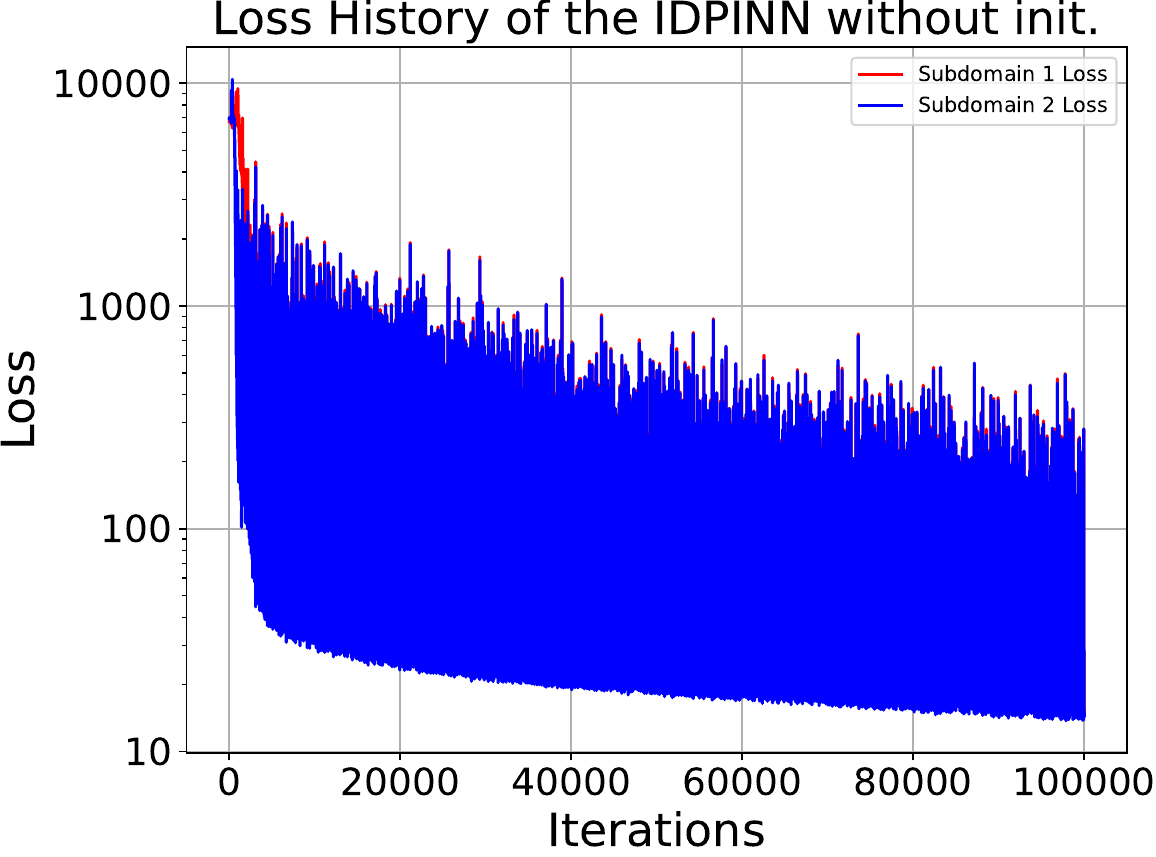}  
    \end{subfigure}
    \begin{subfigure}{0.31\textwidth}
        \centering
        \includegraphics[width=.95\linewidth]{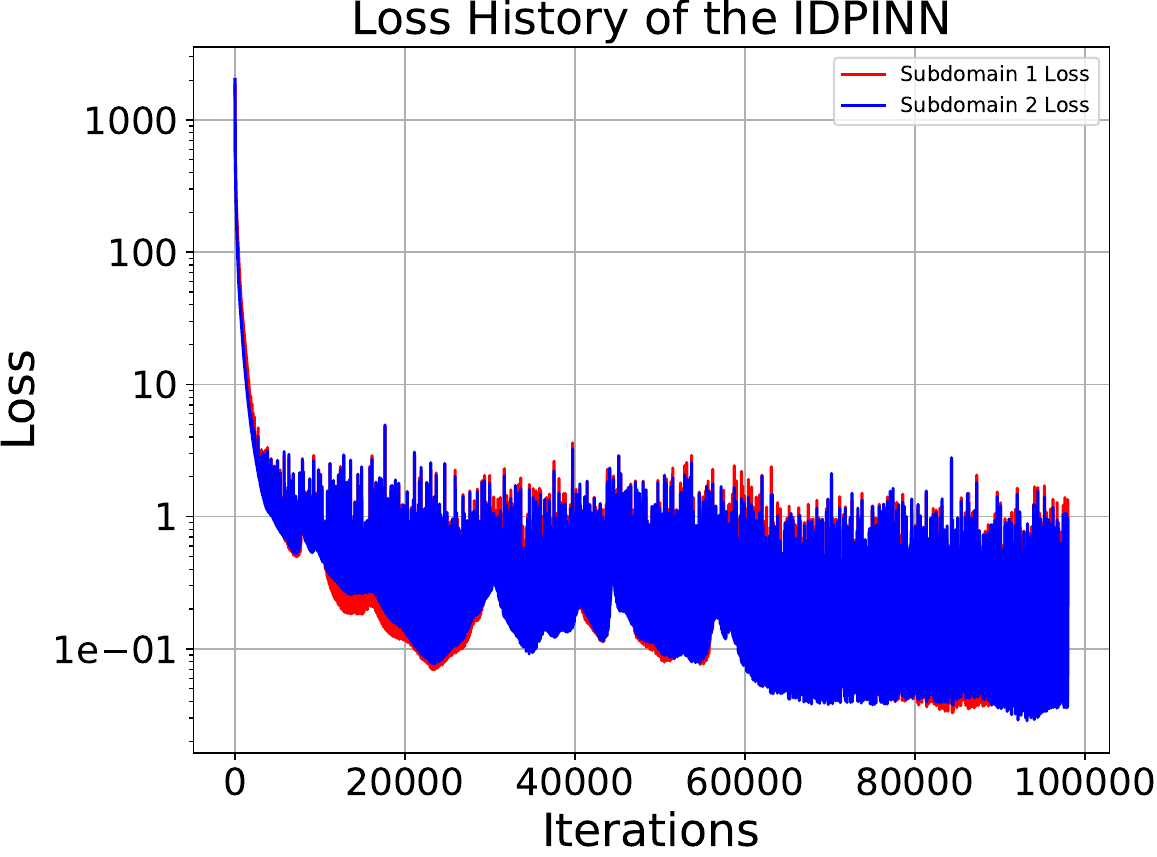}    
    \end{subfigure}
    \begin{subfigure}{0.31\textwidth}
        \centering
        \includegraphics[width=.95\linewidth]{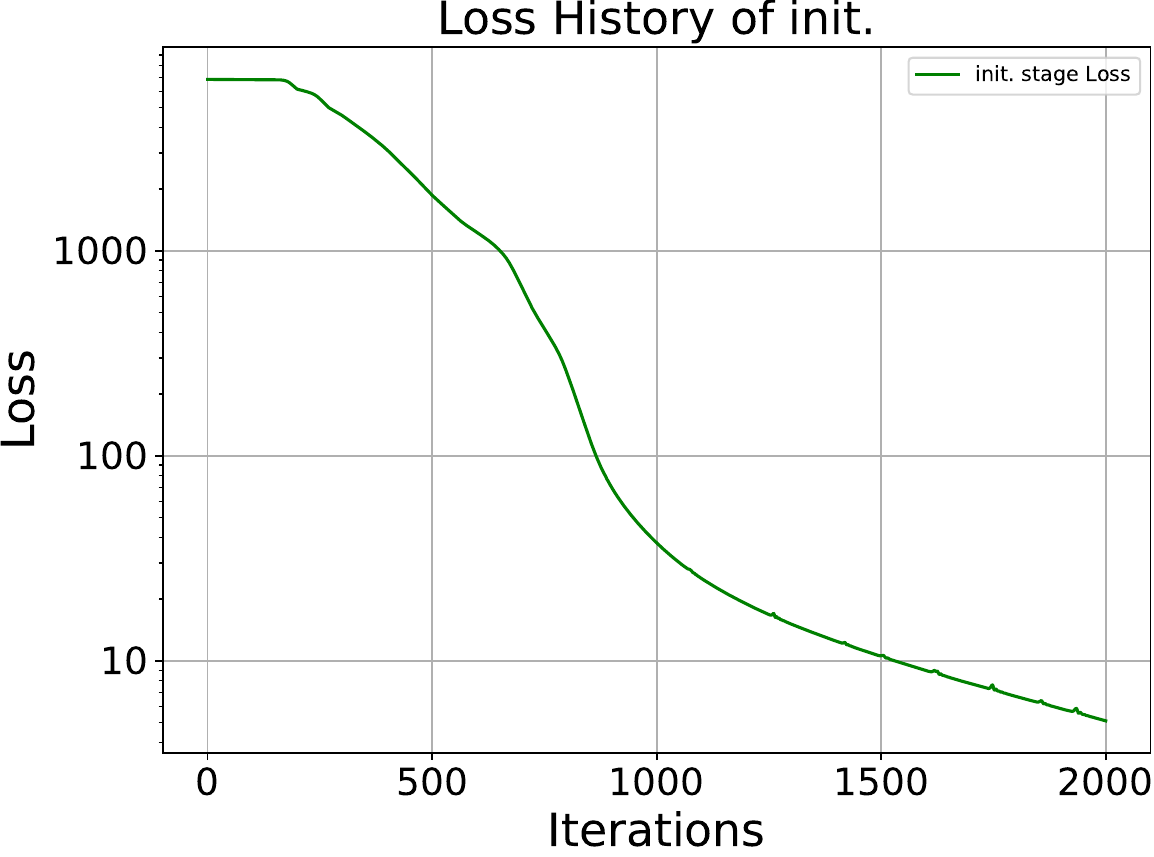}  
    \end{subfigure}
    \caption{First: Loss history for IDPINN without initialization. Second: Loss history for IDPINN-3 after the initialization. Third: Loss history of the IDPINN-init stage.}\label{loss history}
\end{figure}

We also compare the training loss history over two subdomains for IDPINN-3 without and with initialization under the higher learning rate ($1\times 10^{-2}$) in Fig.~\ref{loss history}. With initialization, the initial training loss drops from around $1\times 10^4$ to $1\times 10^3$, and the training loss can decrease to below $1\times 10^{-1}$. However, without initialization, the training loss struggles to reach around 20. This highlights the importance of initialization. Understanding its importance prompts the question: how many iterations are sufficient for the initialization stage? Too many iterations may prolong the training time unnecessarily, while too few iterations may not adequately prepare the model. Achieving the right balance is crucial for efficient training. The next subsection will examine this. 

\subsubsection{Different number of iterations in the IDPINN-init stage}
\begin{figure}[!htb]
 \centering
   \begin{minipage}{0.555\textwidth}
     \centering
     \includegraphics[width=.95\linewidth]{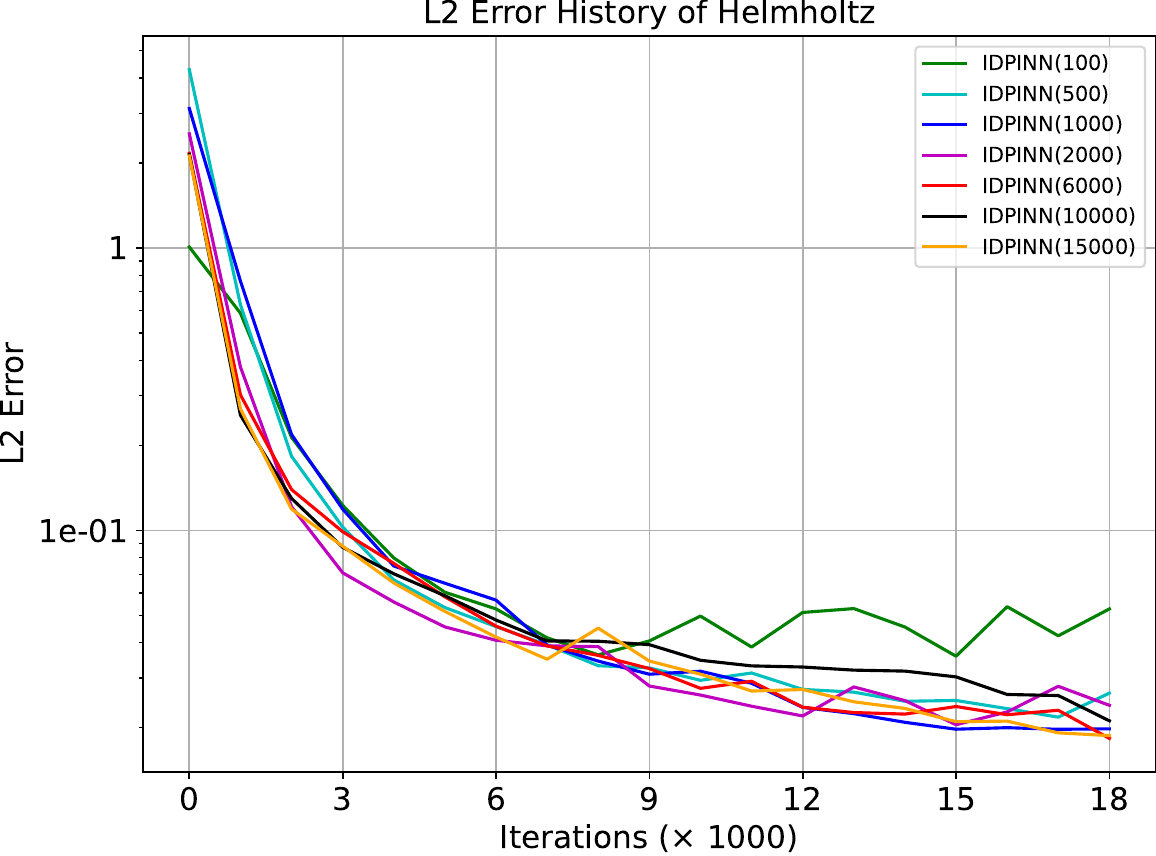}    
   \end{minipage}
    \caption{The L2 error history for IDPINN with different numbers of iterations in the IDPINN-init stage.  Only 100 iterations (green curve) do not produce a satisfactory initialization; when the number of iterations exceeds 500, the performance becomes very similar.}\label{HIST18K}
\end{figure}

The previous subsection demonstrates the importance of initialization, while this subsection focuses on determining the number of iterations in the IDPINN-init stage. Since the IDPINN-init stage initializes the parameter for the IDPINN-main part, finding a good initialization is crucial. We vary the number of iterations in the IDPINN-init stage and compare their performance in the IDPINN-main part in Fig.~\ref{HIST18K}. This figure shows that 100 iterations do not produce a satisfactory initialization, and when the number of iterations exceeds 500, the performance becomes very similar. This suggests that we do not need too many iterations in the initialization stage. Therefore, we only require a certain number of iterations in the IDPINN-init stage to achieve satisfactory performance. 

\begin{figure}[!htb]
 \centering
   \begin{minipage}{0.45\textwidth}
     \centering
     \includegraphics[width=.95\linewidth]{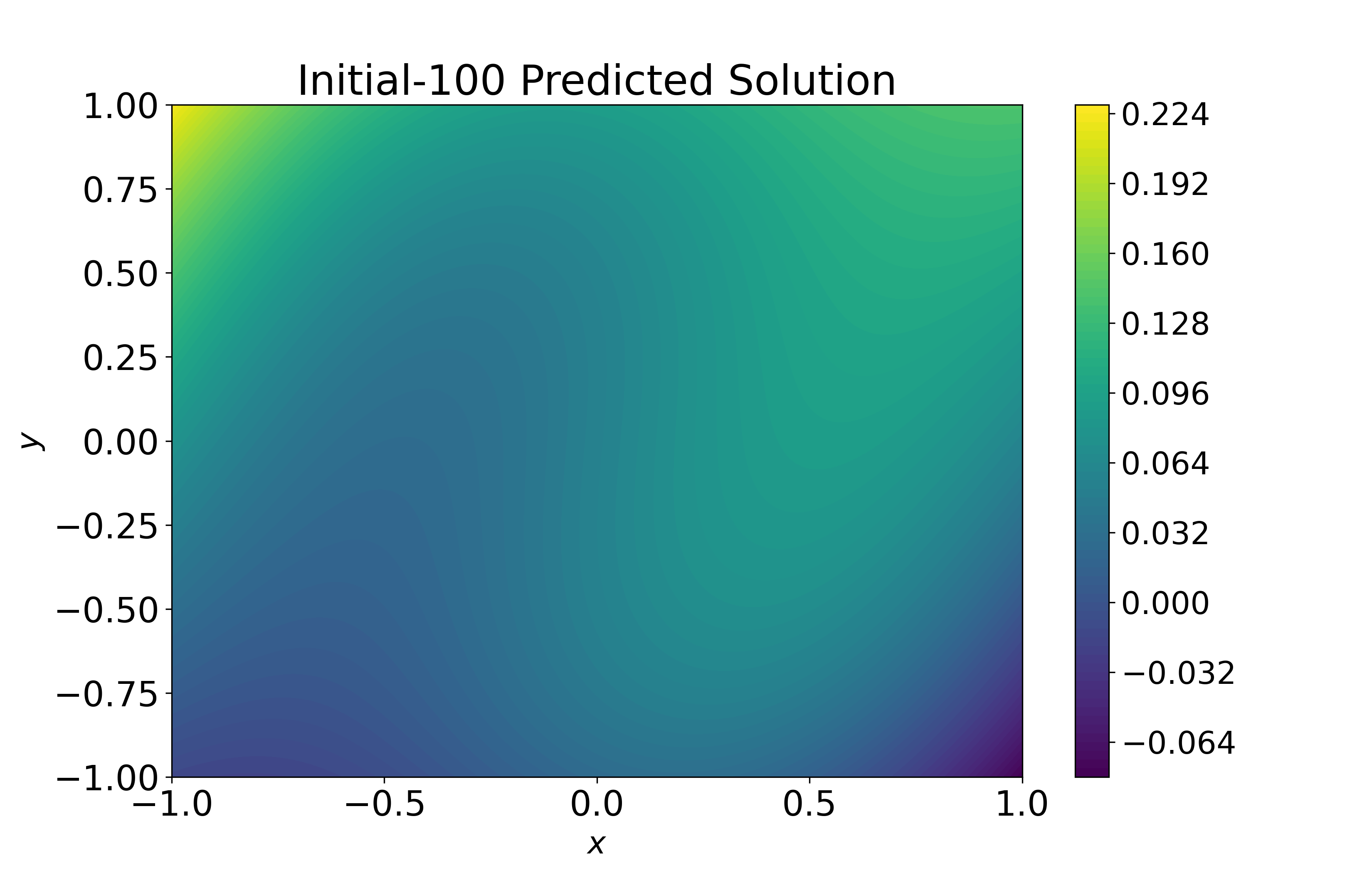}    
   \end{minipage}
      \begin{minipage}{0.45\textwidth}
     \centering
     \includegraphics[width=.95\linewidth]{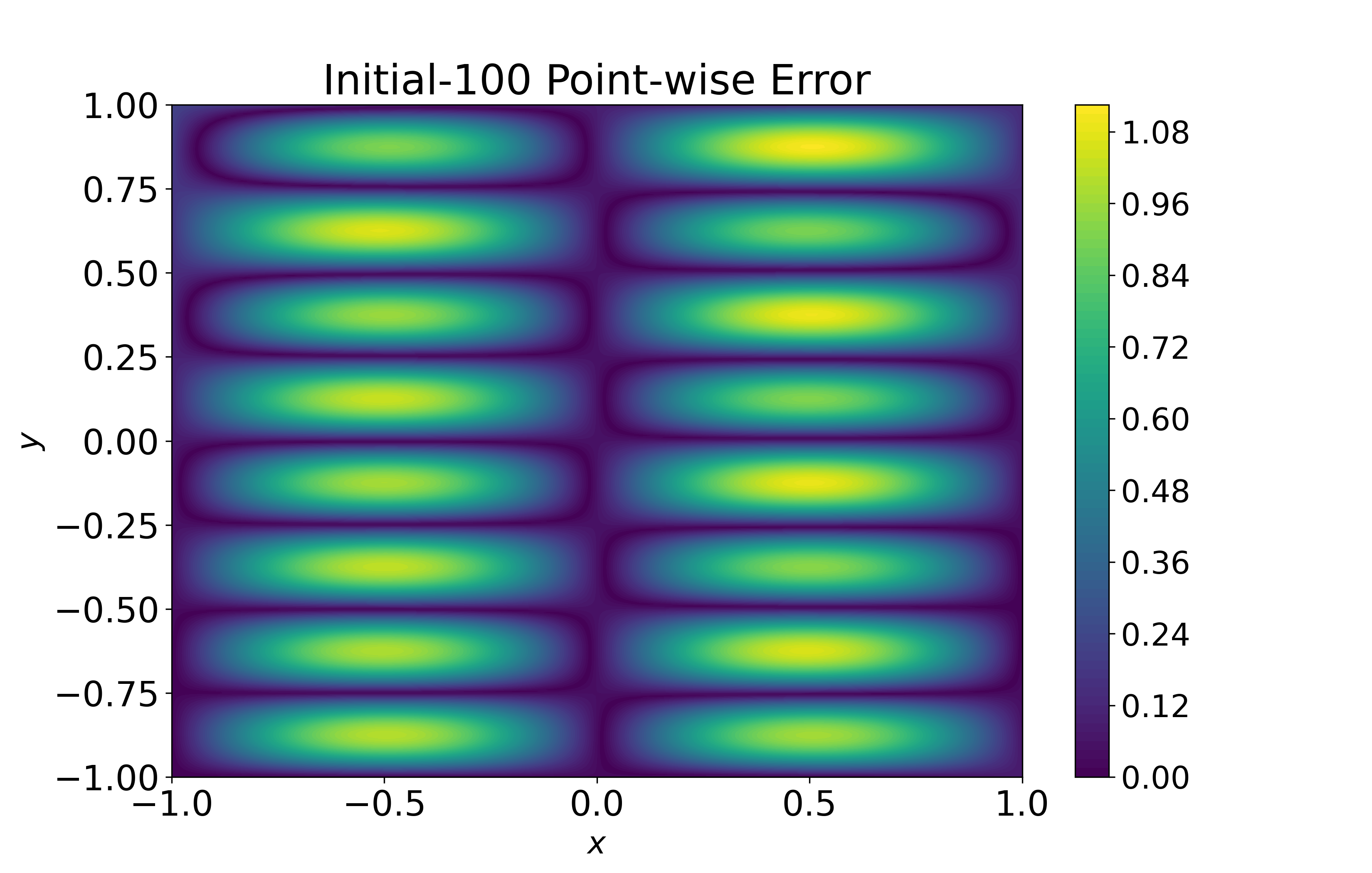}    
   \end{minipage}
      \begin{minipage}{0.45\textwidth}
     \centering
     \includegraphics[width=.95\linewidth]{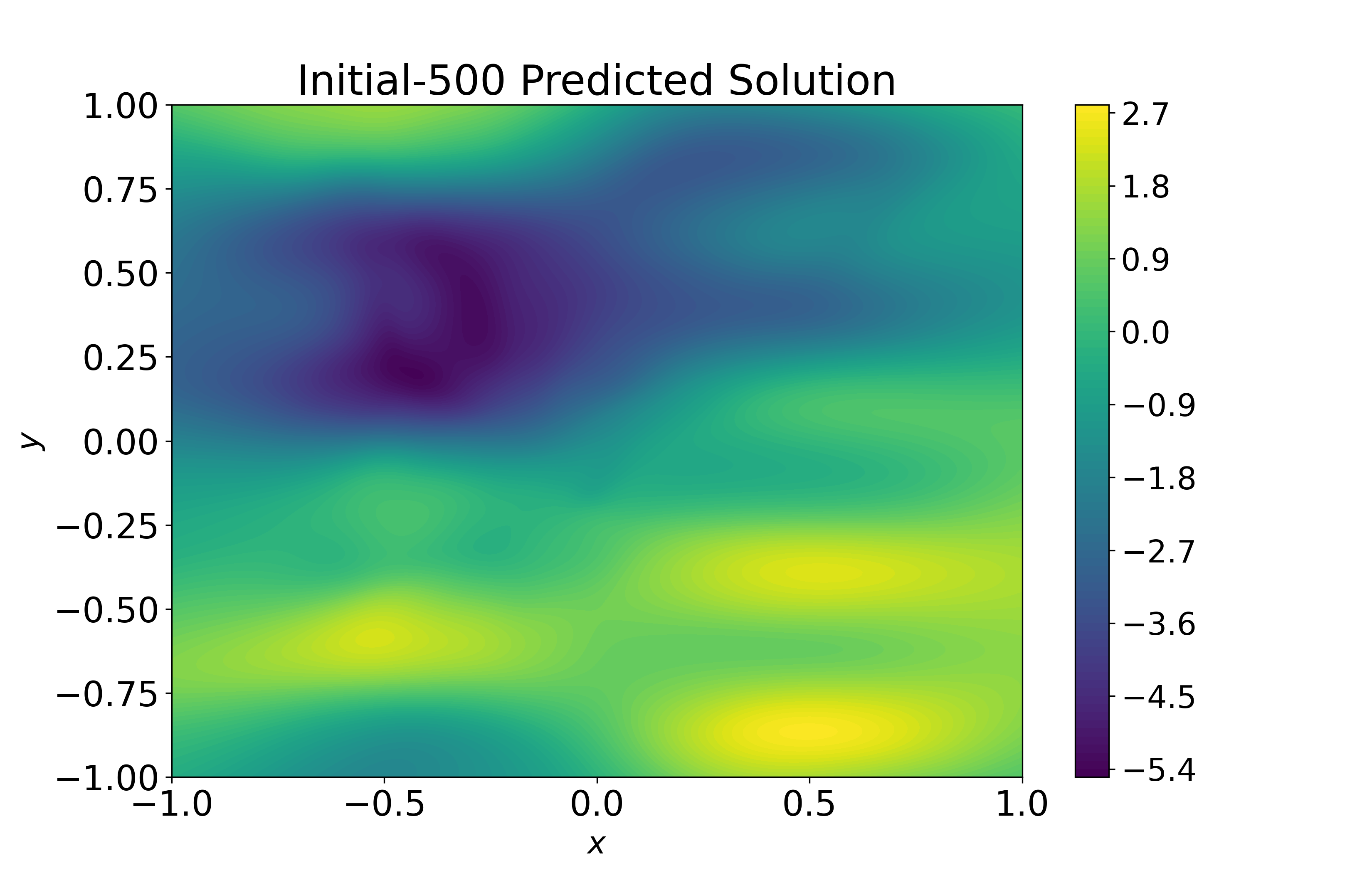}    
   \end{minipage}
      \begin{minipage}{0.45\textwidth}
     \centering
     \includegraphics[width=.95\linewidth]{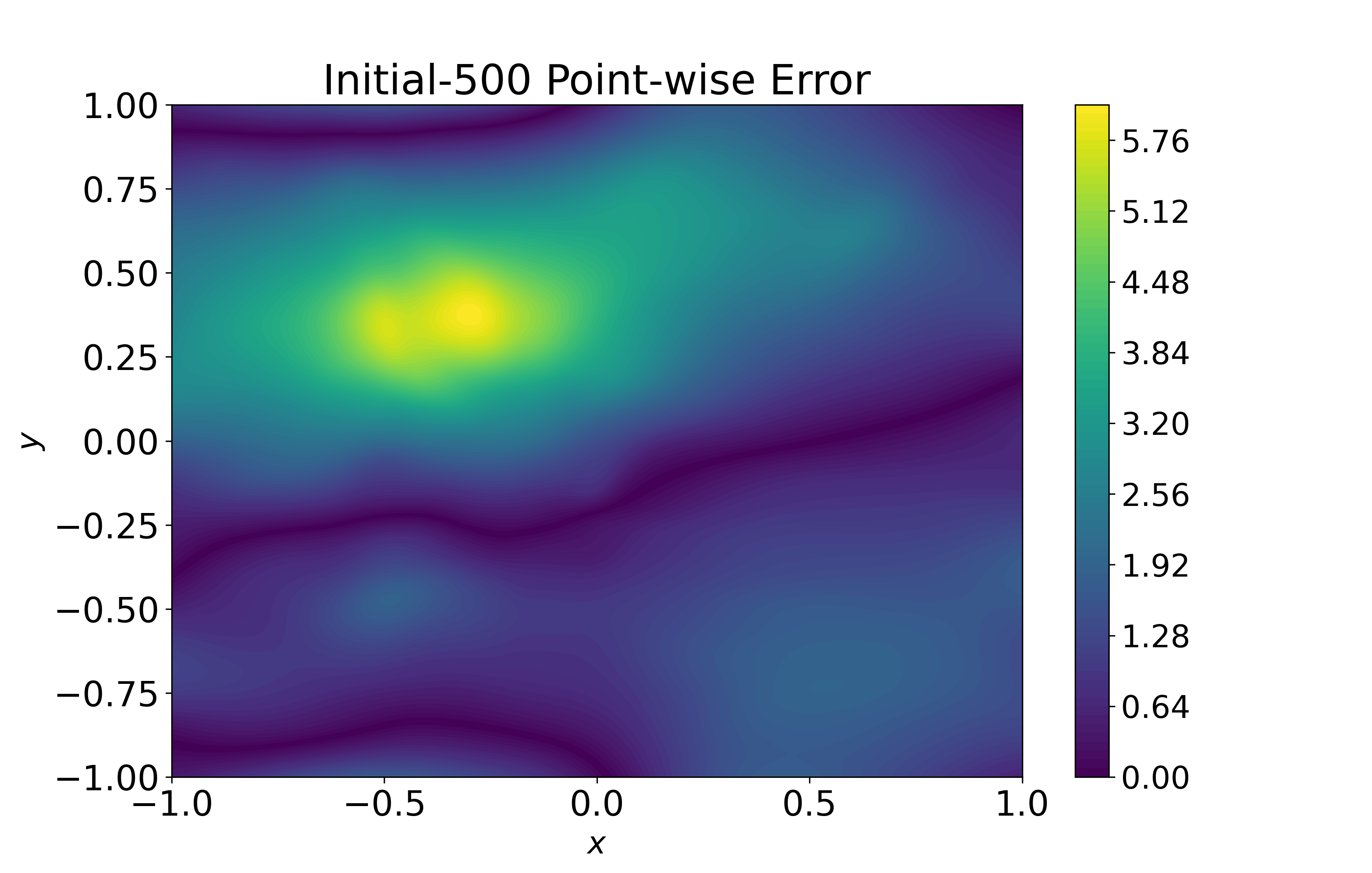}    
   \end{minipage}
      \begin{minipage}{0.45\textwidth}
     \centering
     \includegraphics[width=.95\linewidth]{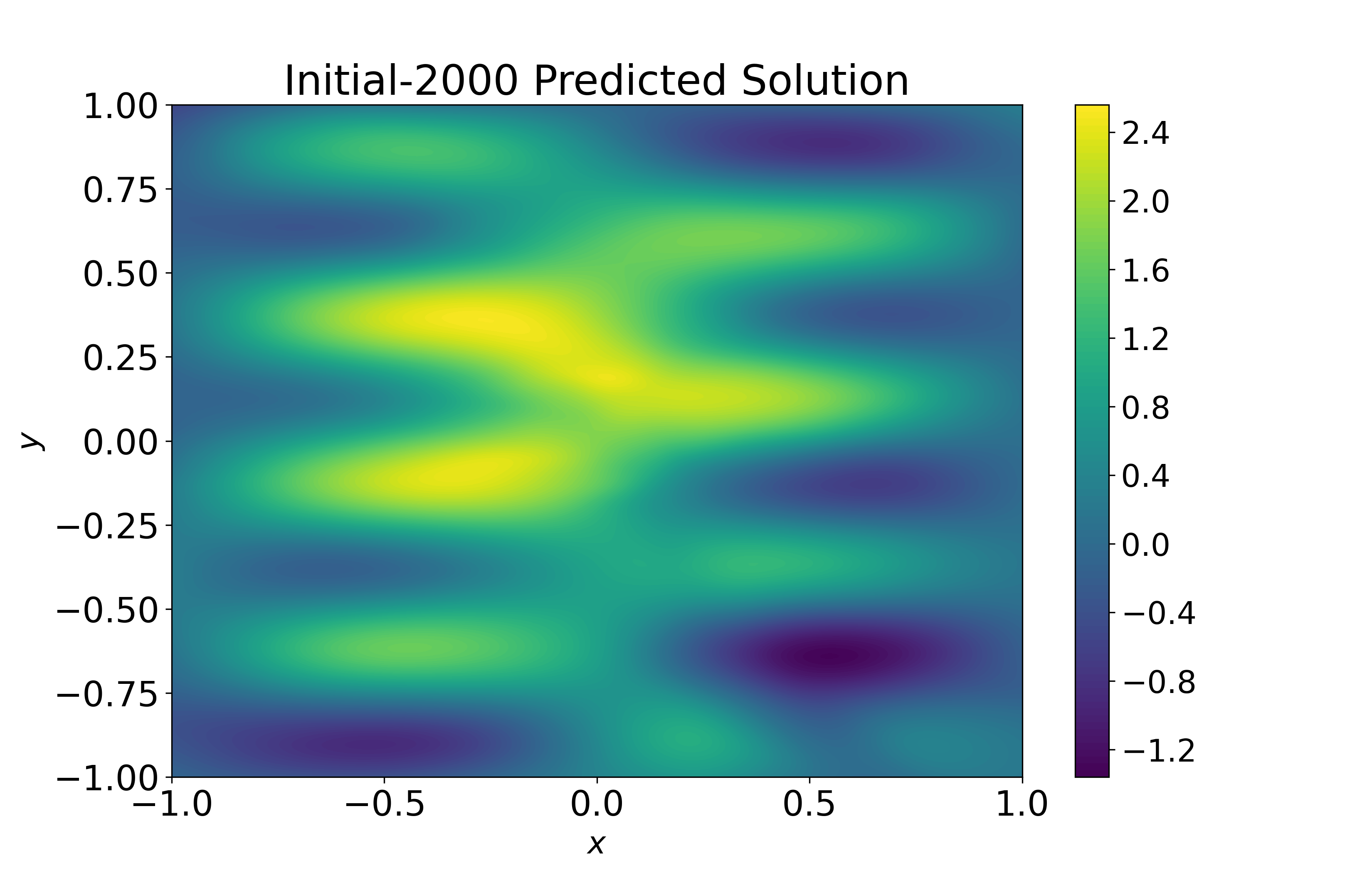}    
   \end{minipage}
      \begin{minipage}{0.45\textwidth}
     \centering
     \includegraphics[width=.95\linewidth]{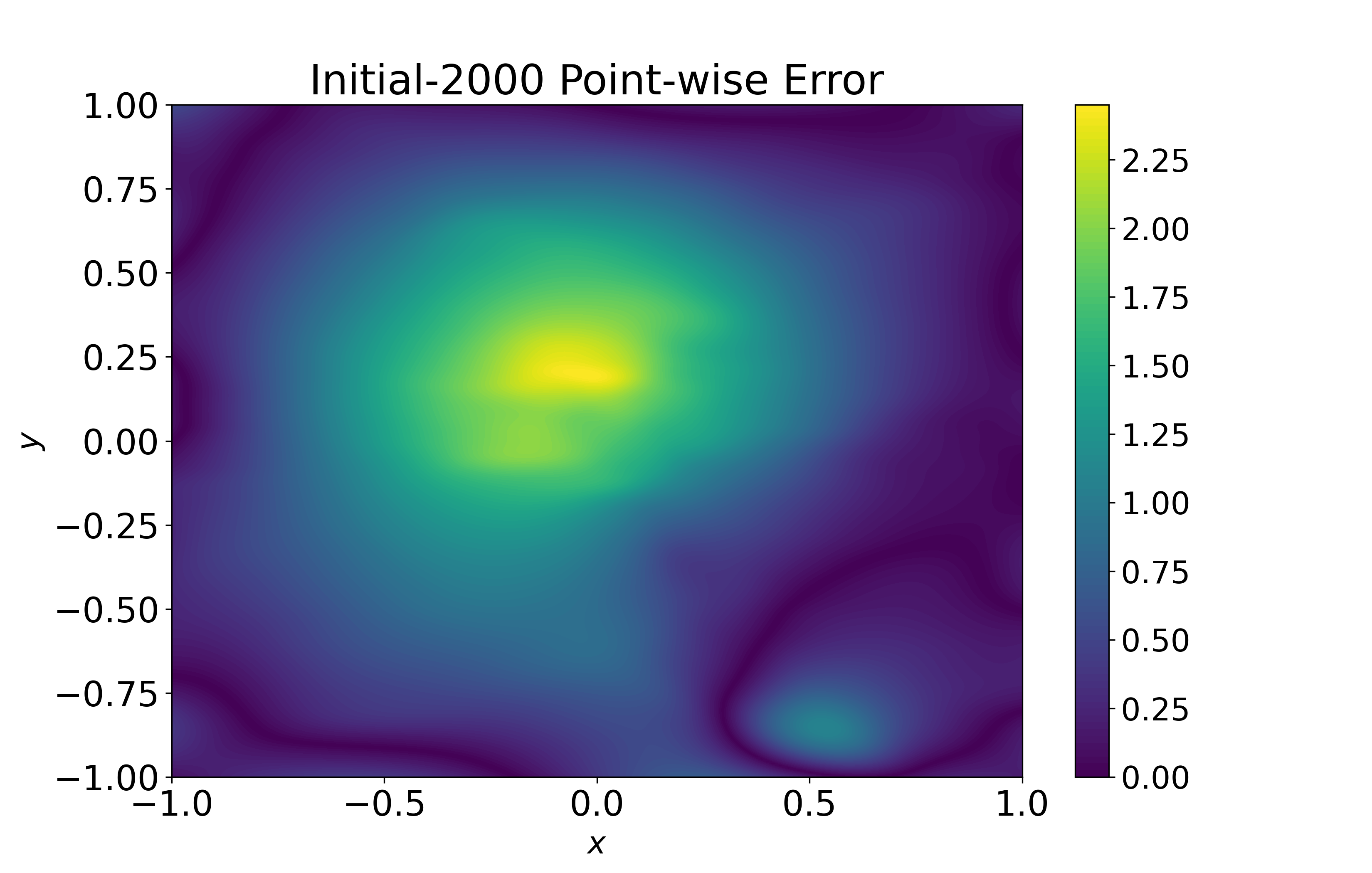}    
   \end{minipage}
    \caption{The prediction and the point-wise error for the IDPINN-init, when the number of training steps is different. (100 iterations for the first row, 500 for the second, and 2000 for the last row.)}\label{init}
\end{figure}

In addition, we examine the solution constructed by initialization and its point-wise error in Fig.~\ref{init}. After 100 iterations, the solution remains very close to zero. However, after 500 iterations, discernible patterns begin to emerge. Comparing the solutions for 500 and 2000 iterations, we observe that the range of the function values becomes closer to the true range $[-1,1]$ as the number of iterations increases. The larger function value also justifies why the initialization with 500 iterations has the highest initial L2 error in Fig.~\ref{HIST18K}.

This and the preceding subsections demonstrate that a certain number of iterations in the initialization stage is adequate to enhance the performance of IDPINN. Without precise knowledge of the exact number of iterations required, opting for a relatively large number is advisable, considering that the initialization step is fast.

\subsection{2D Poisson Equation}
This experiment uses IDPINN-3, which includes smoothness terms \eqref{9} and \eqref{10}, to solve the 2D Poisson equation under the same domain setup as~\cite{JaptapXelta2020}:
\begin{align*}
    \Delta u = f(x,y) \quad (x,y)\in\Omega\subset\mathbb{R}^2.
\end{align*}
The boundary is $\partial\Omega:=\big\{(0.02+r(\theta)\cos\theta,r(\theta)\sin\theta):\theta \in [0,2\pi)\big\}$, where 
\begin{align*}
r(\theta) = 1.5+0.14 \sin(4\theta)+0.12 \cos(6\theta)+0.09\cos(5\theta). \label{Poi boundary}
\end{align*}
The two interface curves are defined by:
$\Gamma_1:=\big\{(-0.4+r_1(\theta)\cos\theta,-0.4+r_1(\theta)\sin\theta):\theta \in [0,2\pi)\big\}$ and
$\Gamma_2:=\big\{(0.5+r_2(\theta)\cos\theta,0.6+r_2(\theta)\sin\theta):\theta \in [0,2\pi)\big\}$, where 
\begin{align*}
r_1(\theta) = & 0.50+0.18\sin(3\theta)+0.08\cos(2\theta)+0.2\cos(5\theta),\\
r_2(\theta) = & 0.34+0.04\sin(5\theta)+0.18 \cos(3\theta)+0.1\cos(6\theta).
\end{align*}
Assume the true solution is $u(x, y) = e^x + e^y$, and we have $f(x,y)=e^x+e^y$. 

\begin{figure}[!htb]
     \centering
     \begin{minipage}{0.45\textwidth}
     \centering
     \includegraphics[width=.95\linewidth]{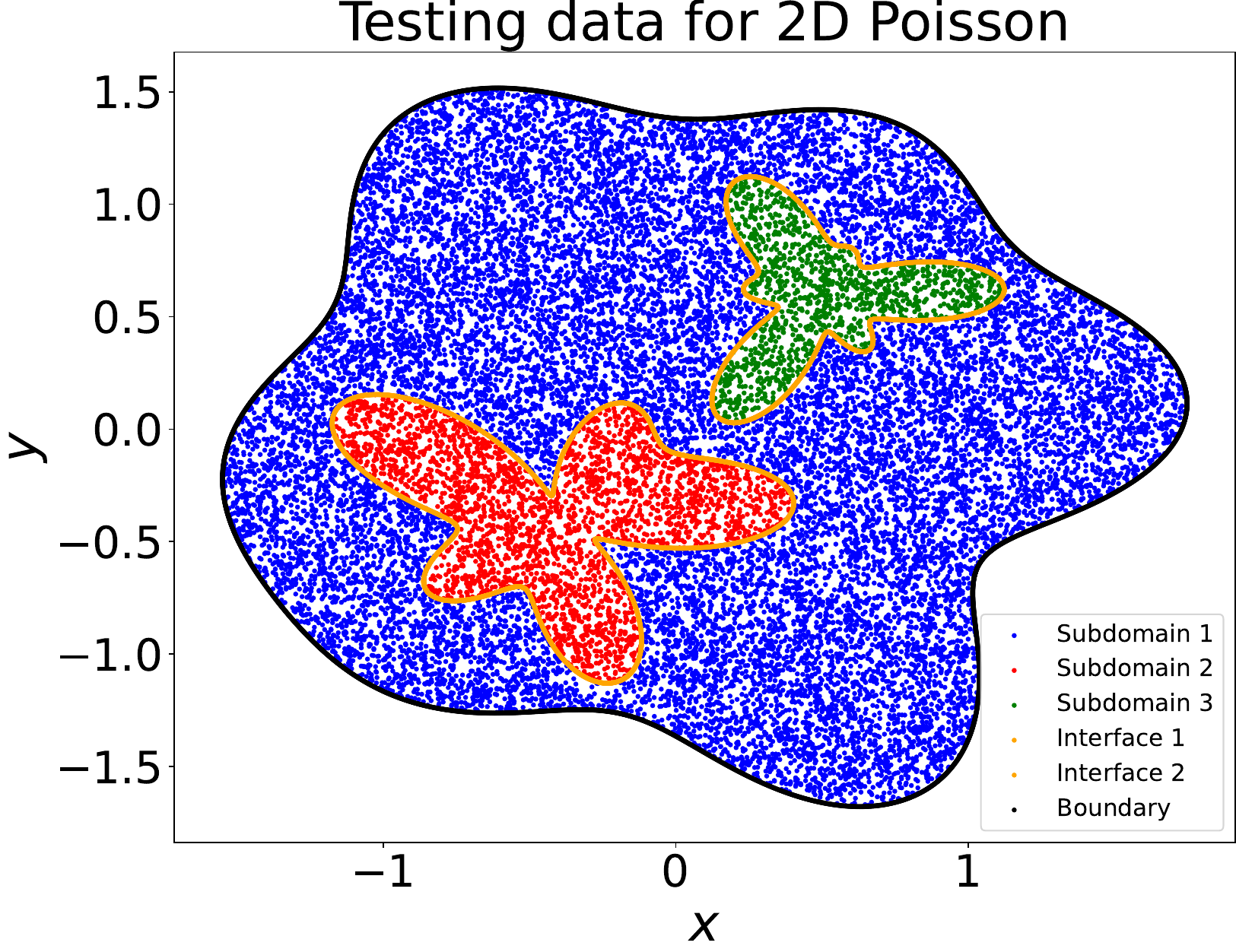}  
     \end{minipage}  
    \caption{Training points for the 2D Poisson equation. There are three subdomains. The training points in each subdomain are represented by blue (subdomain 1), red (subdomain 2), and green (subdomain 3). Points on the interface and boundary are represented by orange and black, respectively.}\label{Poisson data}
\end{figure}

The data points shared with~\cite{JaptapXelta2020} consist of 18211, 2855, and 1291 residual points in subdomains 1, 2, and 3, respectively. Additionally, there are 6284 points each for both interfaces and the boundary. All these data points are used to compute the L2 error. The training points, randomly collected from these points, are depicted in Fig.~\ref{Poisson data}, whose summary information is in Table~\ref{Table 2}. In summary, the IDPINN-main stage uses 1300 residual points, 100 interface points, and 100 boundary points, from which the points in the IDPINN-init stage are selected. 

\begin{table}[!h]
    \centering
    \caption{Number of training points for the 2D Poisson equation. The training points in the IDPINN-init stage are chosen from those in the IDPINN-main stage.}
    \begin{tabular}{|l|l|r|c|r|c|r|}
    \hline
    Stage & Region &  Residual  &  Interface & Boundary & Initial  & Iterations\\
    \hline
    IDPINN-init & all-region & 800 & NA & 50 &  \multirow{4}{*}{NA} &  500\\
    \cline{1-5}    \cline{7-7}
    \multirow{3}{*}{\makecell{IDPINN-main\\XPINN}}  & sub-domain 1 (blue) & 1000 & \multirow{3}{*}{100 (50 per  interface)} & 100 & &  \multirow{3}{*}{3500}\\
    \cline{2-3}
    \cline{5-5}
    & sub-domain 2 (red) & 180 & & \multirow{2}{*}{0} &  & \\
    \cline{2-3}
    & sub-domain 3 (green) & 120 & & &  & \\
    \hline
    \end{tabular}
    \label{Table 2}
\end{table}

For XPINN, we initially adopt the layer structure as defined in~\cite{JaptapXelta2020}, which is [2, 20, 20, 20, 20, 1] for sub-domain 1, [2, 25, 25, 25, 1] for sub-domain 2, and [2, 30, 30, 1] for sub-domain 3. However, utilizing different networks for distinct sub-domains poses challenges in initialization for IDPINN. Thus, we conduct further experiments with XPINN by employing uniform layer configurations for all sub-domains, set as [2, 20, 20, 20, 20, 1]. For simplicity, we denote this model as XPINN-U20. The weighted parameters for the loss functions in both models are set as $\lambda_1 = 1$, $\lambda_2 = 20$, $\lambda_3 = 0$, $\lambda_{residual} = \lambda_{avg} =20$. For IDPINN, the network structure is defined as [2, 20, 20, 20, 20, 1]. The weighted parameters assigned to the loss terms are $\lambda_1 = 1$, $\lambda_2 = 20$, $\lambda_3 = 0$, $\lambda_4 = 2$, $\lambda_5 = 5$, and $\lambda_6 = 5$. The network structures are summarized in Table~\ref{Table network}. 

\begin{table}[!h]
  \centering
  \caption{Network structures of IDPINN and XPINN for the 2D Poisson equation.}
  \begin{tabular}{|l|l|r|}
    \hline
    Model & Region & Network structure\\
    \hline
    IDPINNs  & all sub-domains  & [2, 20, 20, 20, 20, 1]\\
    \cline{2-2}
    \hline
    XPINN-U20  & all sub-domains & [2, 20, 20, 20, 20, 1]\\
    \cline{2-2}
    \hline
    \multirow{3}{*}{XPINN}  & sub-domain 1 (blue) & [2, 20, 20, 20, 20, 1]\\
    \cline{2-3}
    & sub-domain 2 (red) & [2, 25, 25, 25, 1]\\
    \cline{2-3}
    & sub-domain 3 (green) & [2, 30, 30, 1] \\
    \hline
    \end{tabular}
    \label{Table network}
\end{table}

\begin{figure}[!htb]
\centering
    \begin{subfigure}{0.33\textwidth}
     \centering
     \includegraphics[width=\linewidth, height=0.75\textwidth]{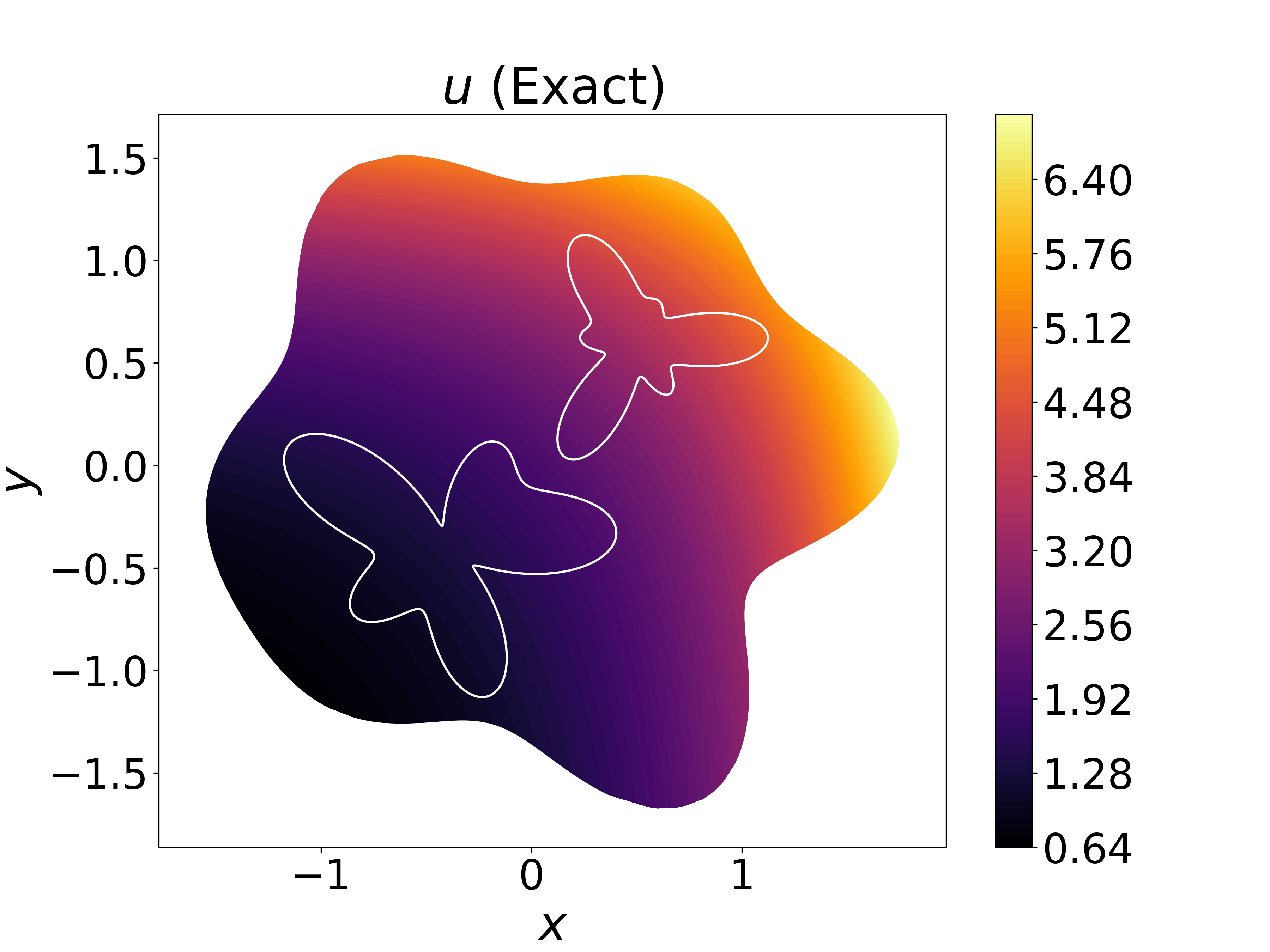}  
     \caption*{\textbf{A} (Exact solution)}
   \end{subfigure}   
   \begin{subfigure}{0.33\textwidth}
     \centering
     \includegraphics[width=\linewidth, height=0.75\textwidth]{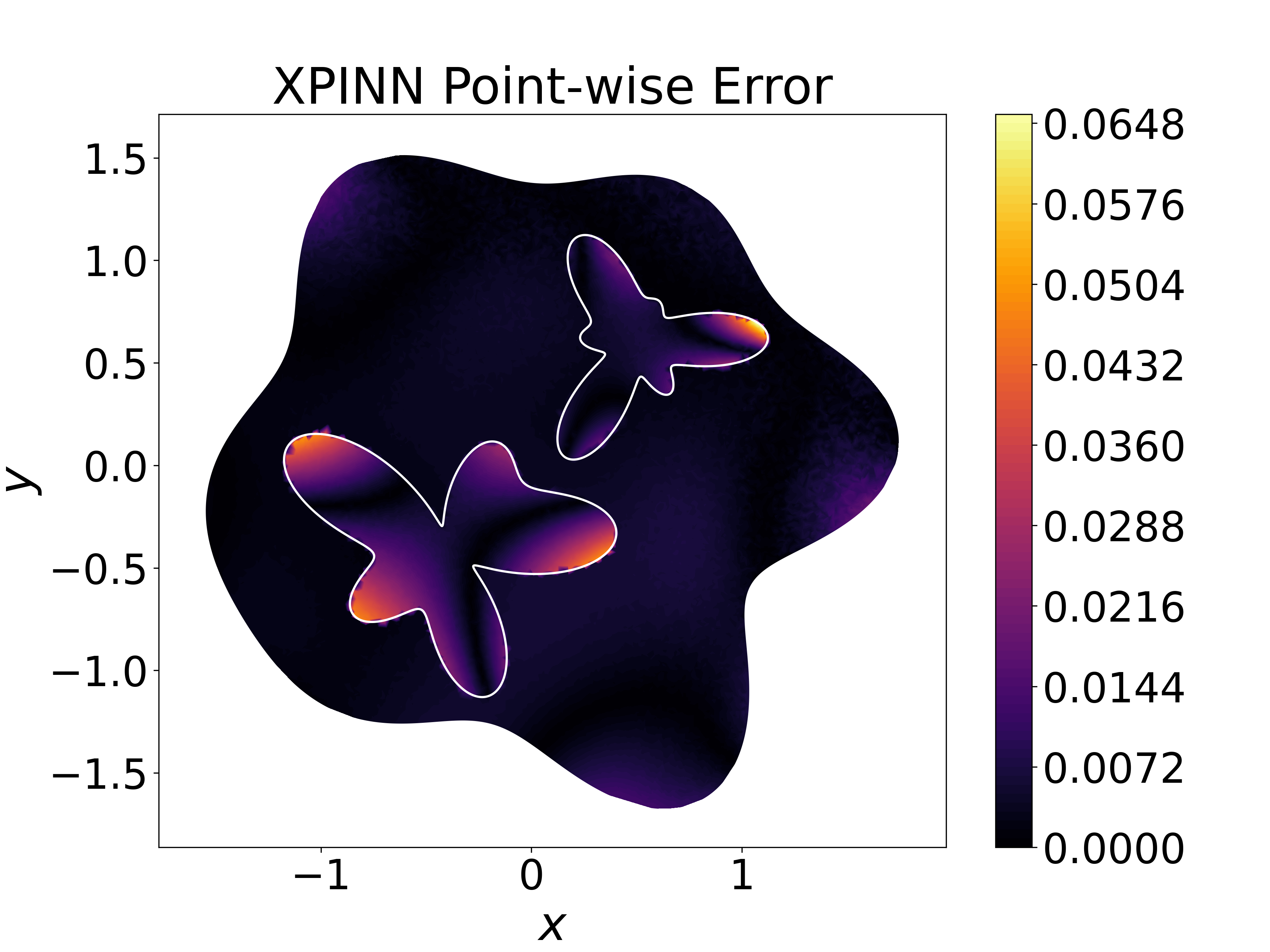} 
     \caption*{\textbf{B} (L2 error: $2.47\times 10^{-3}$)}
   \end{subfigure} 
   \begin{subfigure}{0.33\textwidth}
     \centering
     \includegraphics[width=\linewidth, height=0.75\textwidth]{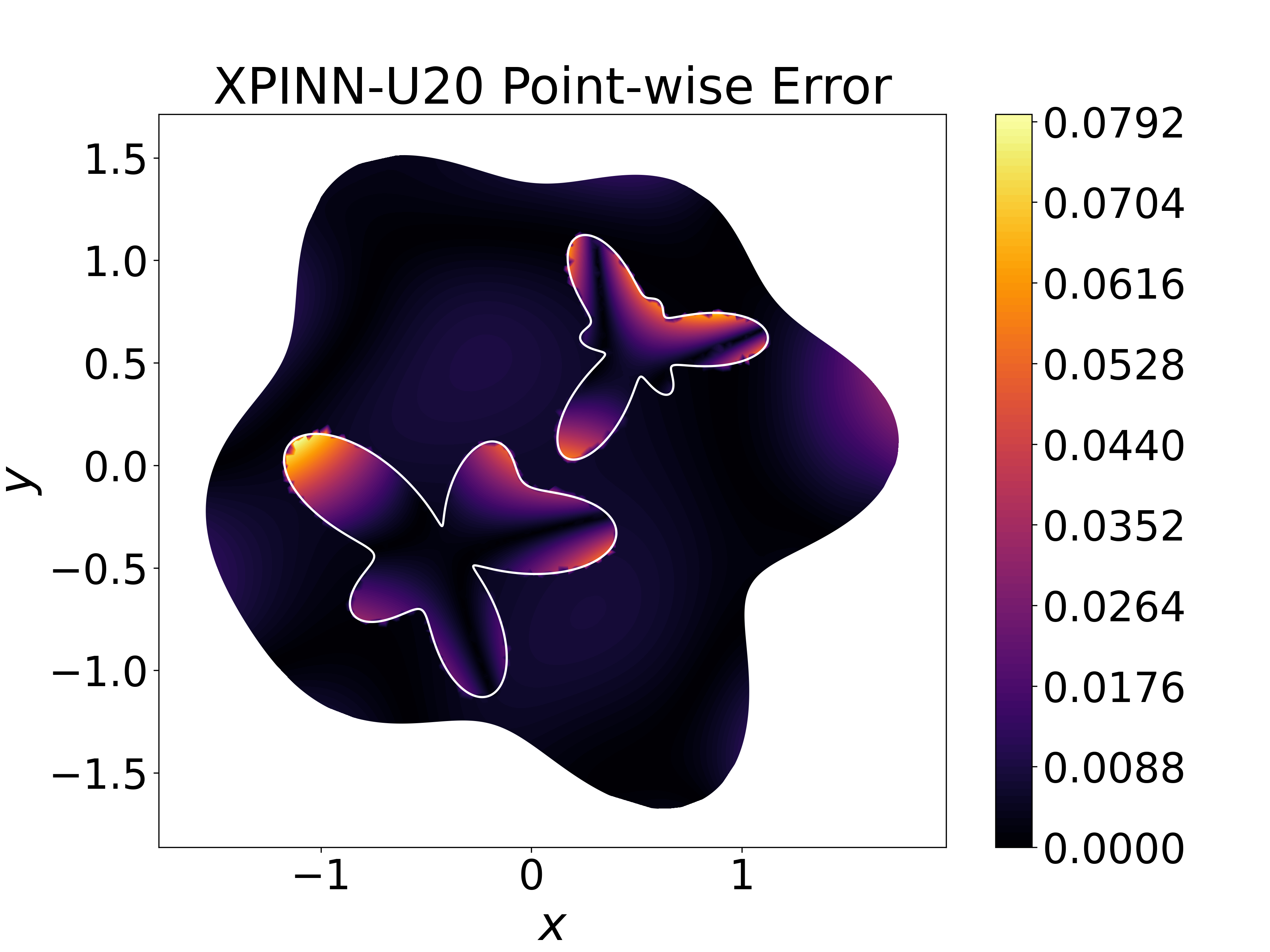} 
     \caption*{\textbf{C} (L2 error: $3.53\times 10^{-3}$)}
   \end{subfigure} 
   
   
    \begin{subfigure}{0.33\textwidth}
     \centering
     \includegraphics[width=\linewidth, height=0.75\textwidth]{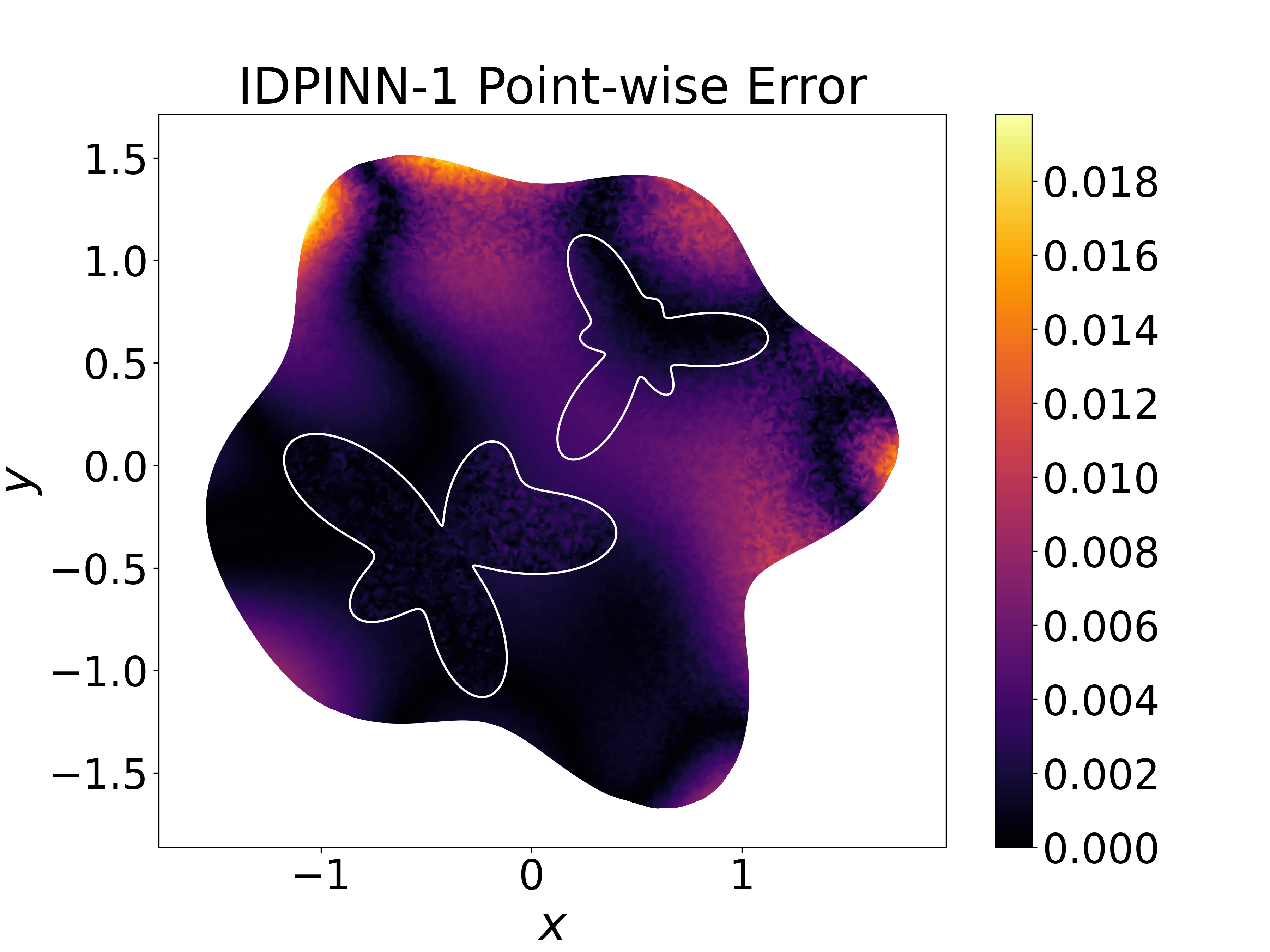}  
     \caption*{\textbf{D} (L2 error: $1.07\times 10^{-3}$)}
   \end{subfigure}
      \begin{subfigure}{0.33\textwidth}
     \centering
     \includegraphics[width=\linewidth, height=0.75\textwidth]{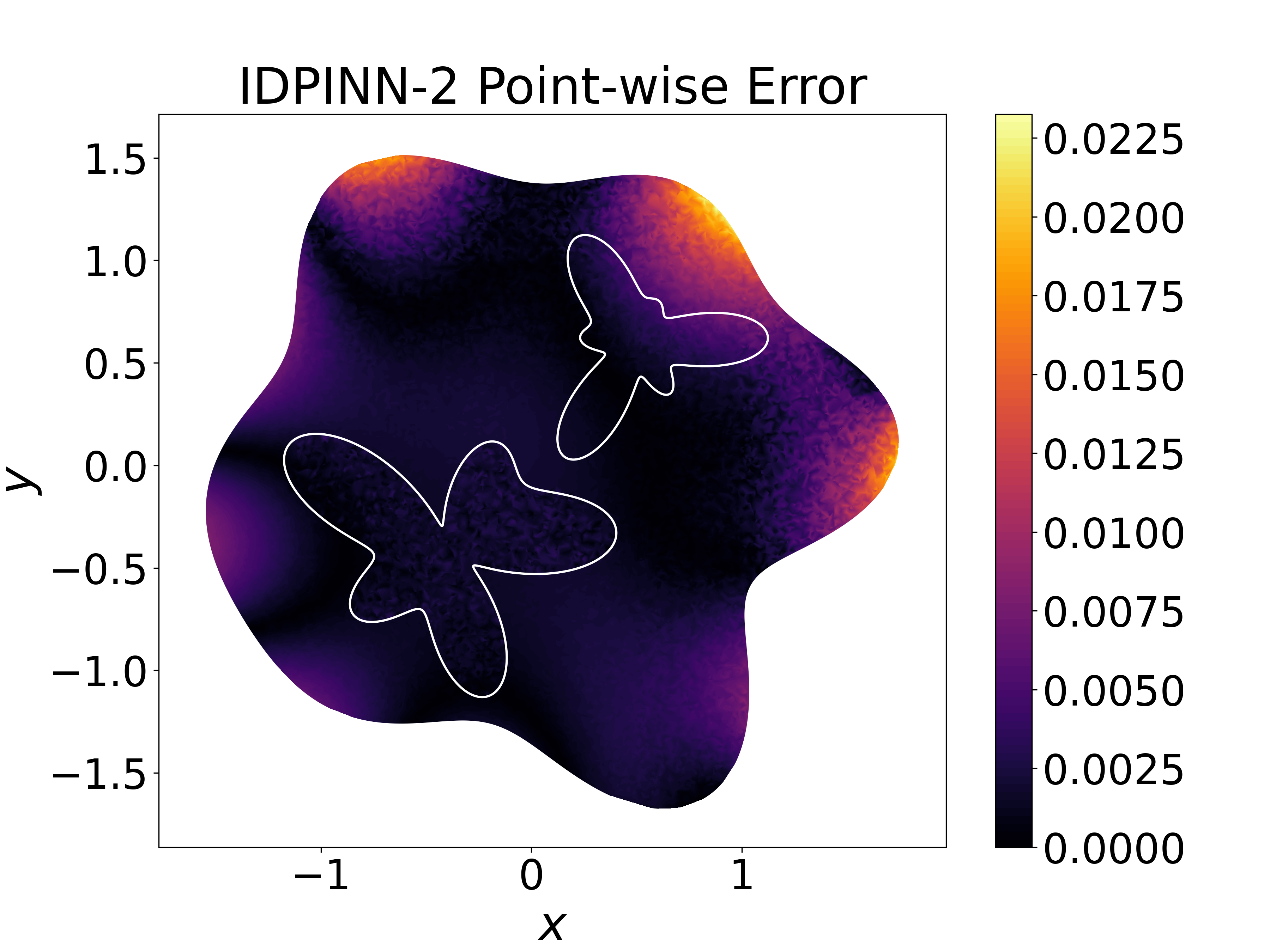}  
     \caption*{\textbf{E} (L2 error: $9.72\times 10^{-4}$)}
     \end{subfigure}
    \begin{subfigure}{0.33\textwidth}
     \centering
     \includegraphics[width=\linewidth, height=0.75\textwidth]{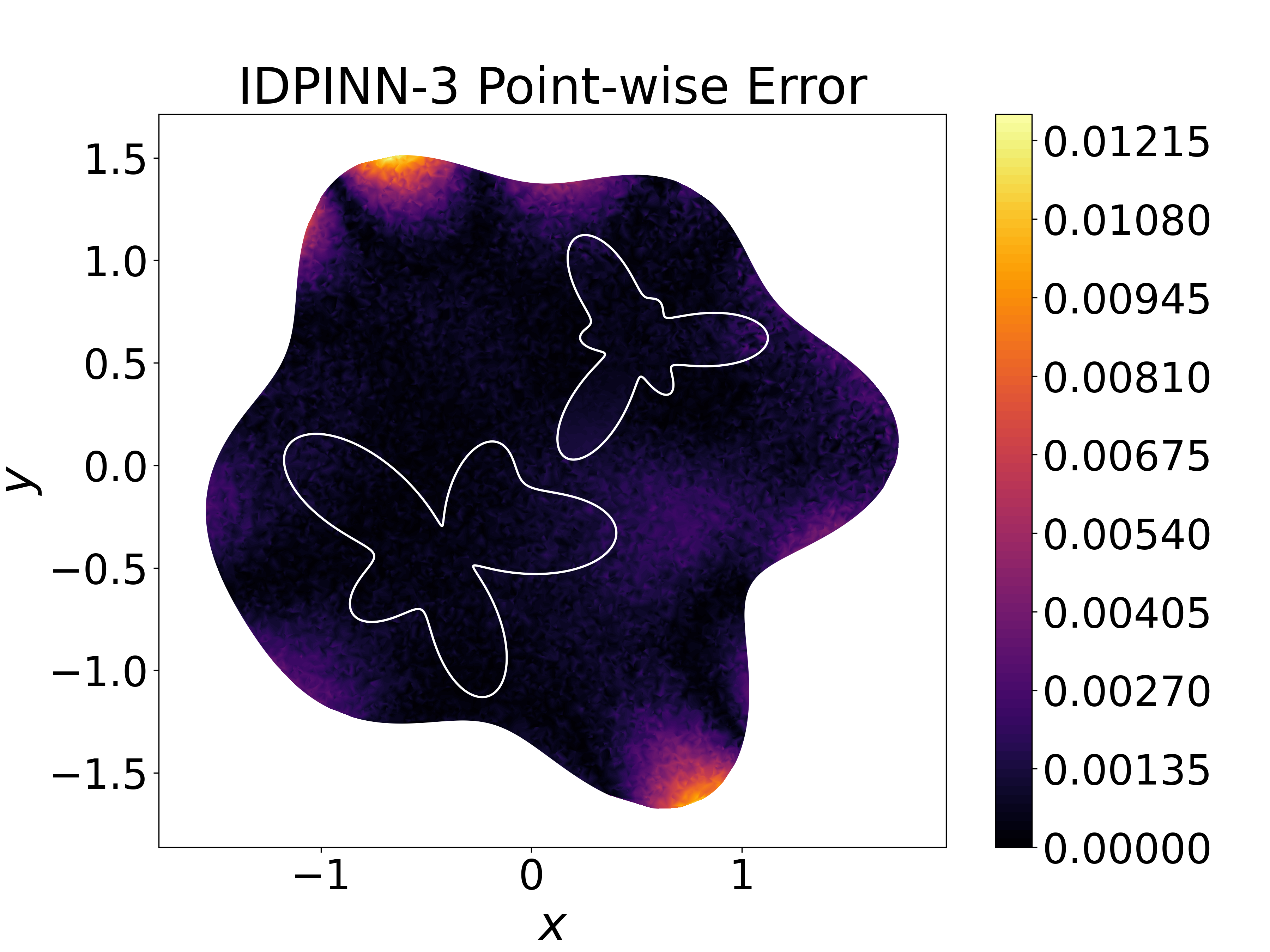}  
     \caption*{\textbf{F} (L2 error: $6.48\times 10^{-4}$)}
   \end{subfigure}
   \caption{Exact solution of the 2D Poisson equation and the point-wise relative error for different methods after 4k XPINN or 4k IDPINN-main iterations. A: Exact solution. B and C: Relative error of XPINN and XPINN-U20. D-F: Relative error of IDPINN-1, IDPINN-2, and IDPINN-3, respectively. The L2 errors are shown in the captions as well.}\label{Poisson result}    
\end{figure}

The performance of different methods is shown in Fig.~\ref{Poisson result}. Notably, IDPINN demonstrates better performance than XPINN, particularly at the interface. The second row shows the benefit of incorporating both smoothness terms. Specifically, IDPINN-3 with both terms performances better than the other two models with only one smooth term. In the next subsection, we will compare the performance with different $\lambda_6$ values.

\subsubsection{Effect of choosing different $\lambda_6$}
\begin{figure}[!htb]
\centering
    \begin{minipage}{0.55\textwidth}
     \centering
     \includegraphics[width=.95\linewidth]{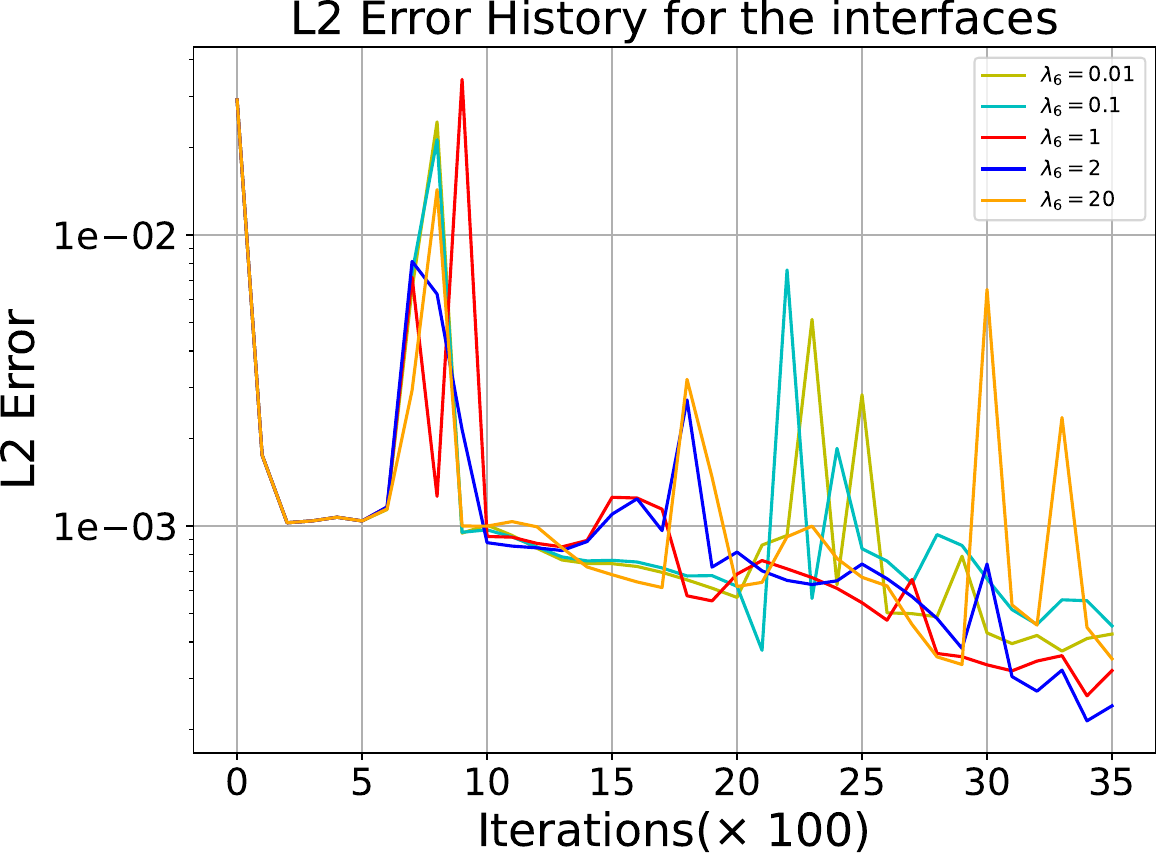}  
   \end{minipage}  
    \caption{The L2 error history of the interfaces for different $\lambda_6$ values in IDPINN-3.}\label{Interface Error history}
\end{figure}

In Fig.~\ref{Poisson result}, we set $\lambda_6 = 2$ for IDPINN-3, and while the model IDPINN-1 corresponds to $\lambda_6=0$. We observed that IDPINN-3 performs better at the interface. Therefore, in this subsection, we test the performance at the interfaces with different $\lambda_6$ values. We choose $\lambda_6\in\{0.01, 0.1, 1, 2, 20\}$ while fixing the other parameters. Fig.~\ref{Interface Error history} shows the L2 error history specifically at the interfaces. When $\lambda_6$ is set to 1 (red) or 2 (blue), the overall performance of the history curve is better than smaller or larger $\lambda_6$ values. 

\subsubsection{Effect of the layer structure}

The layer structure significantly influences the neural network's performance. In this section, we conduct additional experiments using various layer architectures (different numbers of layers and number of neurons in the layer). The L2 error results are summarized in Table~\ref{Poisson layer table}. For each structure, we train the model five times with different random seeds and select the best-performing result. While there is no obvious pattern in the results, we observe that IDPINN-3 performs better when the width of the network is set to 40 or 60. Additionally, increasing the number of hidden layers does not necessarily lead to improved prediction performance.

\begin{table}[h]
  \centering
  \renewcommand{\arraystretch}{1.3}
  \caption{The L2 error of different layer structures on the 2D Poisson equation.}
  \begin{tabular}{|c|c|c|c|c|c|}
  \hline
    \diagbox{neurons}{layers} & 2 & 3 & 4 & 6 & 8\\
    \hline
    20 & $0.8778 \times 10^{-3}$ & $0.7858 \times 10^{-3}$ & $0.6824 \times 10^{-3}$ & $1.7923 \times 10^{-3}$ & $1.1480 \times 10^{-3}$ \\
    \hline
    40 & $1.2391 \times 10^{-3}$ & $0.9278 \times 10^{-3}$ & $0.3857 \times 10^{-3}$ & $0.8020 \times 10^{-3}$ & $0.7896 \times 10^{-3}$ \\
    \hline
    60 & $0.4433 \times 10^{-3}$ & $0.4403 \times 10^{-3}$ & $0.7442 \times 10^{-3}$ & $1.6240 \times 10^{-3}$ & $1.3766 \times 10^{-3}$\\
    \hline
    80 & $1.0639 \times 10^{-3}$ & $1.0544 \times 10^{-3}$ & $1.4418 \times 10^{-3}$ & $1.1049 \times 10^{-3}$ & $4.1409 \times 10^{-3}$ \\
    \hline
    \end{tabular}
  \label{Poisson layer table}
\end{table}

\subsection{Heat equation}
This subsection explores the heat equation, where XPINN performed worse than PINN~\cite{Huelta2020}. We consider the same heat equation:
\begin{align}
    u_t - u_{xx} &= 0, \quad x\in[-1, 1],~t\in[0,1],
\end{align}
with the boundary and initial conditions
\begin{align}
    u(1, t) &= -e^{-\pi^2 t} + 0.6e^{-4\pi^2t} + 0.3e^{4t-4}\cosh(2)+0.1e^{t-1}\sinh(1),\\
    u(-1, t) &= -e^{-\pi^2 t} + 0.6e^{-4\pi^2t} + 0.3e^{4t-4}\cosh(-2)+0.1e^{t-1}\sinh(-1),\\
    u(x, 0) &= \cos(\pi x)+0.6\cos(2\pi x)+0.3e^{-4}\cosh(2x)+0.1e^{-1}\sinh(x).
\end{align}
The exact solution for the heat equation is 
\begin{align}
    u(x,t) = e^{-\pi^2t}\cos{\pi x} + 0.6e^{-4\pi^2t}\cos{2\pi x}+0.3e^{4t-4}\cosh{2x}+0.3e^{t-1}\sinh{x}.
\end{align}

Here, we follow a similar domain decomposition as the Ref.~\cite{Huelta2020}, where the interface is chosen to be $t=0.5$. The region for $t\leq 0.5$ will be denoted as subdomain 1, and the rest will be subdomain 2 in our experiment.

\begin{figure}[!htb]
\centering
   \begin{minipage}{0.45\textwidth}
     \centering
     \includegraphics[width=\linewidth, height=0.75\textwidth]{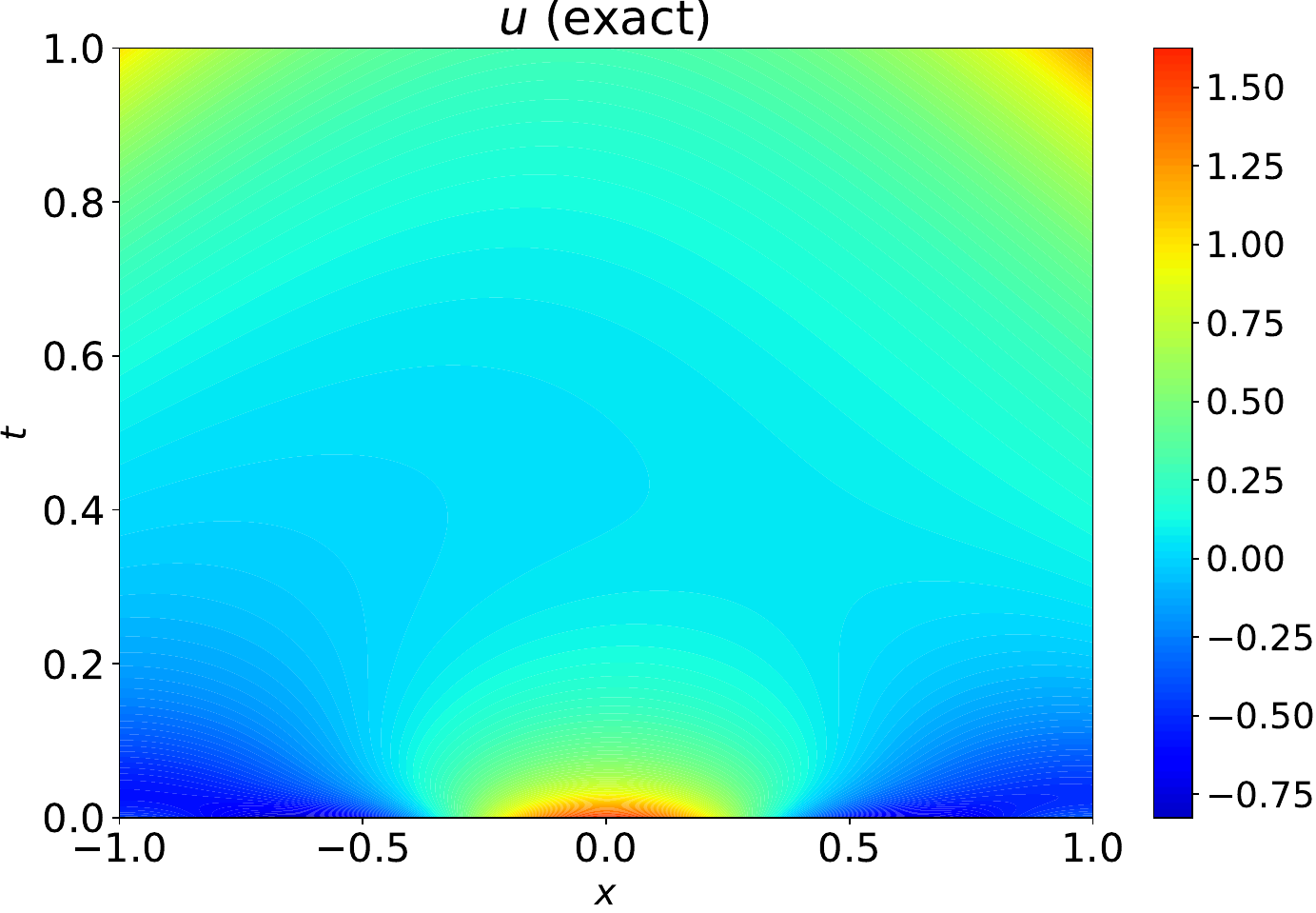}  
   \end{minipage}
   \begin{minipage}{0.45\textwidth}
     \centering
     \includegraphics[width=\linewidth, height=0.75\textwidth]{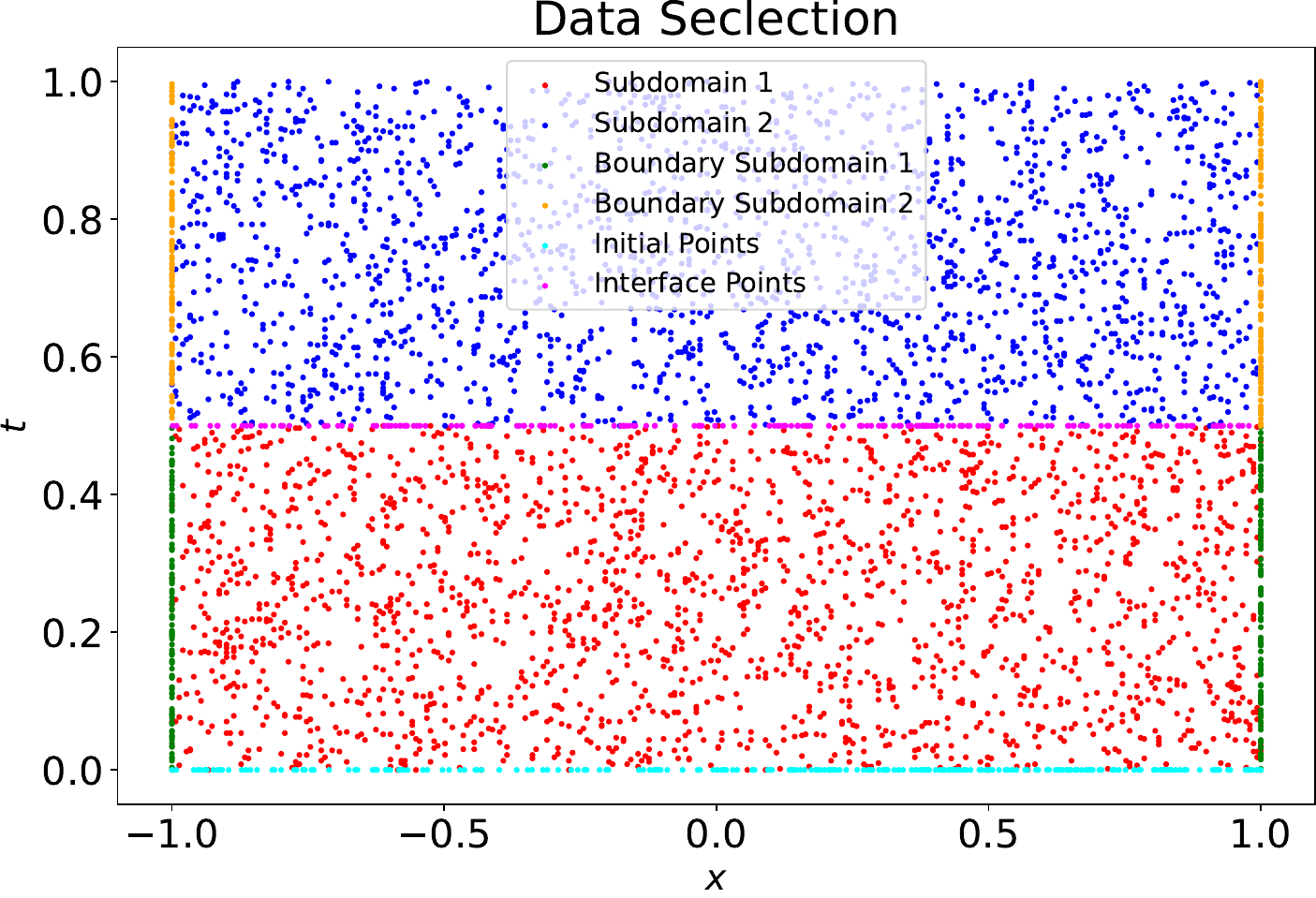} 
   \end{minipage}
    \caption{Exact solution and the data selection of the Heat equation.}\label{Heat data}    
\end{figure}

Table~\ref{Table 3} and Fig.~\ref{Heat data} give information about the training data points. We generate a $300\times 300$ grid uniformly over the domain $[-1,1]\times [0,1]$ to compute the L2 error. Additionally, 5000 points are uniformly generated on the interface $t = 0.5$. We select training points from the total of 95000 generated points as follows: initially, 100 boundary points and 2000 residual points are randomly chosen for each subdomain. Moreover, 200 initial points are randomly selected in subdomain 1. Within the 4400 selected points, we further select 50 boundary points, 50 initial points, and 1000 residual points in the IDPINN-init stage. For the IDPINN-main stage, 200 points are randomly chosen on the interface. Subsequently, the model is trained using these 4600 points for $14$k iterations by the Adam optimizer with a learning rate of $5.5\times 10^{-6}$. 

\begin{figure}[!htb]
\centering
    \begin{minipage}{0.45\textwidth}
    \centering
    \includegraphics[width=\linewidth, height=0.75\textwidth]{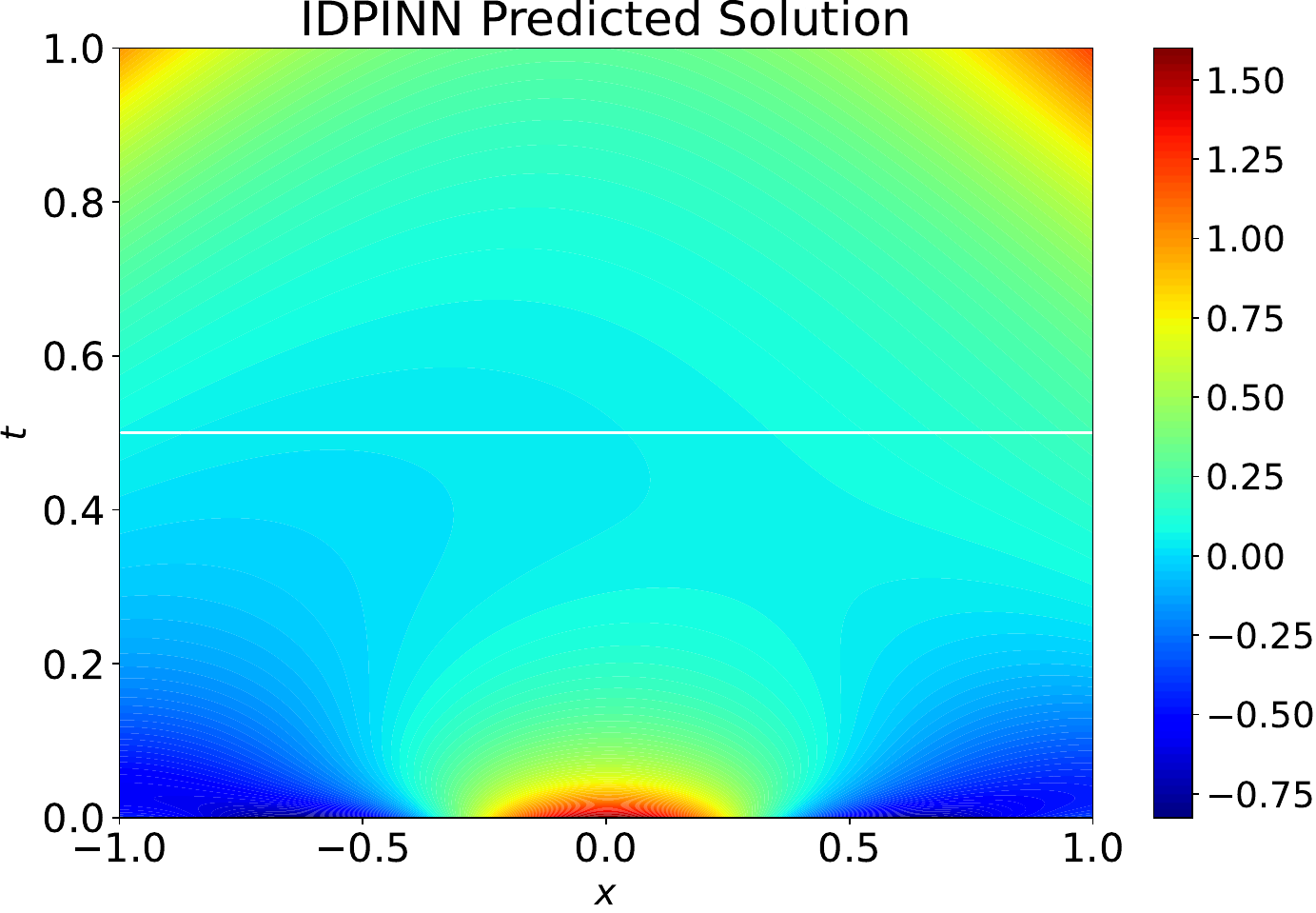}    
  \end{minipage}
  \begin{minipage}{0.45\textwidth}
      \centering
    \includegraphics[width=\linewidth, height=0.75\textwidth]{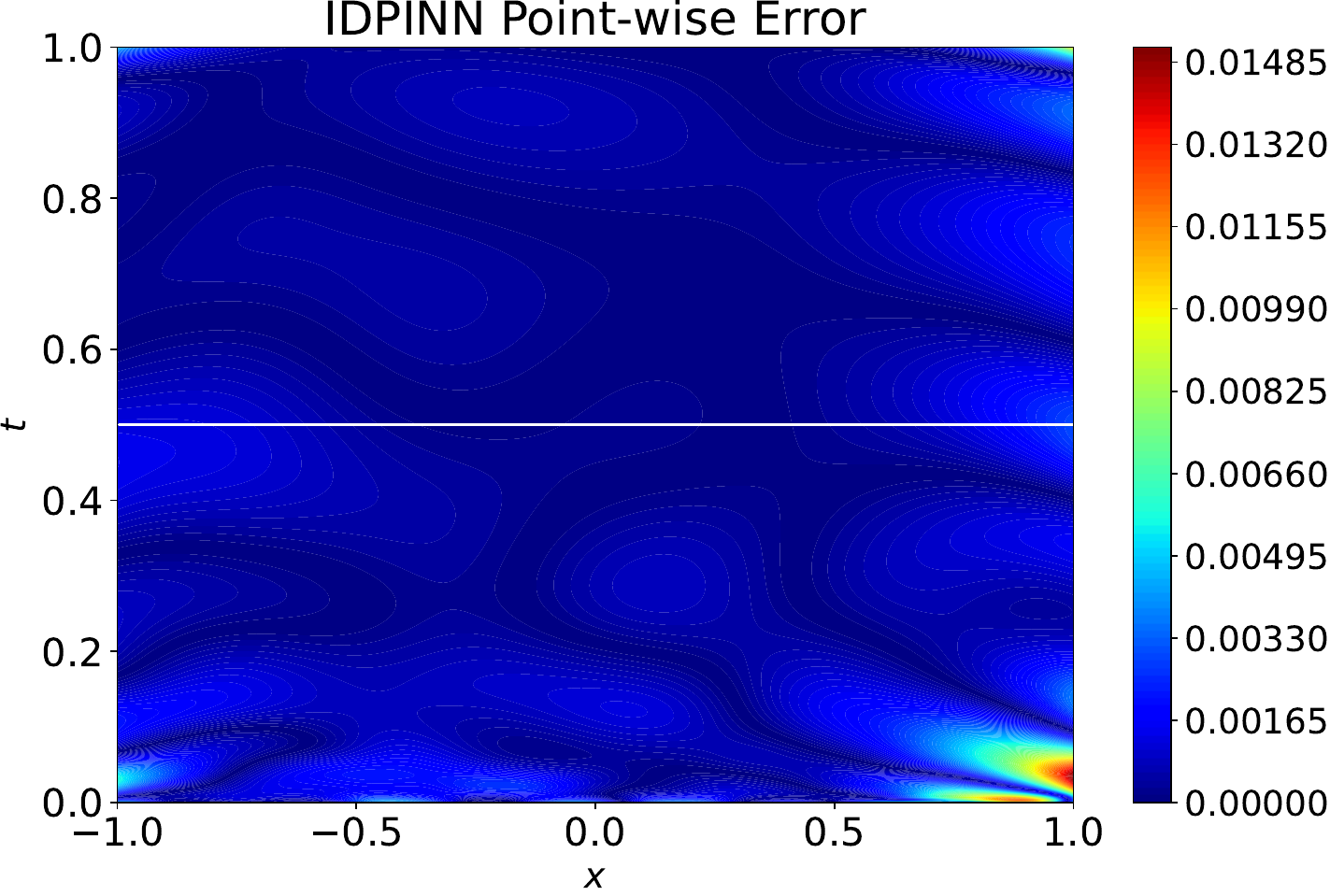}
  \end{minipage}
 \vspace{1em}

  \begin{minipage}{0.45\textwidth}
    \centering
    \includegraphics[width=\linewidth, height=0.75\textwidth]{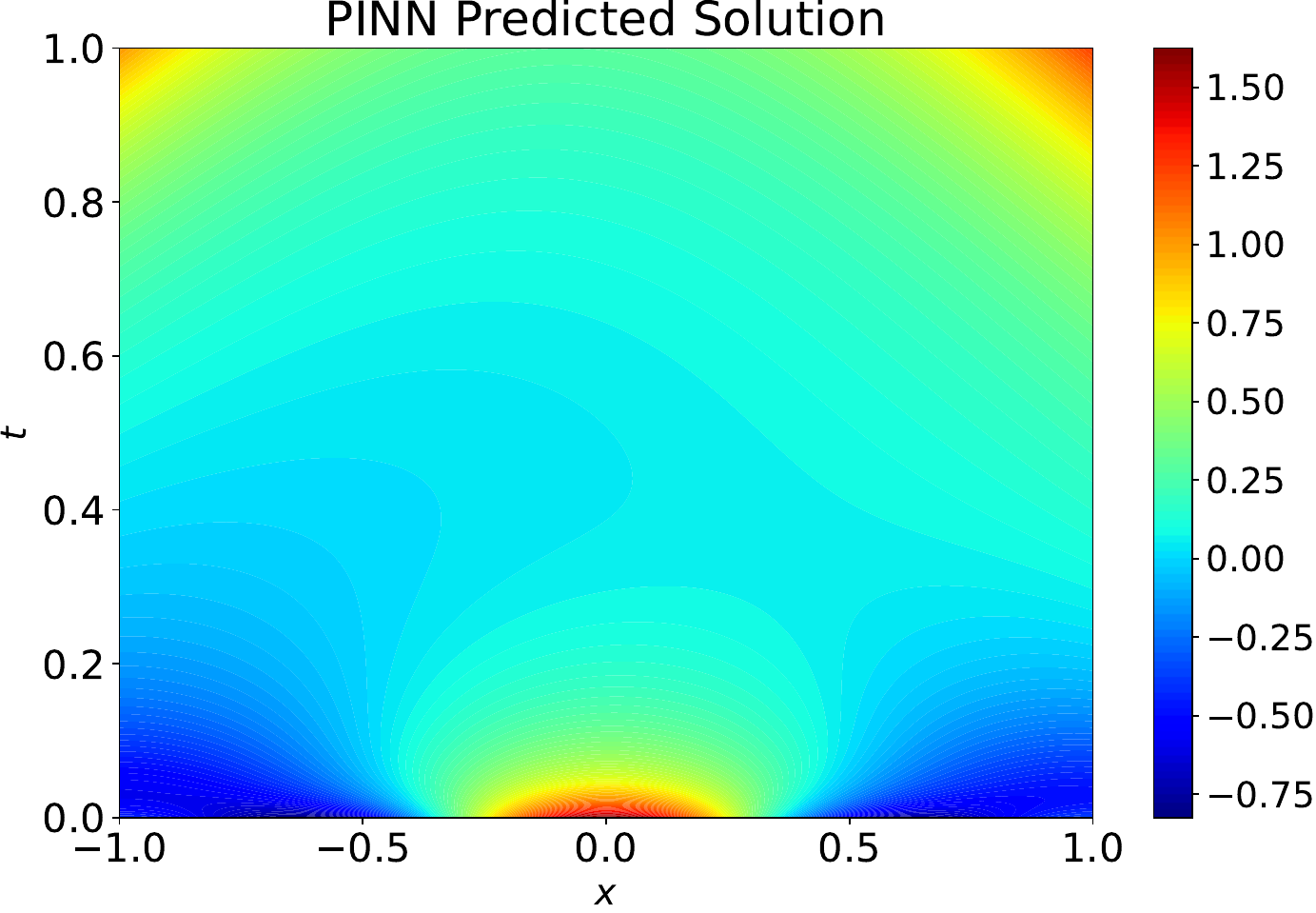}    
  \end{minipage}
  \begin{minipage}{0.45\textwidth}
      \centering
    \includegraphics[width=\linewidth, height=0.75\textwidth]{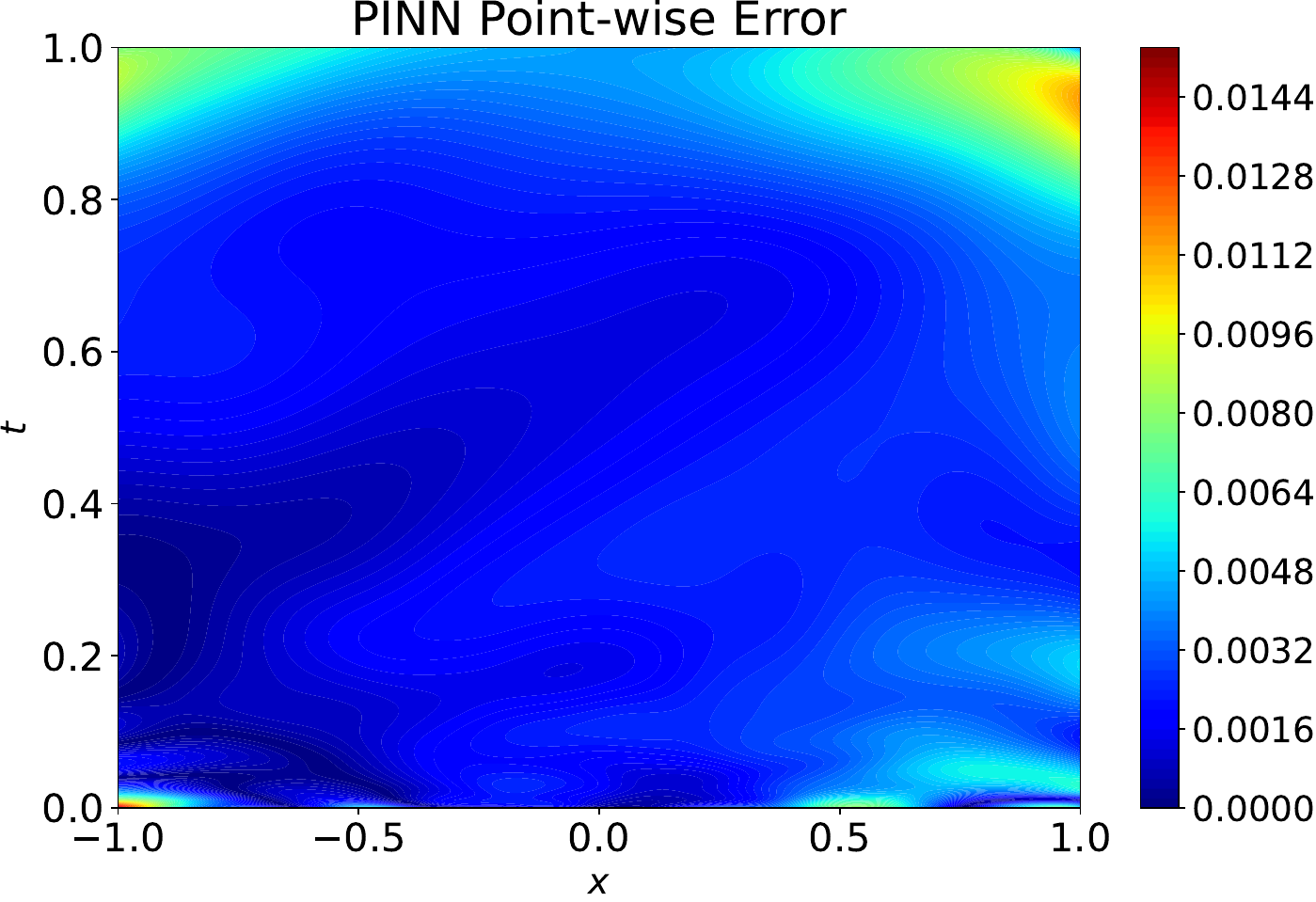}
  \end{minipage}

  \vspace{1em} 

  \begin{minipage}{0.45\textwidth}
    \centering
    \includegraphics[width=\linewidth, height=0.75\textwidth]{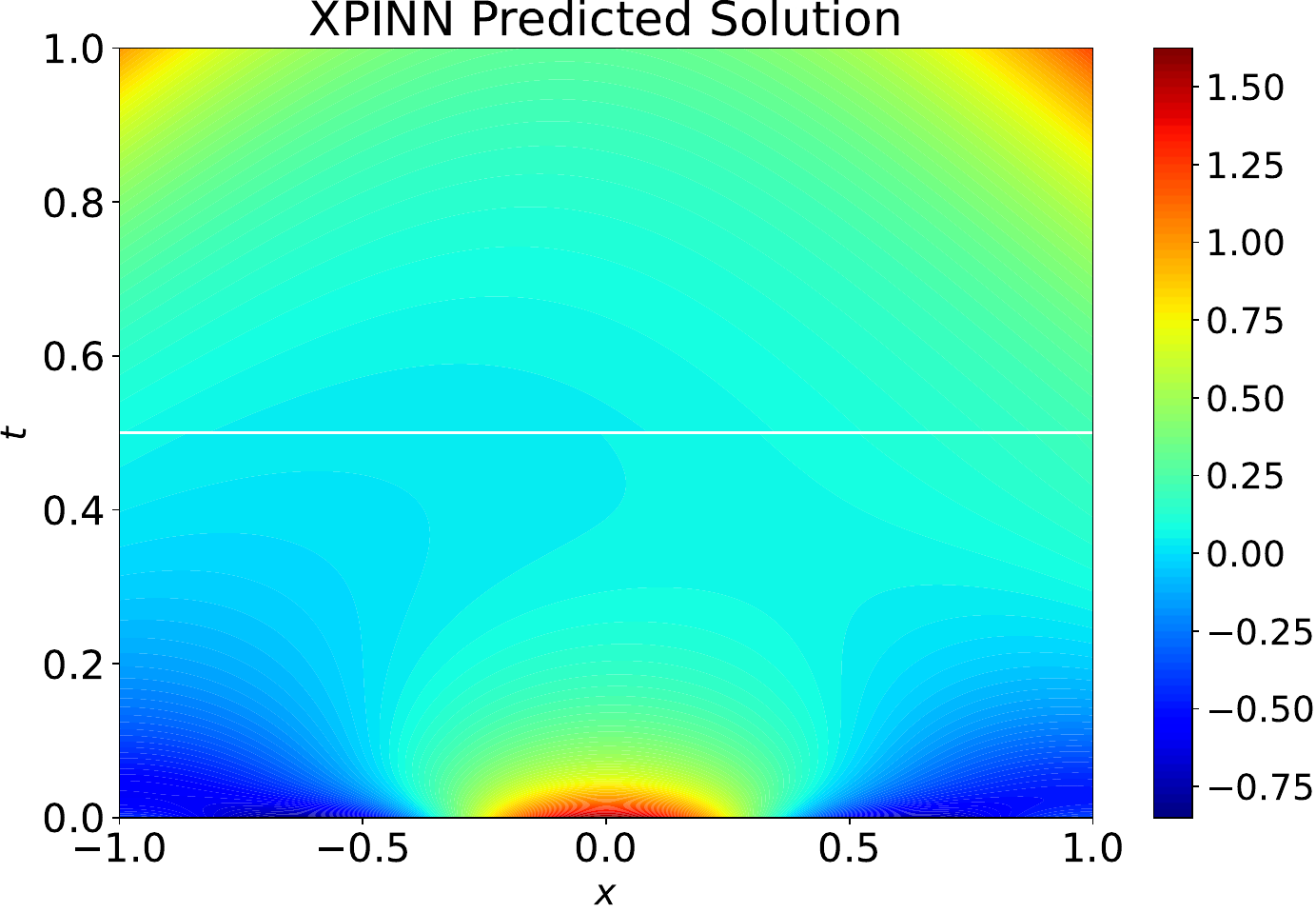}  
  \end{minipage}
  \begin{minipage}{0.45\textwidth}
    \centering
    \includegraphics[width=\linewidth, height=0.75\textwidth]{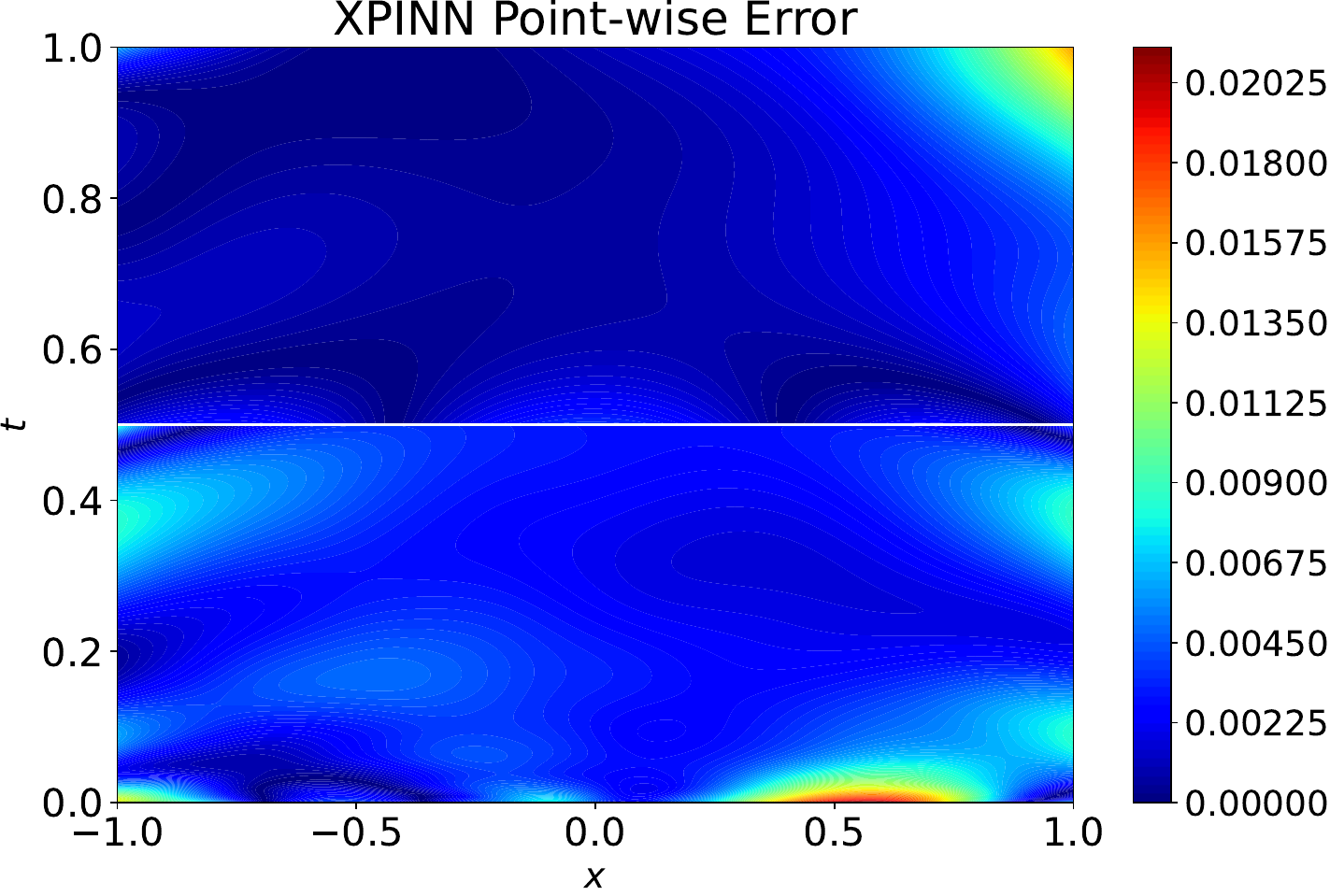}    
  \end{minipage}
  
   \caption{From up to down, left to right: Predicted solution and point-wise error of IDPINN-3 (L2 error: $3.51\times 10^{-3}$), PINN (L2 error: $6.22\times 10^{-3}$), and XPINN (L2 error: $9.30\times 10^{-3}$), to the Heat equation.}\label{Heat result}    
\end{figure}

For the numerical experiment, we use a 9-layer neural network with 20 neurons per layer. The weights for IDPINN-3 are set as $\lambda_1=2$, $\lambda_2 = 10$, $\lambda_3 = 20$, $\lambda_4 = 20$, $\lambda_5 = 5$, and $\lambda_6 = 5$. The XPINN is configured with $\lambda_{residual} = \lambda_{avg} = 20$, while other parameters remain consistent with IDPINN-3. Also, we train a PINN using 4600 training points within the defined grid, employing the same network structure and learning rate. Visualizations of PINN, XPINN, and IDPINN-3 predictions are shown in Fig.~\ref{Heat result}.
\begin{table}[!h]
  \centering
  \caption{Number of points and iterations in each stage (Heat equation). The training data in the IDPINN-init stage are selected from the training points in the IDPINN-main stage.}
  \begin{tabular}{|l|l|c|c|c|c|c|}
    \hline
    stage & region &  residual  &  boundary & initial  & interface & iterations\\
    \hline
    IDPINN-init & all-region  & 1000  & 50 & 50 & NA & 1000\\
    \hline
 \multirow{2}{*}{IDPINN-Main}  & sub-domain 1 (red) & 2000  & 100 & 200 & \multirow{2}{*}{200} & \multirow{2}{*}{14000}\\
    \cline{2-5}
    & sub-domain 2 (blue) & 2000 & 100 & NA & & \\
    \hline
    \end{tabular}
    \label{Table 3}
\end{table}

As shown in Fig.~\ref{Heat result}, XPINN exhibits significant point-wise errors near the interface and boundary. After $15$k iterations, XPINN yields a relative L2 error of $9.30 \times 10^{-3}$, while PINN achieves a slightly lower error of $6.22 \times 10^{-3}$. The experiments with different settings in \cite{Huelta2020} also show that PINN outperforms XPINN for the heat equation, even with an increased number of training points. However, IDPINN-3 achieves an even smaller L2 error of $3.51 \times 10^{-3}$ compared to PINN, showcasing its superiority in accuracy. Moreover, IDPINN-3 exhibits reduced point-wise errors near the interface and boundary, demonstrating its effectiveness in handling boundary-dominated problems.

\subsection{Viscous Burgers equation}
The viscous Burgers equation, expressed as \eqref{Burgers1}-\eqref{Burgers3}, serves as a fundamental PDE in fluid dynamics and nonlinear physics. It provides a simplified model for a wide range of complex phenomena, including turbulence and shock waves. Combining the influences of convection and diffusion, the equation encapsulates the nonlinear advection and viscous dissipation characteristics of fluid behavior.

We consider the whole domain as $(x,t)\in [-1,1]\times [0,1]$ and divide it with a circle of radius $0.3$ centered at $(0.5, 0.5)$. The region outside the circle is designated as subdomain 1, while the region inside is labeled as subdomain 2. For testing, we utilize the same dataset as~\cite{Rasissietal2019}, which provides ground truth for the testing points.

Table~\ref{Table 1} provides details about the training data points. A $300\times 300$ grid is uniformly generated over the domain $[-1,1]\times [0,1]$. Additionally, 5000 points are uniformly generated on the circle $\{(x, t): (x-0.5)^2 + (t-0.5)^2 = 0.3^2\}$. Training points are selected from the 95000 generated points. Initially, 120 boundary points and 2450 residual points are randomly chosen for subdomain 1, along with 450 residual points for subdomain 2. Furthermore, 60 initial points are randomly selected in subdomain 1, along with 100 points on the interface.
\begin{table}[!h]
  \centering
  \caption{Number of points and iterations in each stage (Burgers equation). The training data in the IDPINN-init stage are collected from the training sample in the IDPINN-main stage.}
  \begin{tabular}{|l|l|r|r|r|r|r|}
    \hline
    stage & region &  residual  &  interface & boundary & initial  & iterations\\
    \hline
    IDPINN-init & all-region  & 1000  & NA & 60 & 30 &  1000\\
    \hline
 \multirow{2}{*}{IDPINN-Main}  & sub-domain (inside) & 450& \multirow{2}{*}{100} & 0 & 0 &  \multirow{2}{*}{14000}\\
    \cline{2-3}
    \cline{5-6}
    & sub-domain (outside) & 2450 & & 120 & 60 & \\
    \hline
    PINN & all-region  & 3000  & NA & 120 & 60 &  15000\\
    \hline
    \end{tabular}
    \label{Table 1}
\end{table}

\begin{figure}[!htb]

  \centering
  \includegraphics[width=.4\linewidth]{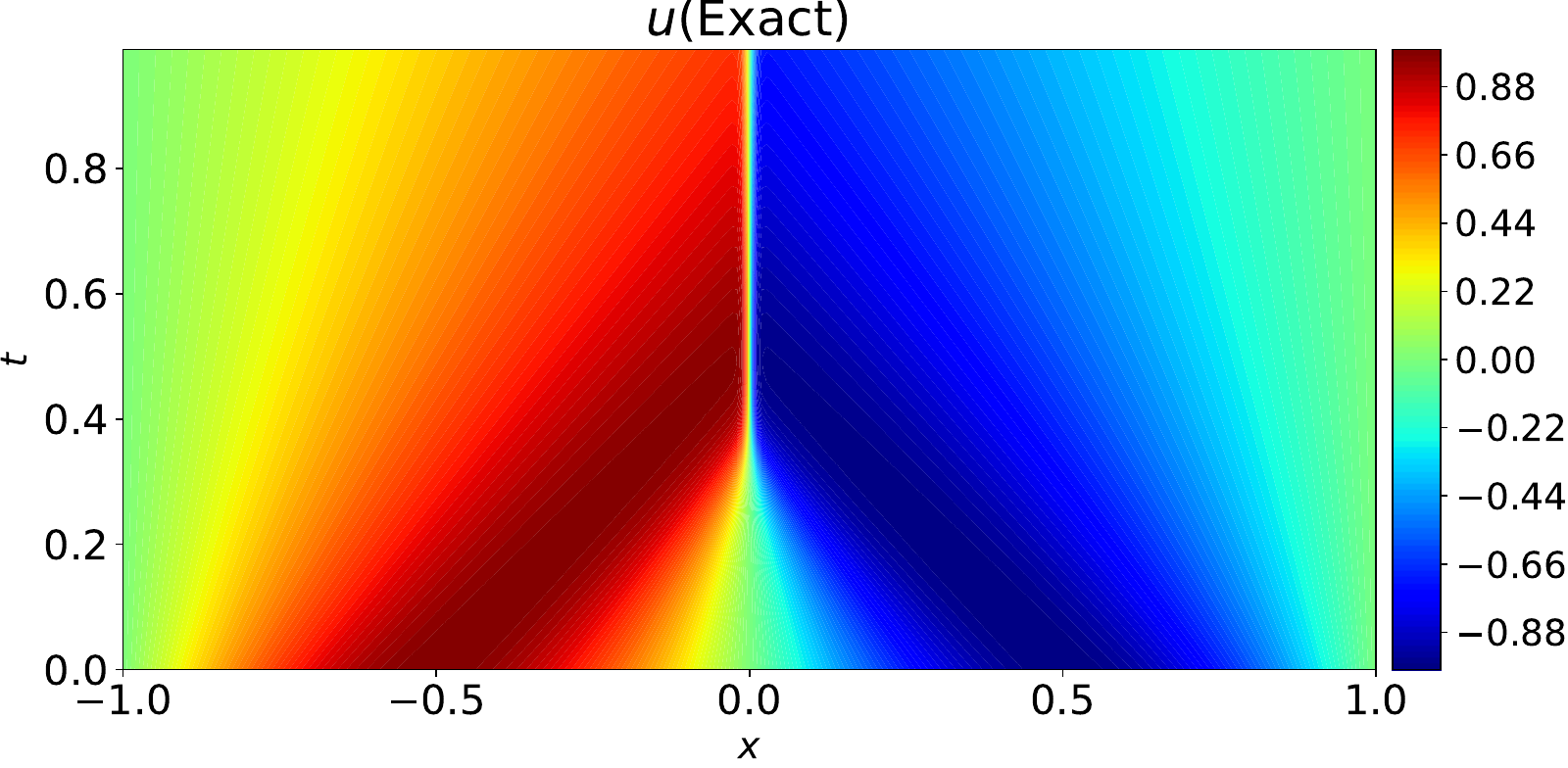}
  \vspace{1em}

  \begin{subfigure}{0.4\textwidth}
    \centering
    \includegraphics[width=\linewidth]{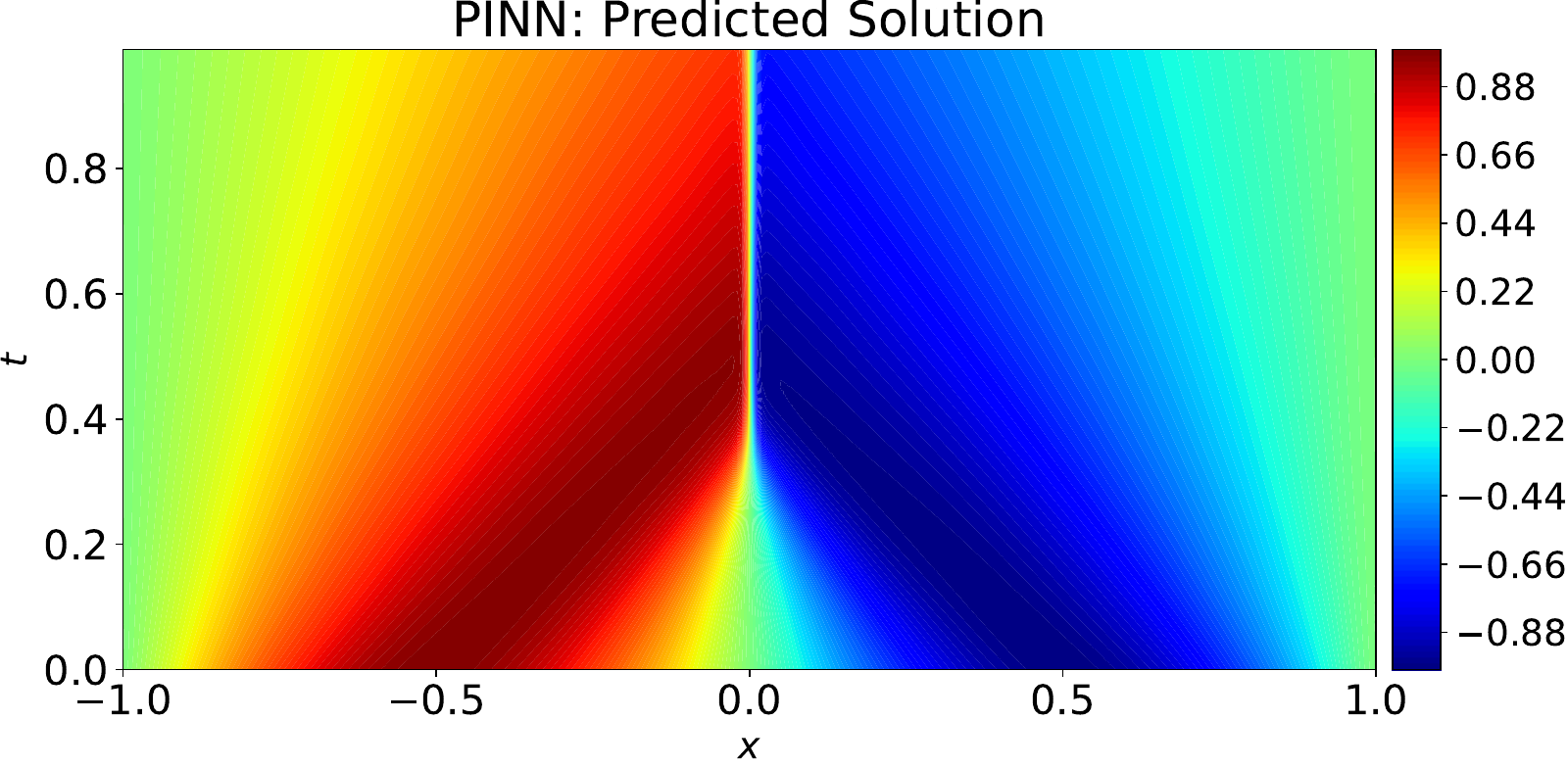}    
  \end{subfigure}
  \begin{subfigure}{0.4\textwidth}
    \centering
    \includegraphics[width=\linewidth]{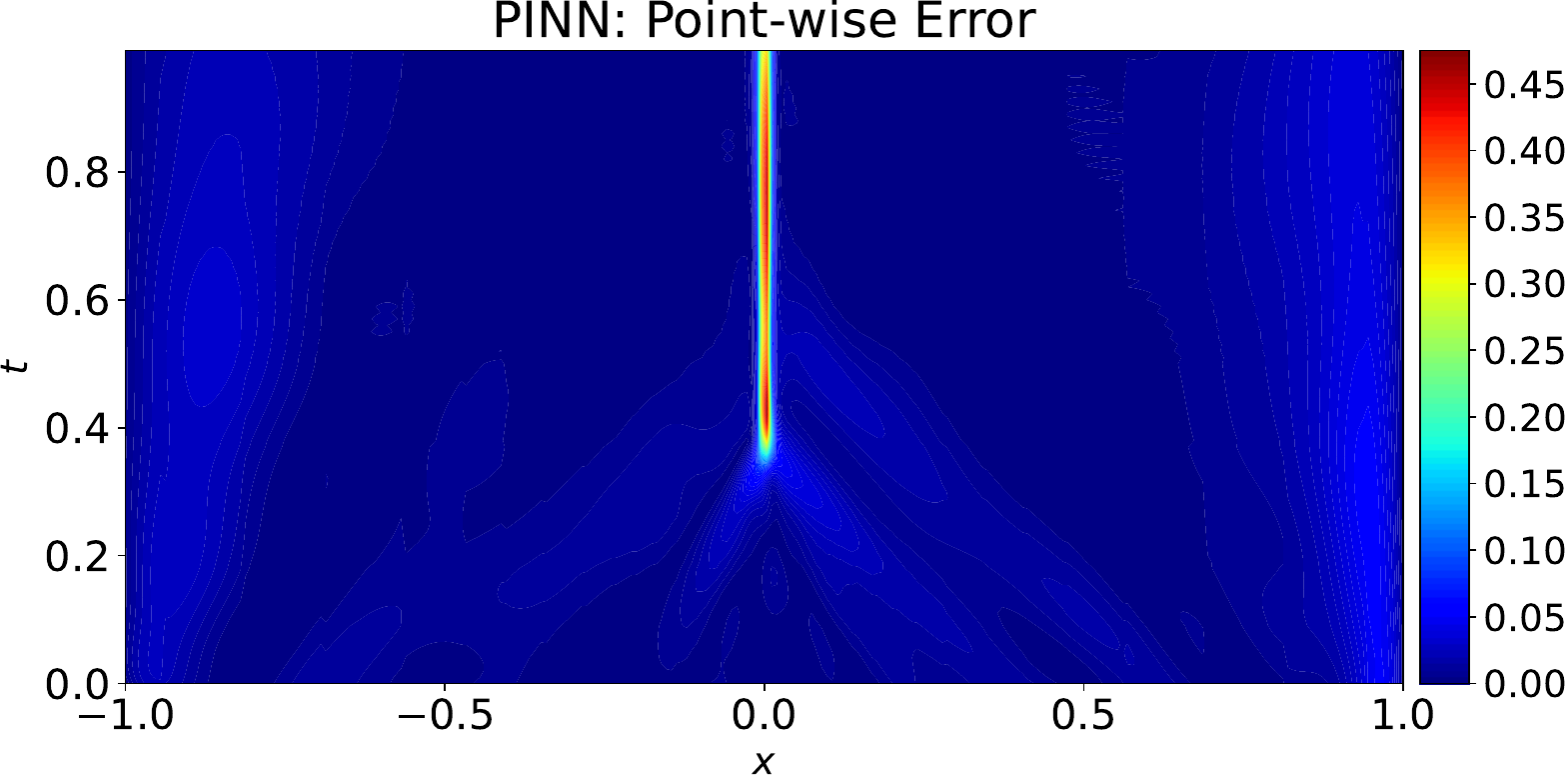}
  \end{subfigure}

  \vspace{1em} 

  \begin{subfigure}{0.4\textwidth}
    \centering
    \includegraphics[width=\linewidth]{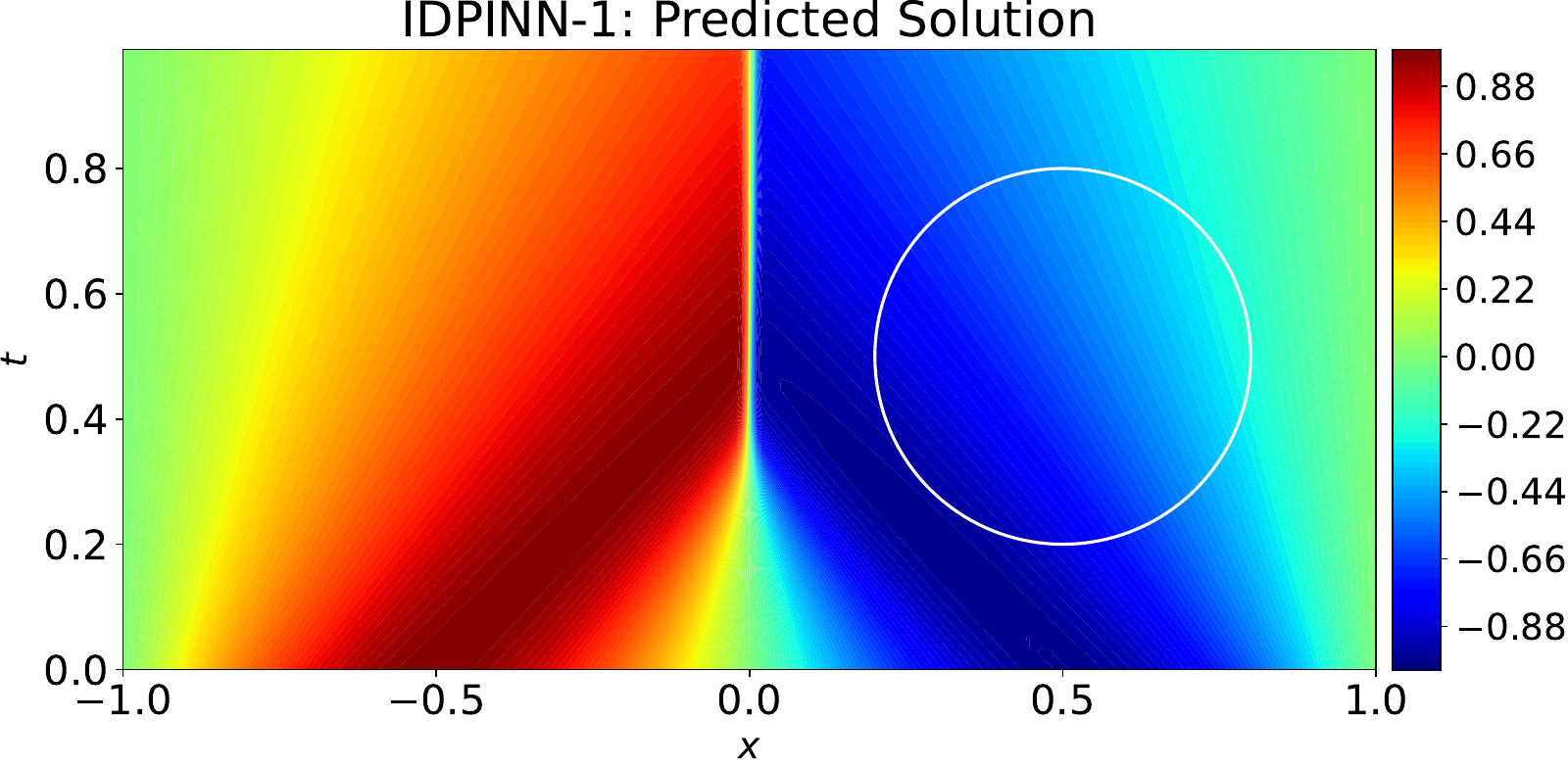}  
  \end{subfigure}
  \begin{subfigure}{0.4\textwidth}
    \centering
    \includegraphics[width=\linewidth]{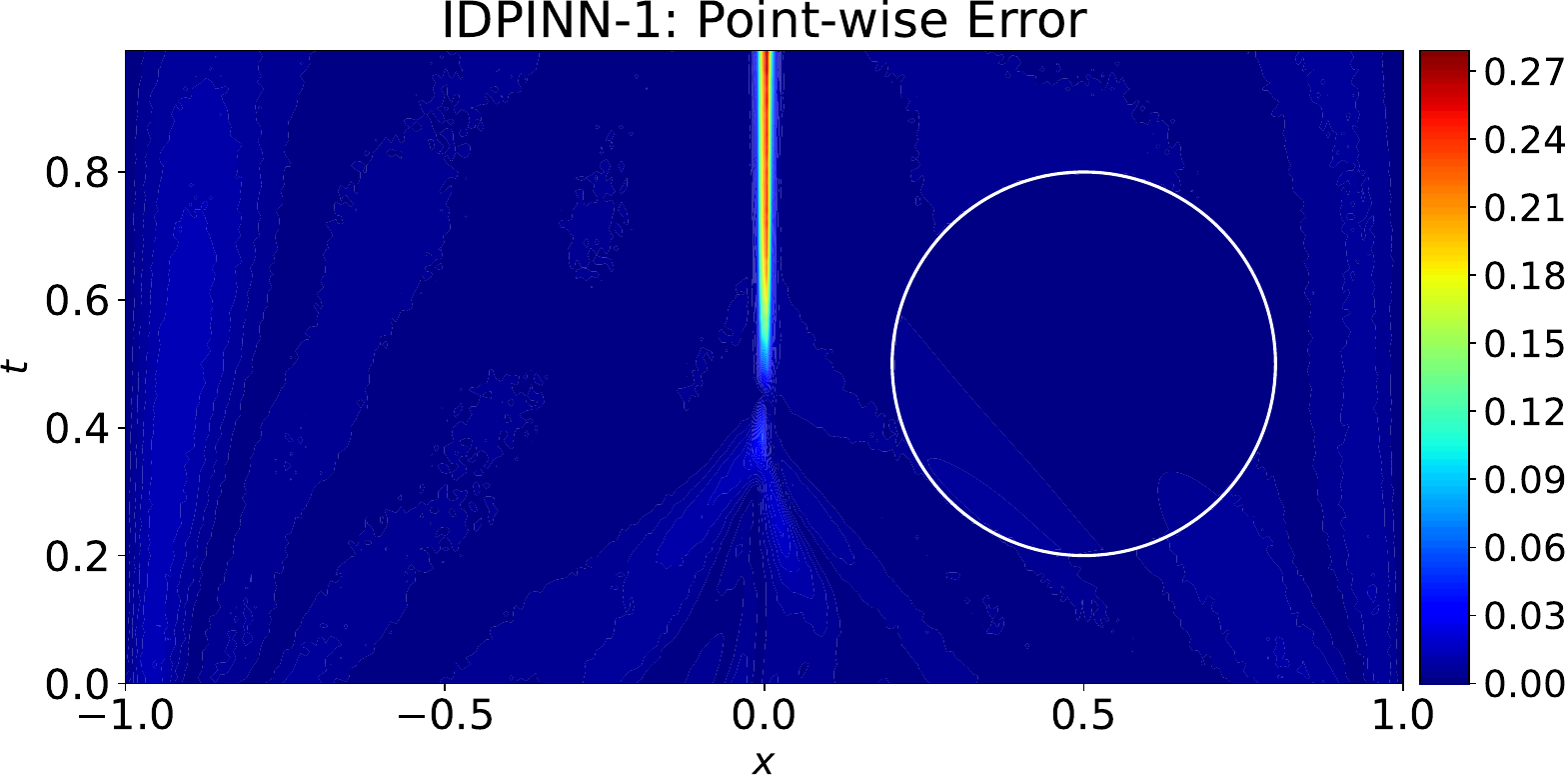}    
  \end{subfigure}
    \vspace{1em} 

  \begin{subfigure}{0.4\textwidth}
    \centering
    \includegraphics[width=\linewidth]{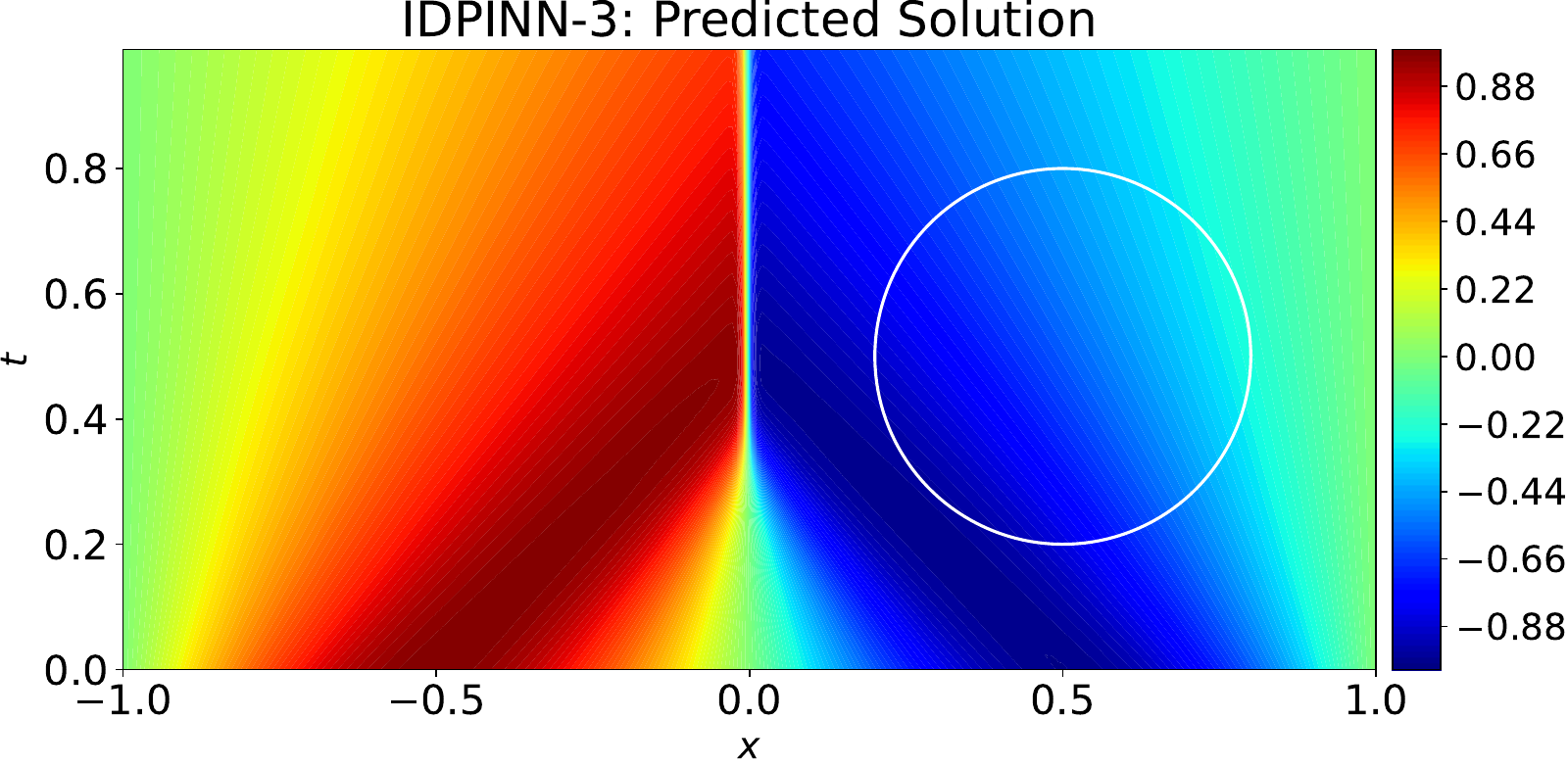}  
  \end{subfigure}
  \begin{subfigure}{0.4\textwidth}
    \centering
    \includegraphics[width=\linewidth]{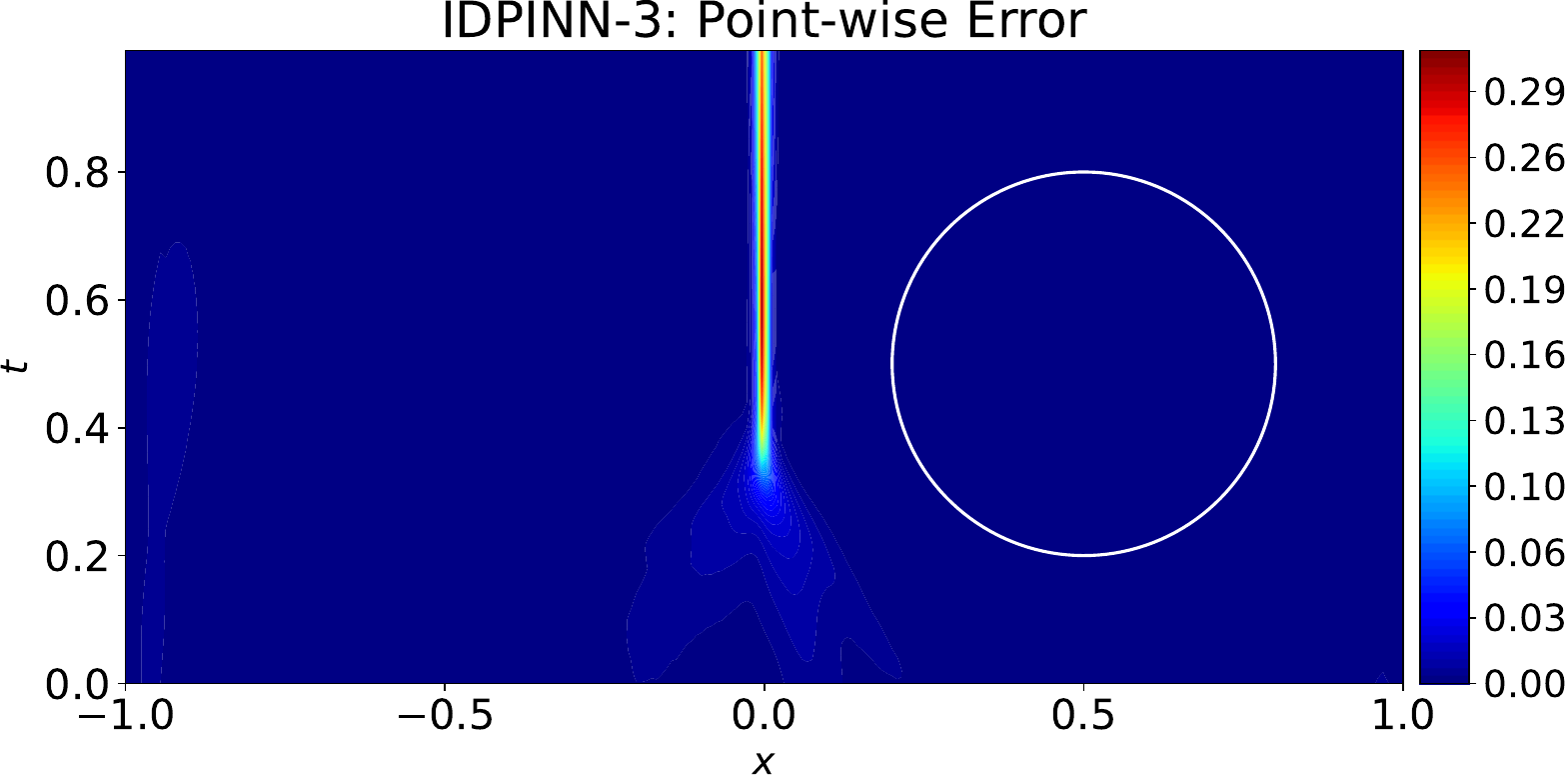}    
  \end{subfigure}
    \caption{The numerical results for the Burgers equation. Top row: the exact solution. Second row: the predicted solution and point-wise error for PINN (L2 error: $4.03\times 10^{-2}$). Third row: the predicted solution and point-wise error for IDPINN-1 (L2 error: $2.92\times 10^{-2}$). Last row: the predicted solution and point-wise error for IDPINN-3 (L2 error: $3.35\times 10^{-2}$).}\label{Burgers result}\end{figure}

We compare IDPINN-1, IDPINN-3, and PINN for this experiment. The network architecture has 7 hidden layers with 20 neurons in each hidden layer. The parameters for IDPINN-1 are set as $\lambda_1 = 1$, $\lambda_2 = 20$, $\lambda_3 = 5$, $\lambda_{4} = 20$, $\lambda_{5} = 2$ and $\lambda_{6} = 0$.  Letting $\lambda_6=2$ yields IDPINN-3. In addition, we use the same parameters ($\lambda_1$, $\lambda_2$, and $\lambda_3$) for PINN. We train the models using the Adam optimizer with a learning rate of $7 \times 10^{-4}$. The numerical results are shown in Fig.~\ref{Burgers result}. 

The improvement of IDPINN over PINN in this experiment may not be substantial, and the majority of the error occurs at $x=0$, where the shock wave appears. Since IDPINN assumes smoothness across the interface, we avoid placing the interface over the shock. For such problems, adaptive sampling strategies may be beneficial~\cite{Chenxielta2023}\cite{dawelta2022}\cite{Gaoelta2023}. These strategies dynamically adjust the sampling density based on the solution behavior, potentially improving accuracy and efficiency in regions of interest.

\section{Conclusion}
This paper introduces IDPINN, a novel framework that combines advanced initialization and domain decomposition techniques to enhance physics-informed neural networks (PINNs). By formulating special loss functions for interface conditions, IDPINN effectively recovers smooth solutions, while its initialization from a small dataset enhances training efficiency, saving time and computing resources. Experimental results demonstrate the superiority of IDPINN over XPINN and PINN. Additionally, we explore IDPINN's applicability to various domain decomposition strategies, including straight lines and irregular closed curves. It demonstrates the versatility of our approach, making it applicable across a spectrum of domain decomposition scenarios. 

\section*{Acknowledgments}
This work was partially supported by the Guangdong Provincial Key Laboratory of Mathematical Foundations for Artificial Intelligence (2023B1212010001), Shenzhen Stability Science Program, and the Shenzhen Science and Technology Program under grant no. ZDSYS20211021111415025. 

\bibliographystyle{unsrt}  
\bibliographystyle{model1-num-names}
\bibliographystyle{plain}

\end{document}